\documentclass{article}
\usepackage[T1]{fontenc}
\usepackage{iclr2027_conference,times}

\usepackage{amsmath,amsfonts,bm}

\DeclareRobustCommand{\Fullref}[2]{\hyperref[#2]{\mbox{#1~\ref*{#2}}}}
\newcommand{\Tabref}[1]{\Fullref{Table}{#1}}
\newcommand{\Appref}[1]{\Fullref{Appendix}{#1}}
\DeclareRobustCommand{\Tabrange}[2]{\hyperref[#1]{\mbox{Tables~\ref*{#1}--\ref*{#2}}}}
\newcommand{\Tabpair}[2]{\Tabref{#1} and \Tabref{#2}}

\def\Figref#1{\Fullref{Figure}{#1}}

\def\Secref#1{\Fullref{Section}{#1}}

\def\eqref#1{\Fullref{equation}{#1}}
\def\Eqref#1{\Fullref{Equation}{#1}}

\def\1{\bm{1}}

\DeclareMathAlphabet{\mathsfit}{\encodingdefault}{\sfdefault}{m}{sl}
\SetMathAlphabet{\mathsfit}{bold}{\encodingdefault}{\sfdefault}{bx}{n}

\usepackage{graphicx}
\usepackage{booktabs}
\usepackage{colortbl}
\usepackage{placeins}
\usepackage{flafter}
\usepackage{etoolbox}
\usepackage{needspace}
\AtBeginEnvironment{table}{\setlength{\belowcaptionskip}{6pt}}
\AtBeginEnvironment{table*}{\setlength{\belowcaptionskip}{6pt}}
\definecolor{VolticTint}{HTML}{EDF4FA}
\usepackage{hyperref}
\usepackage{url}

\title{Voltic: Distinguishing Volatility from Stochasticity in Recurrent Memory}
\author{Parsa Hejabi \quad Morteza Dehghani \quad Payam Piray\\
University of Southern California\\
\texttt{\{hejabi,mdehghan,piray\}@usc.edu}}
\iclrfinalcopy

\begin{document}
\maketitle
\lhead{Preprint}

\begin{abstract}

Recurrent sequence models must decide how strongly to overwrite their memory at each token. Read as Bayesian filtering, this write is the gain of a Kalman update, set by uncertainty from two sources that pull it in opposite directions: volatility, how quickly the underlying associations change, and stochasticity, how noisy each observation of them is. First, we show that the update of gated delta-rule memories is the form this filter takes under isotropic uncertainty. Next, we introduce Voltic, a recurrent memory that keeps the covariance anisotropic and makes both noise variances input-dependent, so the write is vector-valued and carries uncertainty accumulated over the sequence. A dense covariance would have to be propagated token by token, ruling out the parallel training these models depend on. We therefore give two assumed-density approximations, diagonal and quasi-diagonal, both of which leave the memory update in delta-rule form and reuse its chunked kernels. On controlled recall tasks in which associations change and observations are corrupted, Voltic leads all baselines. On the task combining volatility and stochasticity, its margin over the strongest baseline is larger at both extrapolation sizes than at the training sizes. In 45M-parameter language models it leads an eight-task reasoning average and achieves higher retrieval accuracy beyond the training context length than gated baselines, at throughput close to those baselines. Deriving the write from an uncertainty recursion therefore makes memory more responsive to change.

\end{abstract}

\section{Introduction}\label{sec:introduction}
Recurrent sequence models compress a stream of tokens into a fixed-size memory, and at every token they must decide how strongly the new token should change what is stored. This decision has a natural statistical reading. If the memory is taken to be the posterior mean of a latent state that the token stream observes, then updating it is Bayesian filtering, and the strength of each write is the Kalman gain \citep{kalman1960new}: how much weight a new observation receives relative to the evidence already accumulated. The gain is set by uncertainty, so the question of how a memory should write becomes a question of what that uncertainty is made of.

Under this view, the uncertainty that sets the gain has two sources: \emph{Volatility} is how quickly the latent state changes, the magnitude of the process noise; \emph{stochasticity} is how noisy each observation is, the magnitude of the observation noise. The two act on the gain in opposite directions. Higher volatility makes past evidence less relevant and calls for a stronger write; higher stochasticity makes the new observation less informative and calls for a weaker one. Both raise uncertainty, so a filter that represents only their combined effect cannot respond appropriately to either. Estimating the two online is a long-standing problem in adaptive filtering \citep{sarkka2009recursive, piray2021computational}, and normative accounts of learning treat their dissociation as central, showing that failing to make it produces characteristic failures of adaptation \citep{fang2026inferring, smith2026early}. These accounts infer both noise parameters jointly with the latent state, using a Rao-Blackwellized particle filter \citep{doucet2000rao, piray2021computational} or a deterministic quantile grid in place of the particles \citep{chen2026categorical}. Such methods are designed for behavioral tasks, a few static parameters inferred from hundreds of observations. Their representation of the noise parameters grows exponentially with the number of parameters, so they do not extend to the input-dependent, per-coordinate rates a high-dimensional memory trained on billions of tokens would require.

Existing recurrent memories can be read as filters in which this structure is heavily restricted. The delta-rule update used in these models \citep{schlag2021linear} is the classical least-mean-squares rule of adaptive filtering, and the Kalman filter is its uncertainty-weighted counterpart, setting the learning rate from the filter's uncertainty rather than fixing it \citep{sutton1992gain, dayan2000learning}. Gated delta-rule models such as GDN \citep{ICLR2025_4904fad1} and KDA \citep{kimiteam2025kimilinearexpressiveefficient} update a key-value memory by erasing its component along the current key and writing the new value, scaled by a learned write strength. We show that this update is a filtering step under isotropic uncertainty: the write strength is the gain of a filter whose uncertainty is the same in every direction of key space, so it is a single scalar per head, and it is produced by a projection of the current token rather than by accumulated uncertainty. Filtering has also been used directly as a sequence mixer, most closely in Kalman Linear Attention \citep{shaj2026kalman}, which filters each channel with a diagonal precision and learned, time-invariant process noise. What remains missing is a memory in which both sources of uncertainty depend on the input, uncertainty varies across key space, and the update keeps the delta-rule form that makes these models efficient to train.

We introduce \textbf{Voltic}, a recurrent memory that treats the sequence as \emph{vol}atile and stochas\emph{tic}, and derives its write strength accordingly. Both noise variances are learned functions of the input, entering the filter at the prediction and observation steps, so the write responds to the two sources in opposite directions without inference over either. The resulting gain is vector-valued along the key dimension and carries uncertainty accumulated over the sequence. Exact inference would propagate a dense covariance, which cannot be evaluated in parallel across tokens, so we develop two assumed-density approximations \citep{opper1999bayesian, minka2001family}: Voltic-D keeps a diagonal covariance with a learned per-coordinate decay, and Voltic-Q retains covariance within short blocks. Both leave the memory update in delta-rule form and reuse its parallel algorithms. Both properties are ordinary in real sequences: what a name or a value refers to can change over a document, and a mention of it can be wrong. Existing recall benchmarks assume neither, observing each association once and faithfully \citep{ICLR2024_448fc91f}, so a model is never asked to respond differently to a change and to a corrupted observation. We therefore vary the two sources separately and together.

We make four contributions. (i) We introduce Voltic, a recurrent memory whose write strength is derived from a filtering recursion rather than learned as a free gate, with two tractable approximations that keep the update in delta-rule form. (ii) We give efficient forward and backward kernels for both variants, bringing throughput close to gated recurrent-attention baselines at the sequence lengths and batch sizes used for language-model training. (iii) We introduce three controlled recall tasks in which associations change during the sequence, observations of them are corrupted, or both. (iv) We evaluate Voltic against recurrent-attention, state-space, and full-attention baselines on these tasks, language modeling, in-context retrieval, and throughput.
\section{Related Work}\label{sec:related_work}

\paragraph{Recurrent memories and online learning.}
Linear attention expresses sequence mixing through a fixed-size recurrent state \citep{katharopoulos2020transformers}. Delta-rule models improve how this state stores key--value associations through error-correcting writes; \Secref{sec:method_gated} shows that the write strength of their gated extensions plays the role of a filtering gain. State-space models provide another route to recurrent sequence mixing, with input-dependent state updates \citep{gu2024mamba, ICLR2026_8abd2043}. A parallel line reads these updates as solutions to a per-token regression problem, unifying delta-rule and gated variants as online learners \citep{liu2025longhorn, wang2026test}.

\paragraph{Filtering-based sequence models.}
Deep Kalman Filters \citep{krishnan2015deep}, Recurrent Kalman Networks \citep{becker2019recurrent}, and learned latent dynamics for planning \citep{hafner2019learning} combine neural representations with probabilistic state estimation, but their recurrent updates limit use at long context. Closest to this work is Kalman Linear Attention (KLA; \citealp{shaj2026kalman}), which derives sequence mixing from Bayesian filtering and parallelizes its information-form update through a scan. KLA keeps a diagonal precision per channel and treats process noise as a learned time-invariant parameter; Voltic instead carries uncertainty over key coordinates, retaining short-range covariance in Voltic-Q, and makes both noise variances input-dependent. We compare KLA on MQAR, language modeling, in-context retrieval, and throughput; \Appref{app:software} specifies the different KLA backends used for these studies. MesaNet \citep{vonoswald2026mesanet} and Gated KalmaNet \citep{peng2026gated} solve a related regularized least-squares objective at each step, but assume a static latent state and require iterative solvers.

\paragraph{Assumed-density filtering.}
Projecting the posterior back onto a fixed family at each step, minimizing KL divergence to the exact posterior, is a standard route to keeping the state bounded \citep{opper1999bayesian, minka2001family}; \Secref{sec:method_diagonal} applies it here. Temporal parallelization of Bayesian filters has also been studied directly \citep{sarkka2021temporal}, using lifted representations to obtain associativity; we instead keep the memory update in delta-rule form so that it reuses the chunked algorithms already developed for that family.

\section{Voltic: Memory Updates from Uncertainty}
\label{sec:implementation}
\label{sec:method}

\subsection{A filtering model of recurrent memory}
\label{sec:implementation_recurrence}
In language modeling, a model predicts each token from the tokens before it. Linear recurrent models, such as linear attention and Mamba \citep{katharopoulos2020transformers, gu2024mamba}, summarize that history in a fixed-size state, updated as each token arrives. We treat this summary as the result of inference: the token stream is modeled as the output of a dynamical system whose latent state evolves over time, and each observation $x_t\in\mathbb{R}^{P}$, a learned projection of the current token, is generated noisily from it.

The latent state is a matrix $Z_t\in\mathbb{R}^{N\times P}$, and each observation reads it along a direction $k_t\in\mathbb{R}^{N}$, also computed from the token and normalized to unit length. The model makes two structural assumptions: the latent state evolves as a Markov process, and each observation depends only on the current state. For each observation dimension $j$, with latent column $z_{t,j}\in\mathbb{R}^{N}$,
\begin{equation}
 \begin{aligned}
 z_{t,j}\mid z_{t-1,j}&\sim\mathcal{N}\bigl(A_tz_{t-1,j},\,\operatorname{diag}(v_t)\bigr),\\
 x_{t,j}\mid z_{t,j}&\sim\mathcal{N}\bigl(k_t^\top z_{t,j},\,s_t\bigr),
 \end{aligned}
 \label{eq:latent_model}
\end{equation}
with diagonal transition $A_t=\operatorname{diag}(\alpha_t)$, $\alpha_t\in(0,1]^{N}$. The variance of the transition, $v_t\in\mathbb{R}_{>0}^{N}$, is the volatility of the latent state and may differ across its $N$ coordinates; the variance of the observation, $s_t\in\mathbb{R}_{>0}$, is its stochasticity. Volatility is a vector because the latent state is, while stochasticity is a scalar because each observation value is a single number per step. The Gaussian form is chosen for tractability, since it makes the posterior available in closed form; it is not a claim about the distribution of tokens. In the terminology of recurrent memories, $k_t$ is the key, $x_t$ the value, and the posterior mean of $Z_t$ is the memory that maps keys to values.

Given the parameters at each step, the posterior over each latent column is Gaussian and is computed by the Kalman filter \citep{kalman1960new, bishop2006pattern}. The columns share their covariance $U_t$, since they are observed along the same direction with the same noise. The prediction step propagates the covariance through the transition and adds volatility, and the observation step weights the new observation against its prediction:
\begin{equation}
 P_t=A_tU_{t-1}A_t^\top+\operatorname{diag}(v_t),
 \qquad
 g_t=\gamma_t\,P_tk_t,
 \qquad
 \gamma_t=\frac{1}{s_t+k_t^\top P_tk_t},
 \qquad
 U_t=P_t-g_tk_t^\top P_t.
 \label{eq:exact_filter}
\end{equation}
The posterior mean of column $j$ is $m_{t,j}=A_tm_{t-1,j}+g_t\bigl(x_{t,j}-k_t^\top A_tm_{t-1,j}\bigr)$. Because the gain is shared across columns, stacking them into $M_t\in\mathbb{R}^{N\times P}$, the memory, gives
\begin{equation}
 M_t=A_tM_{t-1}+g_t\bigl(x_t^\top-k_t^\top A_tM_{t-1}\bigr)
 =\bigl(I-\gamma_t P_tk_tk_t^\top\bigr)A_tM_{t-1}+\gamma_t P_tk_t\,x_t^\top,
 \label{eq:memory_update}
\end{equation}
where the second form makes the erase-and-write structure explicit: the memory's component along $k_t$ is removed and the new observation written in its place, both scaled by the gain. The readout $o_t=M_t^\top q_t$ is the posterior predictive mean of the observation along a query direction $q_t$, computed from the token. The gain weights the difference between the observation and its prediction: larger predicted uncertainty writes more strongly, and larger observation noise writes less. Because each observation dimension is scalar, the denominator $1/\gamma_t$ is scalar and no matrix inverse is required.
In a neural sequence model, all parameters of \Eqref{eq:latent_model} are supplied by the network: learned projections of the token representation $h_t$ produce the observation, key, and query, and separate projections produce the transition and both noise variances (\Appref{app:rates}). That the variances depend on the input is what inference over them would produce: a learner that treats $v$ and $s$ as unknown maintains a posterior, and the rates entering its state update at time $t$ are posterior-weighted and therefore functions of $x_{1:t}$, as particle-filter and grid-based treatments compute explicitly \citep{piray2021computational, chen2026categorical}. We instead amortize that dependence, producing the rates directly as learned functions of the input and training them end-to-end through the sequence-prediction loss, which treats them as parameters rather than as generative random variables.

One dependence is deliberately absent. A learner that infers $v$ and $s$ updates its beliefs about them from the predictive likelihood of each observation, so the inferred rates depend on the current state estimate. Here the rates depend on the input alone, which leaves the gain computable without the memory and the two recursions separable.

The obstacle to using this filter directly as a recurrent memory is the covariance. The update of $U_t$ subtracts a rank-one term from a matrix that the prediction step has already made dense, so the exact posterior covariance is a full $N\times N$ matrix costing $O(N^2)$ per token. Recurrent memories are trainable at scale only because their state update can be rewritten as matrix products over blocks of tokens, which lets a GPU process many tokens at once instead of stepping through them one at a time \citep{NEURIPS2024_d13a3eae}. The dense covariance admits no such rewriting, and at $O(N^2)$ per token it is as expensive as the memory update it is meant to inform, so it would dominate training cost.

\subsection{The isotropic case: gated delta rules}
\label{sec:method_gated}

Consider the simplest case of \Eqref{eq:latent_model}: a latent state with no preferred directions, in which volatility is shared across coordinates and the covariance is isotropic, $P_t=\rho_tI$. For a unit key, $P_tk_t=\rho_tk_t$, so both factors in \Eqref{eq:memory_update} collapse onto the key,
\begin{equation}
 g_t=\gamma_t\rho_t\,k_t\;\equiv\;\beta_tk_t,
 \qquad
 \beta_t=\frac{\rho_t}{s_t+\rho_t}\in(0,1),
 \label{eq:isotropic_gain}
\end{equation}
and the memory update becomes $M_t=(I-\beta_tk_tk_t^\top)A_tM_{t-1}+\beta_tk_tx_t^\top$. This is exactly the form of two recent recurrent memories, adopted as the sequence mixer in frontier open-weight models such as Qwen3-Next \citep{qwenteam2025qwen3next} and Kimi Linear, and both derived from the delta rule and its gating rather than from a filter:
\begin{align}
 \text{GDN:}\quad M_t&=\alpha_t\bigl(I-\beta_tk_tk_t^\top\bigr)M_{t-1}+\beta_tk_tx_t^\top, \label{eq:gdn}\\
 \text{KDA:}\quad M_t&=\bigl(I-\beta_tk_tk_t^\top\bigr)\operatorname{diag}(\alpha_t)M_{t-1}+\beta_tk_tx_t^\top, \label{eq:kda}
\end{align}
differing in whether the decay $\alpha_t$ is a scalar or a vector along the state dimension \citep{ICLR2025_4904fad1, kimiteam2025kimilinearexpressiveefficient}. The correspondence is one of form, and it says what the two components of these models do: the rank-one erase occupies the position of the filter's covariance correction, and the learned write strength $\beta_t$ that of its gain. GDN-2 \citep{hatamizadeh2026gateddeltanet2decouplingerase} uses separate erase and write vectors and so falls outside this correspondence, but its write is likewise scaled by a quantity that does not vary along the key axis.

In each case the write is fixed to the direction of the key, and its strength is one number per head: a covariance with a single scale produces a gain with a single magnitude. Voltic keeps the covariance anisotropic, so that the gain varies along the key axis according to where uncertainty has accumulated. We develop two approximations that preserve this structure while keeping the per-token uncertainty work well below that of the memory update, so it remains a small addition to a computation that still runs in block form: a diagonal covariance (\Secref{sec:method_diagonal}) and a quasi-diagonal one that retains correlations within short blocks of tokens (\Secref{sec:method_quasidiagonal}).

\subsection{Voltic-D: Diagonal uncertainty}
\label{sec:method_diagonal}

Both variants of Voltic are assumed-density filters \citep{opper1999bayesian, minka2001family}: after each observation, the posterior is projected onto a tractable Gaussian family by minimizing the KL divergence to the exact posterior. For Gaussians this projection matches moments, so projecting onto diagonal covariances keeps the exact marginal variances and discards only the correlations between coordinates. The variants differ in how often they project.

Volitc-D projects after every token, so the covariance is a diagonal $u_t$ throughout. With coordinate-wise decay $\alpha_t$, the update is
\begin{align}
 p_t&=\alpha_t^{\odot2}\odot u_{t-1}+v_t,
 &d_t&=s_t+\sum_{i=1}^{N}p_{t,i}k_{t,i}^{2}, \label{eq:adf_prediction}\\
 g_t&=(p_t\odot k_t)/d_t,
 &u_t&=p_t-(p_t\odot k_t)^{\odot2}/d_t, \label{eq:adf_gain}
\end{align}
where $\odot$ denotes element-wise multiplication and $d_t=1/\gamma_t$. The gain is the key reweighted by predicted uncertainty: coordinates in which the state is more uncertain receive a larger share of the write. Because the covariance is shared across the $P$ columns of the memory, this costs $O(N)$ work and $N$ stored scalars per head, against the $NP$ scalars of the memory itself.

\subsection{Voltic-Q: Quasi-diagonal uncertainty}
\label{sec:method_quasidiagonal}
Discarding correlations at every token is the coarsest projection that keeps the uncertainty cheap enough to run alongside the memory update. Voltic-Q keeps more of the covariance: within a block of $Q$ tokens the exact recursion runs with no projection, so correlations between coordinates accumulate and are discarded only at the block boundary. Inside a block,
\begin{equation}
 P_t=U_{t-1}+\operatorname{diag}(v_t),\qquad
 g_t=\gamma_t P_tk_t,\qquad
 U_t=P_t-\gamma_t(P_tk_t)(P_tk_t)^\top,
 \label{eq:qadf_update}
\end{equation}
and at the boundary the off-diagonal entries of $U_t$ are set to zero, the same moment-matching projection applied every $Q$ tokens rather than every one, with the diagonal passed to the next block along with the memory. The two variants are endpoints of one family: $Q=1$ recovers \Secref{sec:method_diagonal}, and $Q$ equal to the sequence length would be the exact filter. Because the projection applies to the uncertainty alone, $Q$ controls the covariance approximation without restricting the context length.

Voltic-Q uses fixed decay, $A_t=I$. A coordinate-wise decay inside a block would require rescaling the accumulated terms by cumulative products of the decays, which shrink toward zero over a block and destabilize the gradients; with $A_t=I$ no such rescaling arises.

Within a block the covariance never has to be formed. Each observation adds a diagonal term through the prediction step and subtracts a rank-one term through the correction, so $U_t$ is always a known diagonal minus at most $Q$ rank-one terms, and the gain follows from those terms alone. This costs $O(QN)$ work per token, a factor $Q/P$ of the memory update's $O(NP)$, so the uncertainty remains the cheaper of the two computations as long as $Q$ stays well below $P$ (\Appref{app:qadf_implementation}). The main experiments use $Q=4$; \Appref{app:covariance_blocks} examines the effect of the block length.

\subsection{Efficient implementation}
\label{sec:implementation_optimizations}

The efficiency of the gated linear-attention family rests on one algorithm: updates of the form $M_t=T_tM_{t-1}+w_tx_t^\top$, with $T_t$ diagonal plus rank one, can be evaluated with matrix products over chunks of tokens rather than stepping through them \citep{NEURIPS2024_d13a3eae, yang2024fla}. Voltic inherits it: because the gain does not depend on the memory or the observation, \Eqref{eq:memory_update} has exactly this form, with transition $(I-g_tk_t^\top)A_t$, and we evaluate it with a specialized version of the corresponding FLA operator. The per-coordinate gain costs nothing here: $g_t$ enters where a gated delta rule places a scalar multiple of the key, and the kernel does not care which.

What the filter adds is the uncertainty recursion, which is sequential. Its cost per token is $O(N)$ for Voltic-D and $O(QN)$ for Voltic-Q, against $O(NP)$ for the memory update, so it is smaller by a factor of $P$ and $P/Q$. Note that $Q$ sets the covariance approximation, while the chunk length only organizes the same recurrence. Fused projections, a fused kernel for the noise rates, and reuse of the gain's matrix products between its write and erase roles reduce memory traffic and kernel launches; uncertainty updates use FP32. The surrounding block follows standard practice for recurrent memories, with short causal convolutions, a gated output, and normalization before the projection back to $d_{\mathrm{model}}$. \Appref{app:implementation} details the block, kernels, and gradients; \Appref{app:resources} lists software, hardware, and training time. \Secref{sec:throughput} reports throughput against gated recurrent-attention baselines.
\section{Experiments and Results}
\label{sec:experiments}

Voltic separates changes in stored associations from noise in new observations. We test this design first with controlled memory tasks, where both sources can be specified, then with language modeling and in-context retrieval, where useful information must be recovered from natural or synthetic text. Training-throughput and cached-decoding measurements test computational cost as sequence length and batch size vary, and when generating tokens one at a time.

\subsection{Models compared}
\label{sec:model_variants}

We compare Voltic-D and Voltic-Q with KLA, which provides another filtering-based memory, and with KDA and GDN-2, which adapt memory through learned gates. Mamba-3 supplies state-space baselines in single-input single-output (SISO) and multiple-input multiple-output (MIMO) forms; the latter uses several input and output directions per update. For MQAR only, xLSTM \citep{NEURIPS2024_c2ce2f27} provides another recurrent-memory comparison, with either two matrix-memory (mLSTM) layers or a matrix/scalar-memory (mLSTM/sLSTM) hybrid. A causal softmax Transformer \citep{NIPS2017_3f5ee243} provides a full-attention comparison for quality and throughput. Each benchmark specifies the included models and matched dimensions.

\subsection{Multi-query associative recall}
\label{sec:mqar_setup}

Multi-query associative recall (MQAR) presents key--value pairs and asks the model to predict the value of each queried key \citep{ICLR2024_448fc91f}. Since associations are sampled separately for each sequence, the task isolates the memory update: nothing can be answered from the weights, and performance depends on what the model writes and retains. We use four matched conditions with three observations per key: associations may change during the sequence or stay fixed, and observations may be corrupted or faithful. In our Standard MQAR condition, associations stay fixed and all observations are faithful. This gives a $2\times2$ design that varies volatility and stochasticity while holding the observation and query layout fixed. Keys and values are categorical tokens with no metric, so none of the variants matches the Gaussian model of \Secref{sec:implementation_recurrence}; we use them to test the structural distinction that model encodes rather than its distributional form.

\textbf{Tasks.} Each sequence has $K$ distinct keys, three observation rounds, and one query per key, with length $T=8K$. Keys and values come from disjoint parts of an 8,192-token vocabulary. An observation is two consecutive tokens; a query supplies a key and asks for its value. \emph{Standard MQAR} repeats each unchanged association in all three rounds. In \emph{Volatile MQAR}, each key's hidden value changes independently with probability $0.30$ before each of the last two rounds. The target is the most recent value, so the model must replace those associations while retaining untouched ones. \emph{Stochastic MQAR} holds every hidden value fixed, but replaces each displayed value with an incorrect one with probability $0.25$. A conflicting observation is therefore always noise, and the target is the hidden value, which the model must recover from the observations. \emph{Vol \& Sto MQAR} combines the two: the hidden value changes with probability $0.30$ before each of the last two rounds, and each displayed value is independently corrupted with probability $0.25$. A conflicting observation can now reflect either a change or noise, so the latest observation is insufficient on its own, and neither source is observable to the model.

\textbf{Training and evaluation.} Models share two layers, hidden width 128, training data, and 10,000 AdamW updates \citep{loshchilov2018decoupled} at $K=16/64$. We average accuracy at the two memory sizes equally within each run, then report the mean and its standard error (SEM) over five runs. \Appref{app:mqar} gives the full protocol, individual-seed results, and extrapolation tests.

\begin{table}[!htbp]
\centering
\caption{MQAR accuracy (\%), averaging $K=16$ and $K=64$ equally within each run: mean $\pm$ SEM over five runs. Bold/underline mark the top two displayed means, including ties.}
\label{tab:mqar_results}
\small
\setlength{\tabcolsep}{3.5pt}
\begin{tabular}{lrrrr}
\toprule
Model & Standard & Volatile & Stochastic & Vol \& Sto \\
\midrule
\rowcolor{VolticTint}
Voltic-D & $\underline{99.98}\,\pm\,0.02$ & $\underline{99.29}\,\pm\,0.31$ & $\mathbf{88.33}\,\pm\,0.20$ & $69.86\,\pm\,1.63$ \\
\rowcolor{VolticTint}
Voltic-Q & $\mathbf{100.00}\,\pm\,0.00$ & $\mathbf{99.48}\,\pm\,0.23$ & $\underline{87.61}\,\pm\,0.33$ & $\mathbf{72.89}\,\pm\,0.26$ \\
KLA & $84.94\,\pm\,2.14$ & $75.26\,\pm\,3.53$ & $69.44\,\pm\,1.55$ & $47.62\,\pm\,8.10$ \\
KDA & $\mathbf{100.00}\,\pm\,0.00$ & $48.76\,\pm\,19.41$ & $84.08\,\pm\,3.55$ & $\underline{70.88}\,\pm\,0.31$ \\
GDN-2 & $99.85\,\pm\,0.14$ & $70.25\,\pm\,16.43$ & $71.40\,\pm\,16.53$ & $65.12\,\pm\,6.83$ \\
Mamba-3 SISO & $11.67\,\pm\,2.08$ & $5.14\,\pm\,2.00$ & $4.84\,\pm\,0.94$ & $1.55\,\pm\,0.63$ \\
Mamba-3 MIMO & $12.96\,\pm\,1.51$ & $2.65\,\pm\,0.64$ & $26.72\,\pm\,10.25$ & $0.29\,\pm\,0.08$ \\
xLSTM (mLSTM) & $99.95\,\pm\,0.02$ & $86.39\,\pm\,2.73$ & $84.52\,\pm\,1.52$ & $66.78\,\pm\,1.70$ \\
xLSTM (hybrid) & $99.93\,\pm\,0.03$ & $87.11\,\pm\,1.04$ & $85.64\,\pm\,0.08$ & $64.97\,\pm\,2.64$ \\
\midrule
Transformer & $\underline{99.98}\,\pm\,0.00$ & $2.96\,\pm\,0.01$ & $3.39\,\pm\,0.02$ & $3.16\,\pm\,0.01$ \\
\bottomrule
\end{tabular}
\end{table}

\phantomsection\label{sec:mqar_results}\textbf{Results.} Standard MQAR is saturated: seven models are above 99.8\%, so it does not separate them (\Tabref{tab:mqar_results}). Varying either source does, and the baselines separate unevenly. KDA, tied for the best Standard accuracy, is among the weakest once values change, and both it and GDN-2 succeed on some seeds and fail on others. xLSTM holds up best among the baselines but trails on all three variants. The Voltic variants are the only models strong across all four tasks, with Voltic-Q leading Volatile MQAR and Vol \& Sto and Voltic-D leading Stochastic MQAR; Voltic-Q also exceeds every baseline on Vol \& Sto at both larger memory sizes (\Tabref{tab:mqar_matched}). \Figref{fig:mqar_seed_accuracy} shows individual-run accuracy. The Transformer saturates Standard MQAR at a lower learning rate, and warm-up training on a one-observation task recovers much of its accuracy on the other three at matched total budget, so its failure under the shared recipe reflects optimization rather than capacity (\Appref{app:mqar_transformer}).

\textbf{Variance ablations.} Fixing $v_t$, $s_t$, or both to $0.5$ during evaluation tests how trained models rely on their noise branches (\Tabref{tab:mqar_variance_ablations}). Fixing $v_t$ has two effects at once: a constant volatility makes the predicted covariance isotropic, so the gain becomes a scalar multiple of the key, and accuracy collapses on every task, including Standard MQAR, where no value changes. This is a direct test of the anisotropy of \Secref{sec:method_gated}, and removing it costs most of the model's accuracy. Fixing $s_t$ instead leaves the covariance anisotropic and removes only the stochasticity response, with the largest effect on Stochastic MQAR.

\subsection{Language modeling}
\label{sec:lm_setup}

Language modeling tests whether Voltic's memory supports prediction in natural text.

\textbf{Data, models, training, and evaluation.} Models share FineWeb-Edu data \citep{NEURIPS2024_370df50c}, a tokenizer, and ordered 2,048-token sequences, respecting document and sequence boundaries. The approximately 45.5M-parameter models share depth, width, and 4,800 AdamW updates over about 2.52 billion input tokens. Voltic, KLA, KDA, GDN-2, and both Mamba-3 variants receive the same base-learning-rate search. Selection and final training use separate seeds. Results use final checkpoints from three fresh seeds, reported as mean $\pm$ sample standard deviation. \Tabref{tab:lm_configuration} gives model dimensions and selected rates; \Appref{app:lm} gives the full recipe. We report zero-shot accuracy with shared prompts through Language Model Evaluation Harness \citep{biderman2026lessonstrenchesreproducibleevaluation}, and record FineWeb-Edu NLL and WikiText-103 perplexity \citep{merity2017pointer} as a convergence check (\Tabref{tab:lm45_likelihood}). The eight-task average reflects reasoning and comprehension tasks; the nine-task average adds LAMBADA's final-word prediction \citep{paperno-etal-2016-lambada}. 

\begin{figure}[!htbp]
\centering
\includegraphics[width=\linewidth]{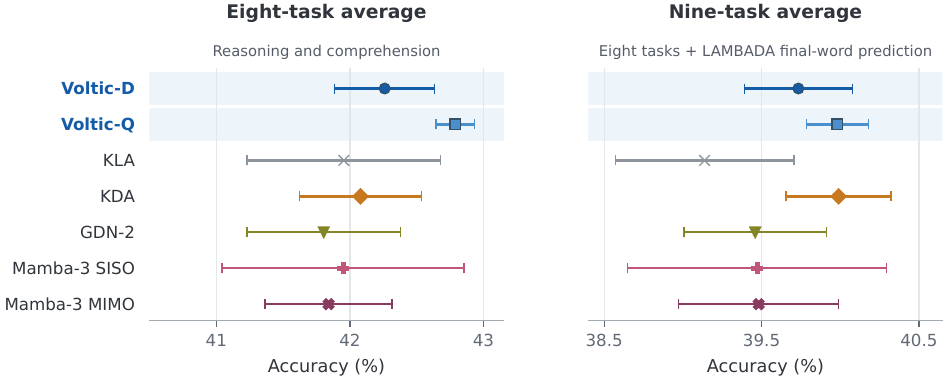}
\caption{45M downstream accuracy after 2.52 billion training tokens: mean $\pm$ sample SD over three seeds. The nine-task average adds LAMBADA to the eight-task average (\Appref{app:lm_evaluation}).}
\label{fig:lm45_results}
\end{figure}

\textbf{Results.} Voltic-Q achieves the highest eight-task average and Voltic-D the second, with the seven recurrent models spanning about one point (\Figref{fig:lm45_results}). Voltic-Q and KDA have nine-task averages within 0.01 percentage points. Voltic-Q has the highest mean on ARC-Challenge \citep{clark2018thinksolvedquestionanswering}, BoolQ \citep{clark-etal-2019-boolq}, and OpenBookQA \citep{mihaylov-etal-2018-suit}. \Tabref{tab:lm45_tasks} reports individual task scores.

\subsection{In-context retrieval}
\label{sec:retrieval}

In-context retrieval tests whether a trained language model can generate answers from details supplied in its prompt. For a recurrent model this is a direct test of the memory: the answer must have been written to a fixed-size state and survived everything written after it. We compare the recurrent language models from \Secref{sec:lm_setup}, without retrieval fine-tuning or prompt tuning. The full-attention RULER comparison appears in \Figref{fig:retrieval45_contexts}.

\textbf{Tasks and evaluation.} Four RULER tasks \citep{hsieh2024ruler} test key--value retrieval at 1K--8K tokens, exceeding the training length at 4K/8K without adaptation. Six BASED tasks \citep{pmlr-v235-arora24a} test natural-text field extraction and question answering within 2K tokens.

Selected LongBench tasks \citep{bai-etal-2024-longbench} test question answering, summarization, and classification; BABILong tasks \citep{NEURIPS2024_c0d62e70} require one to three supporting facts. These short-context subsets complement RULER, without covering either full benchmark.

Models share examples, prompts, and greedy decoding. Groups average task/context scores equally within each seed, then report mean and SD across seeds. \Appref{app:retrieval} gives the full protocol.

\begin{table}[!htbp]
\centering
\caption{45M in-context scores (0--100) for recurrent models: mean $\pm$ sample SD over three seeds. LongBench combines F1, ROUGE-L \citep{lin-2004-rouge}, and classification scores; other groups use answer-matching accuracy. Bold/underline mark the top two means.}
\label{tab:retrieval_summary}
\small
\setlength{\tabcolsep}{3pt}
\begin{tabular}{lrrrrr}
\toprule
Model & \shortstack{RULER\\All 16} & \shortstack{RULER\\4K/8K} & \shortstack{BASED} & LongBench & BABILong \\
\midrule
\rowcolor{VolticTint}
Voltic-D & $3.12\,\pm\,3.73$ & $0.49\,\pm\,0.29$ & $10.23\,\pm\,0.49$ & $\mathbf{7.04}\,\pm\,0.78$ & $\mathbf{7.26}\,\pm\,1.91$ \\
\rowcolor{VolticTint}
Voltic-Q & $\mathbf{15.15}\,\pm\,8.55$ & $\mathbf{9.53}\,\pm\,8.42$ & $9.29\,\pm\,0.37$ & $4.99\,\pm\,0.38$ & $3.89\,\pm\,1.28$ \\
KLA & $0.07\,\pm\,0.08$ & $0.02\,\pm\,0.01$ & $7.37\,\pm\,0.17$ & $4.52\,\pm\,0.91$ & $3.41\,\pm\,2.06$ \\
KDA & $2.03\,\pm\,0.26$ & $0.33\,\pm\,0.20$ & $\mathbf{11.58}\,\pm\,0.40$ & $\underline{6.46}\,\pm\,0.84$ & $6.70\,\pm\,1.73$ \\
GDN-2 & $\underline{5.05}\,\pm\,2.85$ & $\underline{0.82}\,\pm\,0.61$ & $\underline{10.86}\,\pm\,0.21$ & $6.44\,\pm\,0.35$ & $6.30\,\pm\,1.83$ \\
Mamba-3 SISO & $0.56\,\pm\,0.15$ & $0.07\,\pm\,0.09$ & $9.47\,\pm\,0.32$ & $4.92\,\pm\,1.04$ & $6.19\,\pm\,1.53$ \\
Mamba-3 MIMO & $1.37\,\pm\,1.13$ & $0.46\,\pm\,0.45$ & $9.83\,\pm\,0.49$ & $5.99\,\pm\,0.81$ & $\underline{7.15}\,\pm\,0.61$ \\
\bottomrule
\end{tabular}
\end{table}

\textbf{Results.} Voltic variants have the highest mean on four of the five groups: both RULER groups, LongBench, and BABILong (\Tabref{tab:retrieval_summary}). Voltic-Q leads the average over all sixteen RULER conditions by 10.10 points over GDN-2, and leads the 4K/8K group in all three training seeds; its largest advantage is on S1, which asks for a numeric value in repeated filler text, while long identifiers and multiple key--value pairs remain more difficult. Voltic-D has the highest mean on the LongBench tasks and a small lead over Mamba-3 MIMO on BABILong, smaller than the variation across seeds. The two variants differ in both covariance structure and decay. Voltic-D's learned decay removes content that is never queried again, which plausibly accounts for its advantage where many passages compete. Voltic-Q's block covariance and fixed decay instead favor preserving a single stored value, since nothing decays and the block covariance suppresses repeated writes along directions already used. We do not separate the two factors here.

\subsection{Throughput}
\label{sec:throughput}

Voltic computes an uncertainty state on top of the memory all recurrent models carry. This computation is much smaller than the memory update, but we evaluate it sequentially rather than as a parallel scan \citep{shaj2026kalman}, so the question is what it costs in practice. We measure training throughput in two sweeps, varying sequence length at a fixed token budget and batch size at a fixed length, then measure single-token decoding latency, the regime in which recurrent models are deployed.

\begin{figure}[!htbp]
\centering
\includegraphics[width=0.8\linewidth]{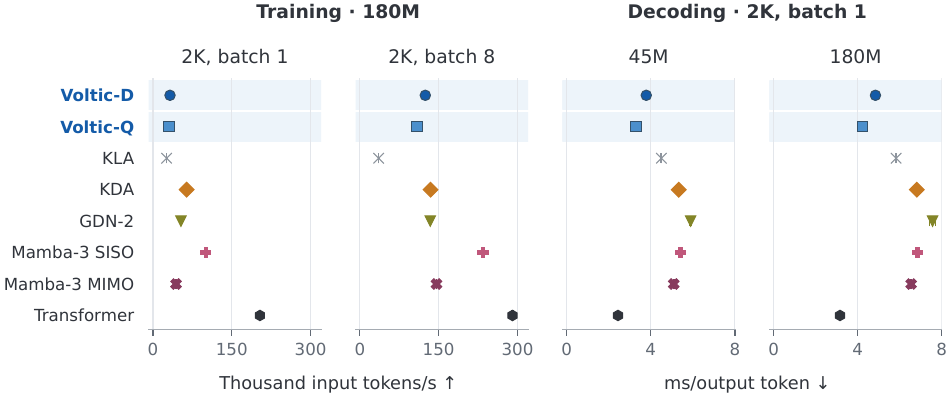}
\caption{Training throughput (180M) and cached decoding latency (45M/180M) at 2K context on one H200 NVL: mean $\pm$ SD over three timing processes. Decoding excludes prompt processing.}
\label{fig:throughput_main}
\end{figure}

\textbf{Models and measurement.} We benchmark the seven recurrent language-model architectures and the Transformer on one physical H200 NVL, using shared inputs and matched 45M/180M configurations distinct from the trained language models. Training timing covers forward, loss, and backward, excluding optimizer updates, data loading, compilation, and warmup. \Appref{app:throughput} specifies dimensions, precision, execution settings, and repetitions.

\textbf{Results.} We hold 8,192 input tokens per operation, increasing length from 1K to 8K as batch size falls from eight to one (\Figref{fig:throughput_sequence}). The Transformer is fastest, followed by Mamba-3 SISO. Voltic-D and Voltic-Q exceed the updated KLA training backend at all these settings except 45M, 8K, batch one, where KLA is faster. This design also reduces the number of independent sequences as length grows. At 180M, the change from 1K to 8K reduces Voltic-D throughput by $65.4\%$, compared with $3.7\%$ for KDA. Holding sequence length fixed separates the batch-size effect.

Next, we test batches of 1, 2, 4, and 8; \Figref{fig:throughput_main} shows the 180M endpoints at 2K tokens. Larger batches substantially improve Voltic's throughput. At 180M, 2K, and batch eight, Voltic-D reaches $92.6\%$ of KDA and $93.0\%$ of GDN-2 throughput. It is the fastest Voltic variant at this setting and reaches 3.5 times the throughput of KLA's updated training backend. At 180M and 8K tokens, increasing the batch from two to four raises Voltic-D throughput by $67.6\%$, compared with $5.2\%$ for KDA. The uncertainty recursion processes tokens in order within a head, while independent sequences provide parallel work. 

\textbf{Cached autoregressive decoding.} Training throughput does not determine decoding latency, which is the regime in which recurrent models are deployed. At batch one, timing covers output tokens including greedy selection and excludes prompt processing. Among the recurrent models, Voltic-Q and Voltic-D have the lowest latency at both scales and all prompt lengths, with one exception at 8K where KLA is slightly faster than Voltic-D. The Transformer is faster still at these lengths, though its per-token cost grows with context while the recurrent models' does not. \Figref{fig:throughput_main} shows 2K prompts; \Appref{app:throughput_decoding} gives the timing protocol and cache checks.

\FloatBarrier
\section{Conclusion}\label{sec:conclusion}
Voltic derives recurrent writes from separate volatility and stochasticity while retaining efficient delta-rule computation. Its two variants lead the three MQAR tasks that vary these sources, four of five in-context retrieval groups, and the eight-task language-model average. Three limitations remain. Quality evaluation is limited to 45M models. Throughput depends on batch parallelism, and Voltic falls furthest behind gated baselines at long sequences with small batches. The two variants differ in both covariance structure and decay, so their complementary strengths are not attributed to one mechanism. Larger-scale validation remains future work. Deriving the write from an uncertainty recursion nonetheless makes recurrent memory more responsive to change, at a cost that stays close to gated baselines where batch parallelism is available.

\label{sec:main_end}

\clearpage
\subsection*{AI use statement}
The authors used OpenAI's Codex assistant with GPT-6 Astra \citep{openai2026codex,openai2026gpt6astra} for drafting assistance and language editing, including revisions to improve the manuscript's grammar, clarity, and organization. The assistant also helped find and verify references, implement code and correctness tests, execute experiments across multiple GPUs, aggregate results, and prepare figures. The authors are responsible for the final text, claims, and artifacts.

\subsection*{Ethics statement}
This work studies recurrent sequence models using synthetic memory tasks and existing text datasets; it does not involve new human-subject data collection. We use public corpora and benchmarks under their applicable terms. Filtering educational web text does not eliminate biases, personal information, or harmful content, and the reported benchmarks do not establish suitability for deployment in sensitive settings. The work concerns memory and computational efficiency rather than deployment of a user-facing language model.

\subsection*{Reproducibility statement}
\Appref{app:filtering} derives the updates, and \Appref{app:implementation} specifies their optimized evaluation and gradients. \Appref{app:resources} records software versions, hardware, and approximate training time. The benchmark appendices document data generation and preparation, model dimensions, optimizer searches, seeds, and scoring (\Appref{app:mqar}, \Appref{app:lm}, \Appref{app:retrieval}, and \Appref{app:throughput}). MQAR error bars report standard errors of the mean across training runs; the other benchmarks report sample standard deviations across training runs or timing processes, as specified in their captions. We will release the implementation and experiment configurations upon publication.

\FloatBarrier
\bibliography{references}

\begin{thebibliography}{76}
\providecommand{\natexlab}[1]{#1}
\providecommand{\url}[1]{\texttt{#1}}
\expandafter\ifx\csname urlstyle\endcsname\relax
  \providecommand{\doi}[1]{doi: #1}\else
  \providecommand{\doi}{doi: \begingroup \urlstyle{rm}\Url}\fi

\bibitem[Arora et~al.(2023)Arora, Yang, Eyuboglu, Narayan, Hojel, Trummer, and
  R{\'e}]{arora2023evaporate}
Simran Arora, Brandon Yang, Sabri Eyuboglu, Avanika Narayan, Andrew Hojel,
  Immanuel Trummer, and Christopher R{\'e}.
\newblock Language models enable simple systems for generating structured views
  of heterogeneous data lakes.
\newblock \emph{Proc. VLDB Endow.}, 17\penalty0 (2):\penalty0 92--105, October
  2023.
\newblock ISSN 2150-8097.
\newblock \doi{10.14778/3626292.3626294}.
\newblock URL \url{https://doi.org/10.14778/3626292.3626294}.

\bibitem[Arora et~al.(2024{\natexlab{a}})Arora, Eyuboglu, Timalsina, Johnson,
  Poli, Zou, Rudra, and Re]{ICLR2024_448fc91f}
Simran Arora, Sabri Eyuboglu, Aman Timalsina, Isys Johnson, Michael Poli,
  James~Y Zou, Atri Rudra, and Christopher Re.
\newblock Zoology: Measuring and improving recall in efficient language models.
\newblock In B.~Kim, Y.~Yue, S.~Chaudhuri, K.~Fragkiadaki, M.~Khan, and Y.~Sun
  (eds.), \emph{International Conference on Learning Representations}, volume
  2024, pp.\  15664--15730, 2024{\natexlab{a}}.
\newblock URL
  \url{https://proceedings.iclr.cc/paper_files/paper/2024/file/448fc91f669c15d10364ee01d512cc10-Paper-Conference.pdf}.

\bibitem[Arora et~al.(2024{\natexlab{b}})Arora, Eyuboglu, Zhang, Timalsina,
  Alberti, Zou, Rudra, and Re]{pmlr-v235-arora24a}
Simran Arora, Sabri Eyuboglu, Michael Zhang, Aman Timalsina, Silas Alberti,
  James Zou, Atri Rudra, and Christopher Re.
\newblock Simple linear attention language models balance the recall-throughput
  tradeoff.
\newblock In Ruslan Salakhutdinov, Zico Kolter, Katherine Heller, Adrian
  Weller, Nuria Oliver, Jonathan Scarlett, and Felix Berkenkamp (eds.),
  \emph{Proceedings of the 41st International Conference on Machine Learning},
  volume 235 of \emph{Proceedings of Machine Learning Research}, pp.\
  1763--1840. PMLR, 21--27 Jul 2024{\natexlab{b}}.
\newblock URL \url{https://proceedings.mlr.press/v235/arora24a.html}.

\bibitem[Bai et~al.(2024)Bai, Lv, Zhang, Lyu, Tang, Huang, Du, Liu, Zeng, Hou,
  Dong, Tang, and Li]{bai-etal-2024-longbench}
Yushi Bai, Xin Lv, Jiajie Zhang, Hongchang Lyu, Jiankai Tang, Zhidian Huang,
  Zhengxiao Du, Xiao Liu, Aohan Zeng, Lei Hou, Yuxiao Dong, Jie Tang, and
  Juanzi Li.
\newblock {L}ong{B}ench: A bilingual, multitask benchmark for long context
  understanding.
\newblock In Lun-Wei Ku, Andre Martins, and Vivek Srikumar (eds.),
  \emph{Proceedings of the 62nd Annual Meeting of the Association for
  Computational Linguistics (Volume 1: Long Papers)}, pp.\  3119--3137,
  Bangkok, Thailand, August 2024. Association for Computational Linguistics.
\newblock \doi{10.18653/v1/2024.acl-long.172}.
\newblock URL \url{https://aclanthology.org/2024.acl-long.172/}.

\bibitem[Beck et~al.(2024)Beck, P\"{o}ppel, Spanring, Auer, Prudnikova, Kopp,
  Klambauer, Brandstetter, and Hochreiter]{NEURIPS2024_c2ce2f27}
Maximilian Beck, Korbinian P\"{o}ppel, Markus Spanring, Andreas Auer,
  Oleksandra Prudnikova, Michael Kopp, G\"{u}nter Klambauer, Johannes
  Brandstetter, and Sepp Hochreiter.
\newblock {xLSTM}: Extended long short-term memory.
\newblock In A.~Globerson, L.~Mackey, D.~Belgrave, A.~Fan, U.~Paquet,
  J.~Tomczak, and C.~Zhang (eds.), \emph{Advances in Neural Information
  Processing Systems}, volume~37, pp.\  107547--107603. Curran Associates,
  Inc., 2024.
\newblock \doi{10.52202/079017-3417}.
\newblock URL
  \url{https://proceedings.neurips.cc/paper_files/paper/2024/file/c2ce2f2701c10a2b2f2ea0bfa43cfaa3-Paper-Conference.pdf}.

\bibitem[Becker et~al.(2019)Becker, Pandya, Gebhardt, Zhao, Taylor, and
  Neumann]{becker2019recurrent}
Philipp Becker, Harit Pandya, Gregor Gebhardt, Cheng Zhao, C.~James Taylor, and
  Gerhard Neumann.
\newblock {Recurrent Kalman Networks}: Factorized inference in high-dimensional
  deep feature spaces.
\newblock In Kamalika Chaudhuri and Ruslan Salakhutdinov (eds.),
  \emph{Proceedings of the 36th International Conference on Machine Learning},
  volume~97 of \emph{Proceedings of Machine Learning Research}, pp.\  544--552.
  PMLR, 09--15 Jun 2019.
\newblock URL \url{https://proceedings.mlr.press/v97/becker19a.html}.

\bibitem[Biderman et~al.(2026)Biderman, Schoelkopf, Sutawika, Gao, Tow, Abbasi,
  Aji, Ammanamanchi, Black, Clive, DiPofi, Etxaniz, Fattori, Forde, Foster,
  Hsu, Jaiswal, Lee, Li, Lovering, Muennighoff, Pavlick, Phang, Skowron, Tan,
  Tang, Wang, Winata, Yvon, and
  Zou]{biderman2026lessonstrenchesreproducibleevaluation}
Stella Biderman, Hailey Schoelkopf, Lintang Sutawika, Leo Gao, Jonathan Tow,
  Baber Abbasi, Alham~Fikri Aji, Pawan~Sasanka Ammanamanchi, Sidney Black,
  Jordan Clive, Anthony DiPofi, Julen Etxaniz, Benjamin Fattori, Jessica~Zosa
  Forde, Charles Foster, Jeffrey Hsu, Mimansa Jaiswal, Wilson~Y. Lee, Haonan
  Li, Charles Lovering, Niklas Muennighoff, Ellie Pavlick, Jason Phang, Aviya
  Skowron, Samson Tan, Xiangru Tang, Kevin~A. Wang, Genta~Indra Winata,
  François Yvon, and Andy Zou.
\newblock Lessons from the trenches on reproducible evaluation of language
  models, 2026.
\newblock URL \url{https://arxiv.org/abs/2405.14782}.

\bibitem[Bishop(2006)]{bishop2006pattern}
Christopher~M Bishop.
\newblock \emph{Pattern recognition and machine learning}.
\newblock Springer, 2006.
\newblock ISBN 978-0-387-31073-2.

\bibitem[Bisk et~al.(2020)Bisk, Zellers, Le~bras, Gao, and Choi]{Bisk_2020}
Yonatan Bisk, Rowan Zellers, Ronan Le~bras, Jianfeng Gao, and Yejin Choi.
\newblock {PIQA}: Reasoning about physical commonsense in natural language.
\newblock \emph{Proceedings of the AAAI Conference on Artificial Intelligence},
  34\penalty0 (05):\penalty0 7432--7439, April 2020.
\newblock ISSN 2159-5399.
\newblock \doi{10.1609/aaai.v34i05.6239}.
\newblock URL \url{https://doi.org/10.1609/aaai.v34i05.6239}.

\bibitem[Chen \& Piray(2026)Chen and Piray]{chen2026categorical}
Junxi Chen and Payam Piray.
\newblock Categorical {Bayes} filtering for computational phenotyping in
  adaptive learning.
\newblock \emph{bioRxiv}, 2026.
\newblock \doi{10.64898/2026.05.14.725268}.
\newblock URL
  \url{https://www.biorxiv.org/content/early/2026/05/18/2026.05.14.725268}.

\bibitem[Clark et~al.(2019)Clark, Lee, Chang, Kwiatkowski, Collins, and
  Toutanova]{clark-etal-2019-boolq}
Christopher Clark, Kenton Lee, Ming-Wei Chang, Tom Kwiatkowski, Michael
  Collins, and Kristina Toutanova.
\newblock {B}ool{Q}: Exploring the surprising difficulty of natural yes/no
  questions.
\newblock In Jill Burstein, Christy Doran, and Thamar Solorio (eds.),
  \emph{Proceedings of the 2019 Conference of the North {A}merican Chapter of
  the Association for Computational Linguistics: Human Language Technologies,
  Volume 1 (Long and Short Papers)}, pp.\  2924--2936, Minneapolis, Minnesota,
  June 2019. Association for Computational Linguistics.
\newblock \doi{10.18653/v1/N19-1300}.
\newblock URL \url{https://aclanthology.org/N19-1300/}.

\bibitem[Clark et~al.(2018)Clark, Cowhey, Etzioni, Khot, Sabharwal, Schoenick,
  and Tafjord]{clark2018thinksolvedquestionanswering}
Peter Clark, Isaac Cowhey, Oren Etzioni, Tushar Khot, Ashish Sabharwal, Carissa
  Schoenick, and Oyvind Tafjord.
\newblock Think you have solved question answering? try {ARC}, the {AI2}
  reasoning challenge, 2018.
\newblock URL \url{https://arxiv.org/abs/1803.05457}.

\bibitem[Dao(2023)]{dao2023causalconv}
Tri Dao.
\newblock {causal-conv1d}: Causal depthwise convolution in {CUDA}.
\newblock Software, 2023.
\newblock URL \url{https://github.com/Dao-AILab/causal-conv1d}.

\bibitem[Dao \& Gu(2023)Dao and Gu]{dao2023mambasoftware}
Tri Dao and Albert Gu.
\newblock {Mamba}: State-space model software.
\newblock Software, 2023.
\newblock URL \url{https://github.com/state-spaces/mamba}.

\bibitem[Dasigi et~al.(2021)Dasigi, Lo, Beltagy, Cohan, Smith, and
  Gardner]{dasigi-etal-2021-dataset}
Pradeep Dasigi, Kyle Lo, Iz~Beltagy, Arman Cohan, Noah~A. Smith, and Matt
  Gardner.
\newblock A dataset of information-seeking questions and answers anchored in
  research papers.
\newblock In Kristina Toutanova, Anna Rumshisky, Luke Zettlemoyer, Dilek
  Hakkani-Tur, Iz~Beltagy, Steven Bethard, Ryan Cotterell, Tanmoy Chakraborty,
  and Yichao Zhou (eds.), \emph{Proceedings of the 2021 Conference of the North
  American Chapter of the Association for Computational Linguistics: Human
  Language Technologies}, pp.\  4599--4610, Online, June 2021. Association for
  Computational Linguistics.
\newblock \doi{10.18653/v1/2021.naacl-main.365}.
\newblock URL \url{https://aclanthology.org/2021.naacl-main.365/}.

\bibitem[Dayan et~al.(2000)Dayan, Kakade, and Montague]{dayan2000learning}
Peter Dayan, Sham Kakade, and P.~Read Montague.
\newblock Learning and selective attention.
\newblock \emph{Nature Neuroscience}, 3\penalty0 (11):\penalty0 1218--1223,
  2000.
\newblock \doi{10.1038/81504}.
\newblock URL \url{https://doi.org/10.1038/81504}.

\bibitem[Doucet et~al.(2000)Doucet, Freitas, Murphy, and
  Russell]{doucet2000rao}
Arnaud Doucet, Nando~de Freitas, Kevin~P. Murphy, and Stuart~J. Russell.
\newblock Rao-blackwellised particle filtering for dynamic bayesian networks.
\newblock In \emph{Proceedings of the 16th Conference on Uncertainty in
  Artificial Intelligence}, UAI '00, pp.\  176--183, San Francisco, CA, USA,
  2000. Morgan Kaufmann Publishers Inc.
\newblock ISBN 1558607099.

\bibitem[Dua et~al.(2019)Dua, Wang, Dasigi, Stanovsky, Singh, and
  Gardner]{dua-etal-2019-drop}
Dheeru Dua, Yizhong Wang, Pradeep Dasigi, Gabriel Stanovsky, Sameer Singh, and
  Matt Gardner.
\newblock {DROP}: A reading comprehension benchmark requiring discrete
  reasoning over paragraphs.
\newblock In Jill Burstein, Christy Doran, and Thamar Solorio (eds.),
  \emph{Proceedings of the 2019 Conference of the North {A}merican Chapter of
  the Association for Computational Linguistics: Human Language Technologies,
  Volume 1 (Long and Short Papers)}, pp.\  2368--2378, Minneapolis, Minnesota,
  June 2019. Association for Computational Linguistics.
\newblock \doi{10.18653/v1/N19-1246}.
\newblock URL \url{https://aclanthology.org/N19-1246/}.

\bibitem[Elfwing et~al.(2018)Elfwing, Uchibe, and Doya]{Elfwing_2018}
Stefan Elfwing, Eiji Uchibe, and Kenji Doya.
\newblock Sigmoid-weighted linear units for neural network function
  approximation in reinforcement learning.
\newblock \emph{Neural Networks}, 107:\penalty0 3--11, November 2018.
\newblock ISSN 0893-6080.
\newblock \doi{10.1016/j.neunet.2017.12.012}.
\newblock URL \url{https://doi.org/10.1016/j.neunet.2017.12.012}.

\bibitem[Fang \& Piray(2026)Fang and Piray]{fang2026inferring}
Xiaotong Fang and Payam Piray.
\newblock Inferring the causes of noise from binary outcomes: A normative
  theory of learning under uncertainty.
\newblock \emph{Psychological Review}, 2026.
\newblock \doi{10.1037/rev0000638}.
\newblock URL \url{https://doi.org/10.1037/rev0000638}.

\bibitem[Gliwa et~al.(2019)Gliwa, Mochol, Biesek, and
  Wawer]{gliwa-etal-2019-samsum}
Bogdan Gliwa, Iwona Mochol, Maciej Biesek, and Aleksander Wawer.
\newblock {SAMS}um corpus: A human-annotated dialogue dataset for abstractive
  summarization.
\newblock In Lu~Wang, Jackie Chi~Kit Cheung, Giuseppe Carenini, and Fei Liu
  (eds.), \emph{Proceedings of the 2nd Workshop on New Frontiers in
  Summarization}, pp.\  70--79, Hong Kong, China, November 2019. Association
  for Computational Linguistics.
\newblock \doi{10.18653/v1/D19-5409}.
\newblock URL \url{https://aclanthology.org/D19-5409/}.

\bibitem[Gu \& Dao(2024)Gu and Dao]{gu2024mamba}
Albert Gu and Tri Dao.
\newblock {Mamba}: Linear-time sequence modeling with selective state spaces.
\newblock In \emph{First Conference on Language Modeling}, 2024.
\newblock URL \url{https://openreview.net/forum?id=tEYskw1VY2}.

\bibitem[Gugger et~al.(2022)Gugger, Debut, Wolf, Schmid, Mueller, Mangrulkar,
  Sun, and Bossan]{accelerate}
Sylvain Gugger, Lysandre Debut, Thomas Wolf, Philipp Schmid, Zachary Mueller,
  Sourab Mangrulkar, Marc Sun, and Benjamin Bossan.
\newblock Accelerate: Training and inference at scale made simple, efficient
  and adaptable.
\newblock \url{https://github.com/huggingface/accelerate}, 2022.

\bibitem[Hafner et~al.(2019)Hafner, Lillicrap, Fischer, Villegas, Ha, Lee, and
  Davidson]{hafner2019learning}
Danijar Hafner, Timothy Lillicrap, Ian Fischer, Ruben Villegas, David Ha,
  Honglak Lee, and James Davidson.
\newblock Learning latent dynamics for planning from pixels.
\newblock In Kamalika Chaudhuri and Ruslan Salakhutdinov (eds.),
  \emph{Proceedings of the 36th International Conference on Machine Learning},
  volume~97 of \emph{Proceedings of Machine Learning Research}, pp.\
  2555--2565. PMLR, 09--15 Jun 2019.
\newblock URL \url{https://proceedings.mlr.press/v97/hafner19a.html}.

\bibitem[Harris et~al.(2020)Harris, Millman, van~der Walt, Gommers, Virtanen,
  Cournapeau, Wieser, Taylor, Berg, Smith, Kern, Picus, Hoyer, van Kerkwijk,
  Brett, Haldane, del Río, Wiebe, Peterson, Gérard-Marchant, Sheppard, Reddy,
  Weckesser, Abbasi, Gohlke, and Oliphant]{Harris_2020}
Charles~R. Harris, K.~Jarrod Millman, Stéfan~J. van~der Walt, Ralf Gommers,
  Pauli Virtanen, David Cournapeau, Eric Wieser, Julian Taylor, Sebastian Berg,
  Nathaniel~J. Smith, Robert Kern, Matti Picus, Stephan Hoyer, Marten~H. van
  Kerkwijk, Matthew Brett, Allan Haldane, Jaime~Fernández del Río, Mark
  Wiebe, Pearu Peterson, Pierre Gérard-Marchant, Kevin Sheppard, Tyler Reddy,
  Warren Weckesser, Hameer Abbasi, Christoph Gohlke, and Travis~E. Oliphant.
\newblock Array programming with {NumPy}.
\newblock \emph{Nature}, 585\penalty0 (7825):\penalty0 357--362, September
  2020.
\newblock ISSN 1476-4687.
\newblock \doi{10.1038/s41586-020-2649-2}.
\newblock URL \url{https://doi.org/10.1038/s41586-020-2649-2}.

\bibitem[Hatamizadeh et~al.(2026)Hatamizadeh, Choi, and
  Kautz]{hatamizadeh2026gateddeltanet2decouplingerase}
Ali Hatamizadeh, Yejin Choi, and Jan Kautz.
\newblock {Gated DeltaNet-2}: Decoupling erase and write in linear attention,
  2026.
\newblock URL \url{https://arxiv.org/abs/2605.22791}.

\bibitem[Hsieh et~al.(2024)Hsieh, Sun, Kriman, Acharya, Rekesh, Jia, and
  Ginsburg]{hsieh2024ruler}
Cheng-Ping Hsieh, Simeng Sun, Samuel Kriman, Shantanu Acharya, Dima Rekesh, Fei
  Jia, and Boris Ginsburg.
\newblock {RULER}: What{\textquoteright}s the real context size of your
  long-context language models?
\newblock In \emph{First Conference on Language Modeling}, 2024.
\newblock URL \url{https://openreview.net/forum?id=kIoBbc76Sy}.

\bibitem[Joshi et~al.(2017)Joshi, Choi, Weld, and
  Zettlemoyer]{joshi-etal-2017-triviaqa}
Mandar Joshi, Eunsol Choi, Daniel Weld, and Luke Zettlemoyer.
\newblock {T}rivia{QA}: A large scale distantly supervised challenge dataset
  for reading comprehension.
\newblock In Regina Barzilay and Min-Yen Kan (eds.), \emph{Proceedings of the
  55th Annual Meeting of the Association for Computational Linguistics (Volume
  1: Long Papers)}, pp.\  1601--1611, Vancouver, Canada, July 2017. Association
  for Computational Linguistics.
\newblock \doi{10.18653/v1/P17-1147}.
\newblock URL \url{https://aclanthology.org/P17-1147/}.

\bibitem[Kalman(1960)]{kalman1960new}
R.~E. Kalman.
\newblock A new approach to linear filtering and prediction problems.
\newblock \emph{Journal of Basic Engineering}, 82\penalty0 (1):\penalty0
  35--45, 03 1960.
\newblock ISSN 0021-9223.
\newblock \doi{10.1115/1.3662552}.
\newblock URL \url{https://doi.org/10.1115/1.3662552}.

\bibitem[Katharopoulos et~al.(2020)Katharopoulos, Vyas, Pappas, and
  Fleuret]{katharopoulos2020transformers}
Angelos Katharopoulos, Apoorv Vyas, Nikolaos Pappas, and Fran{\c{c}}ois
  Fleuret.
\newblock Transformers are {RNN}s: Fast autoregressive transformers with linear
  attention.
\newblock In Hal~Daumé III and Aarti Singh (eds.), \emph{Proceedings of the
  37th International Conference on Machine Learning}, volume 119 of
  \emph{Proceedings of Machine Learning Research}, pp.\  5156--5165. PMLR,
  13--18 Jul 2020.
\newblock URL \url{https://proceedings.mlr.press/v119/katharopoulos20a.html}.

\bibitem[{Kimi Team} et~al.(2025){Kimi Team}, Zhang, Lin, Yao, Hu, Meng, Liu,
  Men, Yang, Li, Li, Lu, Liu, Chen, Xu, Yu, Wang, Fan, Zhong, Yuan, Zhang,
  Zhang, Liu, Wang, Fang, He, Liu, Li, Su, Qiu, Pang, Yan, Jiang, Huang, Yin,
  You, Wei, Wang, Hong, Chen, Chen, Wang, Zheng, Wang, Liu, Dong, Zhang, Pan,
  Wu, Wu, Guan, Tao, Fu, Xu, Wang, Lai, Wu, Zhou, Yang, and
  Du]{kimiteam2025kimilinearexpressiveefficient}
{Kimi Team}, Yu~Zhang, Zongyu Lin, Xingcheng Yao, Jiaxi Hu, Fanqing Meng,
  Chengyin Liu, Xin Men, Songlin Yang, Zhiyuan Li, Wentao Li, Enzhe Lu, Weizhou
  Liu, Yanru Chen, Weixin Xu, Longhui Yu, Yejie Wang, Yu~Fan, Longguang Zhong,
  Enming Yuan, Dehao Zhang, Yizhi Zhang, T.~Y. Liu, Haiming Wang, Shengjun
  Fang, Weiran He, Shaowei Liu, Yiwei Li, Jianlin Su, Jiezhong Qiu, Bo~Pang,
  Junjie Yan, Zhejun Jiang, Weixiao Huang, Bohong Yin, Jiacheng You, Chu Wei,
  Zhengtao Wang, Chao Hong, Yutian Chen, Guanduo Chen, Yucheng Wang, Huabin
  Zheng, Feng Wang, Yibo Liu, Mengnan Dong, Zheng Zhang, Siyuan Pan, Wenhao Wu,
  Yuhao Wu, Longyu Guan, Jiawen Tao, Guohong Fu, Xinran Xu, Yuzhi Wang, Guokun
  Lai, Yuxin Wu, Xinyu Zhou, Zhilin Yang, and Yulun Du.
\newblock {Kimi Linear}: An expressive, efficient attention architecture, 2025.
\newblock URL \url{https://arxiv.org/abs/2510.26692}.

\bibitem[Krishnan et~al.(2015)Krishnan, Shalit, and Sontag]{krishnan2015deep}
Rahul~G. Krishnan, Uri Shalit, and David Sontag.
\newblock {Deep Kalman Filters}, 2015.
\newblock URL \url{https://arxiv.org/abs/1511.05121}.

\bibitem[Kuratov et~al.(2024)Kuratov, Bulatov, Anokhin, Rodkin, Sorokin,
  Sorokin, and Burtsev]{NEURIPS2024_c0d62e70}
Yuri Kuratov, Aydar Bulatov, Petr Anokhin, Ivan Rodkin, Dmitry Sorokin, Artyom
  Sorokin, and Mikhail Burtsev.
\newblock {BABILong}: Testing the limits of {LLM}s with long context
  reasoning-in-a-haystack.
\newblock In A.~Globerson, L.~Mackey, D.~Belgrave, A.~Fan, U.~Paquet,
  J.~Tomczak, and C.~Zhang (eds.), \emph{Advances in Neural Information
  Processing Systems}, volume~37, pp.\  106519--106554. Curran Associates,
  Inc., 2024.
\newblock \doi{10.52202/079017-3381}.
\newblock URL
  \url{https://proceedings.neurips.cc/paper_files/paper/2024/file/c0d62e70dbc659cc9bd44cbcf1cb652f-Paper-Datasets_and_Benchmarks_Track.pdf}.

\bibitem[Kwiatkowski et~al.(2019)Kwiatkowski, Palomaki, Redfield, Collins,
  Parikh, Alberti, Epstein, Polosukhin, Devlin, Lee, Toutanova, Jones, Kelcey,
  Chang, Dai, Uszkoreit, Le, and Petrov]{kwiatkowski-etal-2019-natural}
Tom Kwiatkowski, Jennimaria Palomaki, Olivia Redfield, Michael Collins, Ankur
  Parikh, Chris Alberti, Danielle Epstein, Illia Polosukhin, Jacob Devlin,
  Kenton Lee, Kristina Toutanova, Llion Jones, Matthew Kelcey, Ming-Wei Chang,
  Andrew~M. Dai, Jakob Uszkoreit, Quoc Le, and Slav Petrov.
\newblock Natural questions: A benchmark for question answering research.
\newblock \emph{Transactions of the Association for Computational Linguistics},
  7:\penalty0 452--466, 2019.
\newblock \doi{10.1162/tacl_a_00276}.
\newblock URL \url{https://aclanthology.org/Q19-1026/}.

\bibitem[Lahoti et~al.(2026)Lahoti, Li, Chen, Wang, Bick, Kolter, Dao, and
  Gu]{ICLR2026_8abd2043}
Aakash~Sunil Lahoti, Kevin Li, Berlin Chen, Caitlin Wang, Aviv Bick, Zico
  Kolter, Tri Dao, and Albert Gu.
\newblock {Mamba-3}: Improved sequence modeling using state space principles.
\newblock In C.~Vondrick, B.~Hariharan, C.~Raffel, L.~Pinto, D.~Yang, and
  A.~Faust (eds.), \emph{International Conference on Learning Representations},
  volume 2026, pp.\  86173--86202, 2026.
\newblock URL
  \url{https://proceedings.iclr.cc/paper_files/paper/2026/file/8abd2043b71a074278d5f687947bff9c-Paper-Conference.pdf}.

\bibitem[Lhoest et~al.(2021)Lhoest, Villanova~del Moral, Jernite, Thakur, von
  Platen, Patil, Chaumond, Drame, Plu, Tunstall, Davison, {\v{S}}a{\v{s}}ko,
  Chhablani, Malik, Brandeis, Le~Scao, Sanh, Xu, Patry, McMillan-Major, Schmid,
  Gugger, Delangue, Matussi{\`e}re, Debut, Bekman, Cistac, Goehringer, Mustar,
  Lagunas, Rush, and Wolf]{lhoest-etal-2021-datasets}
Quentin Lhoest, Albert Villanova~del Moral, Yacine Jernite, Abhishek Thakur,
  Patrick von Platen, Suraj Patil, Julien Chaumond, Mariama Drame, Julien Plu,
  Lewis Tunstall, Joe Davison, Mario {\v{S}}a{\v{s}}ko, Gunjan Chhablani,
  Bhavitvya Malik, Simon Brandeis, Teven Le~Scao, Victor Sanh, Canwen Xu,
  Nicolas Patry, Angelina McMillan-Major, Philipp Schmid, Sylvain Gugger,
  Cl{\'e}ment Delangue, Th{\'e}o Matussi{\`e}re, Lysandre Debut, Stas Bekman,
  Pierric Cistac, Thibault Goehringer, Victor Mustar, Fran{\c{c}}ois Lagunas,
  Alexander Rush, and Thomas Wolf.
\newblock Datasets: A community library for natural language processing.
\newblock In Heike Adel and Shuming Shi (eds.), \emph{Proceedings of the 2021
  Conference on Empirical Methods in Natural Language Processing: System
  Demonstrations}, pp.\  175--184, Online and Punta Cana, Dominican Republic,
  November 2021. Association for Computational Linguistics.
\newblock \doi{10.18653/v1/2021.emnlp-demo.21}.
\newblock URL \url{https://aclanthology.org/2021.emnlp-demo.21/}.

\bibitem[Li \& Roth(2002)Li and Roth]{li-roth-2002-learning}
Xin Li and Dan Roth.
\newblock Learning question classifiers.
\newblock In \emph{{COLING} 2002: The 19th International Conference on
  Computational Linguistics}, 2002.
\newblock URL \url{https://aclanthology.org/C02-1150/}.

\bibitem[Lin(2004)]{lin-2004-rouge}
Chin-Yew Lin.
\newblock {ROUGE}: A package for automatic evaluation of summaries.
\newblock In \emph{Text Summarization Branches Out}, pp.\  74--81, Barcelona,
  Spain, July 2004. Association for Computational Linguistics.
\newblock URL \url{https://aclanthology.org/W04-1013/}.

\bibitem[Liu et~al.(2025)Liu, Wang, Wu, Feng, Stone, and Liu]{liu2025longhorn}
Bo~Liu, Rui Wang, Lemeng Wu, Yihao Feng, Peter Stone, and Qiang Liu.
\newblock {Longhorn}: State space models are amortized online learners.
\newblock In Y.~Yue, A.~Garg, N.~Peng, F.~Sha, and R.~Yu (eds.),
  \emph{International Conference on Learning Representations}, volume 2025,
  pp.\  95419--95434, 2025.
\newblock URL
  \url{https://proceedings.iclr.cc/paper_files/paper/2025/file/ee0e45ff4de76cbfdf07015a7839f339-Paper-Conference.pdf}.

\bibitem[Lockard et~al.(2019)Lockard, Shiralkar, and
  Dong]{lockard-etal-2019-openceres}
Colin Lockard, Prashant Shiralkar, and Xin~Luna Dong.
\newblock {O}pen{C}eres: {W}hen open information extraction meets the
  semi-structured web.
\newblock In Jill Burstein, Christy Doran, and Thamar Solorio (eds.),
  \emph{Proceedings of the 2019 Conference of the North {A}merican Chapter of
  the Association for Computational Linguistics: Human Language Technologies,
  Volume 1 (Long and Short Papers)}, pp.\  3047--3056, Minneapolis, Minnesota,
  June 2019. Association for Computational Linguistics.
\newblock \doi{10.18653/v1/N19-1309}.
\newblock URL \url{https://aclanthology.org/N19-1309/}.

\bibitem[Loshchilov \& Hutter(2019)Loshchilov and
  Hutter]{loshchilov2018decoupled}
Ilya Loshchilov and Frank Hutter.
\newblock Decoupled weight decay regularization.
\newblock In \emph{International Conference on Learning Representations}, 2019.
\newblock URL \url{https://openreview.net/forum?id=Bkg6RiCqY7}.

\bibitem[Merity et~al.(2017)Merity, Xiong, Bradbury, and
  Socher]{merity2017pointer}
Stephen Merity, Caiming Xiong, James Bradbury, and Richard Socher.
\newblock Pointer sentinel mixture models.
\newblock In \emph{International Conference on Learning Representations}, 2017.
\newblock URL \url{https://openreview.net/forum?id=Byj72udxe}.

\bibitem[Mihaylov et~al.(2018)Mihaylov, Clark, Khot, and
  Sabharwal]{mihaylov-etal-2018-suit}
Todor Mihaylov, Peter Clark, Tushar Khot, and Ashish Sabharwal.
\newblock Can a suit of armor conduct electricity? a new dataset for open book
  question answering.
\newblock In Ellen Riloff, David Chiang, Julia Hockenmaier, and Jun{'}ichi
  Tsujii (eds.), \emph{Proceedings of the 2018 Conference on Empirical Methods
  in Natural Language Processing}, pp.\  2381--2391, Brussels, Belgium,
  October-November 2018. Association for Computational Linguistics.
\newblock \doi{10.18653/v1/D18-1260}.
\newblock URL \url{https://aclanthology.org/D18-1260/}.

\bibitem[Minka(2001)]{minka2001family}
Thomas~Peter Minka.
\newblock \emph{A family of algorithms for approximate Bayesian inference}.
\newblock PhD thesis, Massachusetts Institute of Technology, 2001.
\newblock URL \url{https://hdl.handle.net/1721.1/86583}.

\bibitem[Moi \& Patry(2023)Moi and Patry]{moi2023tokenizers}
Anthony Moi and Nicolas Patry.
\newblock {HuggingFace's Tokenizers}.
\newblock Software, 2023.
\newblock URL \url{https://github.com/huggingface/tokenizers}.

\bibitem[{NVIDIA}(2025)]{nvidia2025cuda}
{NVIDIA}.
\newblock \emph{{CUDA C++ Programming Guide}}, 2025.
\newblock URL
  \url{https://docs.nvidia.com/cuda/archive/13.0.0/cuda-c-programming-guide/index.html}.
\newblock Release 13.0.

\bibitem[{OpenAI}(2026{\natexlab{a}})]{openai2026codex}
{OpenAI}.
\newblock {Codex}, 2026{\natexlab{a}}.
\newblock URL \url{https://developers.openai.com/learn/codex}.
\newblock Online documentation. Accessed September 26, 2026.

\bibitem[{OpenAI}(2026{\natexlab{b}})]{openai2026gpt6astra}
{OpenAI}.
\newblock {GPT-6 Astra}, 2026{\natexlab{b}}.
\newblock URL \url{https://developers.openai.com/api/docs/models/gpt-6-astra}.
\newblock Online documentation. Accessed September 26, 2026.

\bibitem[Opper(1999)]{opper1999bayesian}
Manfred Opper.
\newblock A {Bayesian} approach to on-line learning.
\newblock In David Saad (ed.), \emph{On-Line Learning in Neural Networks},
  Publications of the Newton Institute, pp.\  363--378. Cambridge University
  Press, Cambridge, 1999.
\newblock \doi{10.1017/CBO9780511569920.017}.
\newblock URL \url{https://doi.org/10.1017/CBO9780511569920.017}.

\bibitem[Paperno et~al.(2016)Paperno, Kruszewski, Lazaridou, Pham, Bernardi,
  Pezzelle, Baroni, Boleda, and Fern{\'a}ndez]{paperno-etal-2016-lambada}
Denis Paperno, Germ{\'a}n Kruszewski, Angeliki Lazaridou, Ngoc~Quan Pham,
  Raffaella Bernardi, Sandro Pezzelle, Marco Baroni, Gemma Boleda, and Raquel
  Fern{\'a}ndez.
\newblock The {LAMBADA} dataset: Word prediction requiring a broad discourse
  context.
\newblock In Katrin Erk and Noah~A. Smith (eds.), \emph{Proceedings of the 54th
  Annual Meeting of the Association for Computational Linguistics (Volume 1:
  Long Papers)}, pp.\  1525--1534, Berlin, Germany, August 2016. Association
  for Computational Linguistics.
\newblock \doi{10.18653/v1/P16-1144}.
\newblock URL \url{https://aclanthology.org/P16-1144/}.

\bibitem[Paszke et~al.(2019)Paszke, Gross, Massa, Lerer, Bradbury, Chanan,
  Killeen, Lin, Gimelshein, Antiga, Desmaison, Kopf, Yang, DeVito, Raison,
  Tejani, Chilamkurthy, Steiner, Fang, Bai, and Chintala]{NEURIPS2019_bdbca288}
Adam Paszke, Sam Gross, Francisco Massa, Adam Lerer, James Bradbury, Gregory
  Chanan, Trevor Killeen, Zeming Lin, Natalia Gimelshein, Luca Antiga, Alban
  Desmaison, Andreas Kopf, Edward Yang, Zachary DeVito, Martin Raison, Alykhan
  Tejani, Sasank Chilamkurthy, Benoit Steiner, Lu~Fang, Junjie Bai, and Soumith
  Chintala.
\newblock {PyTorch}: An imperative style, high-performance deep learning
  library.
\newblock In H.~Wallach, H.~Larochelle, A.~Beygelzimer, F.~d\textquotesingle
  Alch\'{e}-Buc, E.~Fox, and R.~Garnett (eds.), \emph{Advances in Neural
  Information Processing Systems}, volume~32. Curran Associates, Inc., 2019.
\newblock URL
  \url{https://proceedings.neurips.cc/paper_files/paper/2019/file/bdbca288fee7f92f2bfa9f7012727740-Paper.pdf}.

\bibitem[Penedo et~al.(2024)Penedo, Kydl\'{\i}\v{c}ek, allal, Lozhkov,
  Mitchell, Raffel, Von~Werra, and Wolf]{NEURIPS2024_370df50c}
Guilherme Penedo, Hynek Kydl\'{\i}\v{c}ek, Loubna~Ben allal, Anton Lozhkov,
  Margaret Mitchell, Colin Raffel, Leandro Von~Werra, and Thomas Wolf.
\newblock The {FineWeb} datasets: Decanting the web for the finest text data at
  scale.
\newblock In A.~Globerson, L.~Mackey, D.~Belgrave, A.~Fan, U.~Paquet,
  J.~Tomczak, and C.~Zhang (eds.), \emph{Advances in Neural Information
  Processing Systems}, volume~37, pp.\  30811--30849. Curran Associates, Inc.,
  2024.
\newblock \doi{10.52202/079017-0970}.
\newblock URL
  \url{https://proceedings.neurips.cc/paper_files/paper/2024/file/370df50ccfdf8bde18f8f9c2d9151bda-Paper-Datasets_and_Benchmarks_Track.pdf}.

\bibitem[Peng et~al.(2026)Peng, Chattopadhyay, Zancato, Nunez, Xia, and
  Soatto]{peng2026gated}
Liangzu Peng, Aditya Chattopadhyay, Luca Zancato, Elvis Nunez, Wei Xia, and
  Stefano Soatto.
\newblock {Gated KalmaNet}: A fading memory layer through test-time ridge
  regression.
\newblock In \emph{Proceedings of the IEEE/CVF Conference on Computer Vision
  and Pattern Recognition (CVPR)}, pp.\  20212--20222, June 2026.

\bibitem[Piray \& Daw(2021)Piray and Daw]{piray2021computational}
Payam Piray and Nathaniel~D. Daw.
\newblock A model for learning based on the joint estimation of stochasticity
  and volatility.
\newblock \emph{Nature Communications}, 12\penalty0 (1):\penalty0 6587, Nov
  2021.
\newblock ISSN 2041-1723.
\newblock \doi{10.1038/s41467-021-26731-9}.
\newblock URL \url{https://doi.org/10.1038/s41467-021-26731-9}.

\bibitem[{Qwen Team}(2025)]{qwenteam2025qwen3next}
{Qwen Team}.
\newblock {Qwen3-Next}: Towards ultimate training \& inference efficiency.
\newblock Qwen Blog, September 2025.
\newblock URL
  \url{https://qwen.ai/blog?id=4074cca80393150c248e508aa62983f9cb7d27cd}.

\bibitem[Rajpurkar et~al.(2016)Rajpurkar, Zhang, Lopyrev, and
  Liang]{rajpurkar-etal-2016-squad}
Pranav Rajpurkar, Jian Zhang, Konstantin Lopyrev, and Percy Liang.
\newblock {SQ}u{AD}: 100,000+ questions for machine comprehension of text.
\newblock In Jian Su, Kevin Duh, and Xavier Carreras (eds.), \emph{Proceedings
  of the 2016 Conference on Empirical Methods in Natural Language Processing},
  pp.\  2383--2392, Austin, Texas, November 2016. Association for Computational
  Linguistics.
\newblock \doi{10.18653/v1/D16-1264}.
\newblock URL \url{https://aclanthology.org/D16-1264/}.

\bibitem[Sakaguchi et~al.(2020)Sakaguchi, Le~Bras, Bhagavatula, and
  Choi]{Sakaguchi_2020}
Keisuke Sakaguchi, Ronan Le~Bras, Chandra Bhagavatula, and Yejin Choi.
\newblock {WinoGrande}: An adversarial winograd schema challenge at scale.
\newblock \emph{Proceedings of the AAAI Conference on Artificial Intelligence},
  34\penalty0 (05):\penalty0 8732--8740, April 2020.
\newblock ISSN 2159-5399.
\newblock \doi{10.1609/aaai.v34i05.6399}.
\newblock URL \url{https://doi.org/10.1609/aaai.v34i05.6399}.

\bibitem[Sap et~al.(2019)Sap, Rashkin, Chen, Le~Bras, and
  Choi]{sap-etal-2019-social}
Maarten Sap, Hannah Rashkin, Derek Chen, Ronan Le~Bras, and Yejin Choi.
\newblock Social {IQ}a: Commonsense reasoning about social interactions.
\newblock In Kentaro Inui, Jing Jiang, Vincent Ng, and Xiaojun Wan (eds.),
  \emph{Proceedings of the 2019 Conference on Empirical Methods in Natural
  Language Processing and the 9th International Joint Conference on Natural
  Language Processing (EMNLP-IJCNLP)}, pp.\  4463--4473, Hong Kong, China,
  November 2019. Association for Computational Linguistics.
\newblock \doi{10.18653/v1/D19-1454}.
\newblock URL \url{https://aclanthology.org/D19-1454/}.

\bibitem[S{\"a}rkk{\"a} \& Garc{\'\i}a-Fern{\'a}ndez(2021)S{\"a}rkk{\"a} and
  Garc{\'\i}a-Fern{\'a}ndez]{sarkka2021temporal}
Simo S{\"a}rkk{\"a} and {\'A}ngel~F. Garc{\'\i}a-Fern{\'a}ndez.
\newblock Temporal parallelization of {Bayesian} smoothers.
\newblock \emph{IEEE Transactions on Automatic Control}, 66\penalty0
  (1):\penalty0 299--306, 2021.
\newblock \doi{10.1109/TAC.2020.2976316}.
\newblock URL \url{https://ieeexplore.ieee.org/document/9013038/}.

\bibitem[S{\"a}rkk{\"a} \& Nummenmaa(2009)S{\"a}rkk{\"a} and
  Nummenmaa]{sarkka2009recursive}
Simo S{\"a}rkk{\"a} and Aapo Nummenmaa.
\newblock Recursive noise adaptive {Kalman} filtering by variational {Bayesian}
  approximations.
\newblock \emph{IEEE Transactions on Automatic Control}, 54\penalty0
  (3):\penalty0 596--600, 2009.
\newblock \doi{10.1109/TAC.2008.2008348}.
\newblock URL \url{https://ieeexplore.ieee.org/document/4796261/}.

\bibitem[Schlag et~al.(2021)Schlag, Irie, and Schmidhuber]{schlag2021linear}
Imanol Schlag, Kazuki Irie, and J{\"u}rgen Schmidhuber.
\newblock Linear transformers are secretly fast weight programmers.
\newblock In Marina Meila and Tong Zhang (eds.), \emph{Proceedings of the 38th
  International Conference on Machine Learning}, volume 139 of
  \emph{Proceedings of Machine Learning Research}, pp.\  9355--9366. PMLR,
  18--24 Jul 2021.
\newblock URL \url{https://proceedings.mlr.press/v139/schlag21a.html}.

\bibitem[Shaj et~al.(2026)Shaj, Barker, Scannell, Szecsenyi, Crowley, and
  Storkey]{shaj2026kalman}
Vaisakh Shaj, Cameron Barker, Aidan Scannell, Andras Szecsenyi, Elliot~J.
  Crowley, and Amos Storkey.
\newblock {Kalman} linear attention: Parallel bayesian filtering for efficient
  language modeling and state tracking.
\newblock In \emph{Forty-third International Conference on Machine Learning},
  2026.
\newblock URL \url{https://openreview.net/forum?id=9h7sSJe4jN}.

\bibitem[Smith et~al.(2026)Smith, Piray, Teal, Gok, and Pollak]{smith2026early}
Karen~E. Smith, Payam Piray, Alyssa Teal, Selin Gok, and Seth~D. Pollak.
\newblock Early life stress affects human decision making by increasing
  expectations of volatility.
\newblock \emph{Developmental Science}, 29\penalty0 (4):\penalty0 e70238, 2026.
\newblock \doi{10.1111/desc.70238}.
\newblock URL \url{https://onlinelibrary.wiley.com/doi/abs/10.1111/desc.70238}.

\bibitem[Su et~al.(2024)Su, Ahmed, Lu, Pan, Bo, and Liu]{Su_2024}
Jianlin Su, Murtadha Ahmed, Yu~Lu, Shengfeng Pan, Wen Bo, and Yunfeng Liu.
\newblock {RoFormer}: Enhanced transformer with rotary position embedding.
\newblock \emph{Neurocomputing}, 568:\penalty0 127063, February 2024.
\newblock ISSN 0925-2312.
\newblock \doi{10.1016/j.neucom.2023.127063}.
\newblock URL \url{https://doi.org/10.1016/j.neucom.2023.127063}.

\bibitem[Sutton(1992)]{sutton1992gain}
Richard~S. Sutton.
\newblock Gain adaptation beats least squares?
\newblock In \emph{Proceedings of the Seventh Yale Workshop on Adaptive and
  Learning Systems}, pp.\  161--166, New Haven, CT, 1992. Yale University.
\newblock URL
  \url{http://incompleteideas.net/papers/sutton-92b-remastered.pdf}.

\bibitem[Tillet et~al.(2019)Tillet, Kung, and Cox]{Tillet_2019}
Philippe Tillet, H.~T. Kung, and David Cox.
\newblock {Triton}: an intermediate language and compiler for tiled neural
  network computations.
\newblock In \emph{Proceedings of the 3rd ACM SIGPLAN International Workshop on
  Machine Learning and Programming Languages}, PLDI '19, pp.\  10--19. ACM,
  June 2019.
\newblock \doi{10.1145/3315508.3329973}.
\newblock URL \url{https://doi.org/10.1145/3315508.3329973}.

\bibitem[Vaswani et~al.(2017)Vaswani, Shazeer, Parmar, Uszkoreit, Jones, Gomez,
  Kaiser, and Polosukhin]{NIPS2017_3f5ee243}
Ashish Vaswani, Noam Shazeer, Niki Parmar, Jakob Uszkoreit, Llion Jones,
  Aidan~N Gomez, \L~ukasz Kaiser, and Illia Polosukhin.
\newblock Attention is all you need.
\newblock In I.~Guyon, U.~Von Luxburg, S.~Bengio, H.~Wallach, R.~Fergus,
  S.~Vishwanathan, and R.~Garnett (eds.), \emph{Advances in Neural Information
  Processing Systems}, volume~30. Curran Associates, Inc., 2017.
\newblock URL
  \url{https://proceedings.neurips.cc/paper_files/paper/2017/file/3f5ee243547dee91fbd053c1c4a845aa-Paper.pdf}.

\bibitem[von Oswald et~al.(2026)von Oswald, Scherrer, Kobayashi, Versari, Yang,
  Schlegel, Maile, Schimpf, Sieberling, Meulemans, Lajoie, A.~Saurous, Frenkel,
  Pascanu, Aguera~y Arcas, and Sacramento]{vonoswald2026mesanet}
Johannes von Oswald, Nino Scherrer, Seijin Kobayashi, Luca Versari, Songlin
  Yang, Maximilian Schlegel, Kaitlin Maile, Yanick Schimpf, Oliver Sieberling,
  Alexander Meulemans, Guillaume Lajoie, Rif A.~Saurous, Charlotte Frenkel,
  Razvan Pascanu, Blaise Aguera~y Arcas, and Joao Sacramento.
\newblock {MesaNet}: Sequence modeling by locally optimal test-time training.
\newblock In C.~Vondrick, B.~Hariharan, C.~Raffel, L.~Pinto, D.~Yang, and
  A.~Faust (eds.), \emph{International Conference on Learning Representations},
  volume 2026, pp.\  26397--26447, 2026.
\newblock URL
  \url{https://proceedings.iclr.cc/paper_files/paper/2026/file/2c3d5a4eff20d24390a445762d61bfbb-Paper-Conference.pdf}.

\bibitem[Wang et~al.(2026{\natexlab{a}})Wang, Shi, and Fox]{wang2026test}
Ke~Alexander Wang, Jiaxin Shi, and Emily~B. Fox.
\newblock Test-time regression: a unifying framework for designing sequence
  models with associative memory.
\newblock \emph{Journal of Machine Learning Research}, 27\penalty0
  (169):\penalty0 1--41, 2026{\natexlab{a}}.
\newblock URL \url{https://www.jmlr.org/papers/v27/25-0903.html}.

\bibitem[Wang et~al.(2026{\natexlab{b}})Wang, Cheng, Shi, Mo, Tang, Xie, Wu,
  Ma, Xia, Xue, Yang, and Yang]{ICLR2026_76fb9228}
Lei Wang, Yu~Cheng, Yining Shi, Zhiwen Mo, Zhengju Tang, Wenhao Xie, Tong Wu,
  Lingxiao Ma, Yuqing Xia, Jilong Xue, Fan Yang, and Zhi Yang.
\newblock {TileLang}: Bridge programmability and performance in modern neural
  kernels.
\newblock In C.~Vondrick, B.~Hariharan, C.~Raffel, L.~Pinto, D.~Yang, and
  A.~Faust (eds.), \emph{International Conference on Learning Representations},
  volume 2026, pp.\  73099--73134, 2026{\natexlab{b}}.
\newblock URL
  \url{https://proceedings.iclr.cc/paper_files/paper/2026/file/76fb92288bf90360c527efb0d1c2aba6-Paper-Conference.pdf}.

\bibitem[Wolf et~al.(2020)Wolf, Debut, Sanh, Chaumond, Delangue, Moi, Cistac,
  Rault, Louf, Funtowicz, Davison, Shleifer, von Platen, Ma, Jernite, Plu, Xu,
  Le~Scao, Gugger, Drame, Lhoest, and Rush]{wolf-etal-2020-transformers}
Thomas Wolf, Lysandre Debut, Victor Sanh, Julien Chaumond, Clement Delangue,
  Anthony Moi, Pierric Cistac, Tim Rault, Remi Louf, Morgan Funtowicz, Joe
  Davison, Sam Shleifer, Patrick von Platen, Clara Ma, Yacine Jernite, Julien
  Plu, Canwen Xu, Teven Le~Scao, Sylvain Gugger, Mariama Drame, Quentin Lhoest,
  and Alexander Rush.
\newblock Transformers: State-of-the-art natural language processing.
\newblock In Qun Liu and David Schlangen (eds.), \emph{Proceedings of the 2020
  Conference on Empirical Methods in Natural Language Processing: System
  Demonstrations}, pp.\  38--45, Online, October 2020. Association for
  Computational Linguistics.
\newblock \doi{10.18653/v1/2020.emnlp-demos.6}.
\newblock URL \url{https://aclanthology.org/2020.emnlp-demos.6/}.

\bibitem[Yang \& Zhang(2024)Yang and Zhang]{yang2024fla}
Songlin Yang and Yu~Zhang.
\newblock {FLA}: A {Triton}-based library for hardware-efficient
  implementations of linear attention mechanism, January 2024.
\newblock URL \url{https://github.com/fla-org/flash-linear-attention}.

\bibitem[Yang et~al.(2024)Yang, Wang, Zhang, Shen, and
  Kim]{NEURIPS2024_d13a3eae}
Songlin Yang, Bailin Wang, Yu~Zhang, Yikang Shen, and Yoon Kim.
\newblock Parallelizing linear transformers with the delta rule over sequence
  length.
\newblock In A.~Globerson, L.~Mackey, D.~Belgrave, A.~Fan, U.~Paquet,
  J.~Tomczak, and C.~Zhang (eds.), \emph{Advances in Neural Information
  Processing Systems}, volume~37, pp.\  115491--115522. Curran Associates,
  Inc., 2024.
\newblock \doi{10.52202/079017-3668}.
\newblock URL
  \url{https://proceedings.neurips.cc/paper_files/paper/2024/file/d13a3eae72366e61dfdc7eea82eeb685-Paper-Conference.pdf}.

\bibitem[Yang et~al.(2025)Yang, Kautz, and Hatamizadeh]{ICLR2025_4904fad1}
Songlin Yang, Jan Kautz, and Ali Hatamizadeh.
\newblock {Gated Delta Networks}: Improving {Mamba2} with delta rule.
\newblock In Y.~Yue, A.~Garg, N.~Peng, F.~Sha, and R.~Yu (eds.),
  \emph{International Conference on Learning Representations}, volume 2025,
  pp.\  29687--29707, 2025.
\newblock URL
  \url{https://proceedings.iclr.cc/paper_files/paper/2025/file/4904fad153f6434a7bcf04465d4be2cc-Paper-Conference.pdf}.

\bibitem[Zellers et~al.(2019)Zellers, Holtzman, Bisk, Farhadi, and
  Choi]{zellers-etal-2019-hellaswag}
Rowan Zellers, Ari Holtzman, Yonatan Bisk, Ali Farhadi, and Yejin Choi.
\newblock {H}ella{S}wag: Can a machine really finish your sentence?
\newblock In Anna Korhonen, David Traum, and Llu{\'i}s M{\`a}rquez (eds.),
  \emph{Proceedings of the 57th Annual Meeting of the Association for
  Computational Linguistics}, pp.\  4791--4800, Florence, Italy, July 2019.
  Association for Computational Linguistics.
\newblock \doi{10.18653/v1/P19-1472}.
\newblock URL \url{https://aclanthology.org/P19-1472/}.

\bibitem[Zhang \& Sennrich(2019)Zhang and Sennrich]{NEURIPS2019_1e8a1942}
Biao Zhang and Rico Sennrich.
\newblock Root mean square layer normalization.
\newblock In H.~Wallach, H.~Larochelle, A.~Beygelzimer, F.~d\textquotesingle
  Alch\'{e}-Buc, E.~Fox, and R.~Garnett (eds.), \emph{Advances in Neural
  Information Processing Systems}, volume~32. Curran Associates, Inc., 2019.
\newblock URL
  \url{https://proceedings.neurips.cc/paper_files/paper/2019/file/1e8a19426224ca89e83cef47f1e7f53b-Paper.pdf}.

\end{thebibliography}
\bibliographystyle{iclr2027_conference}

\clearpage
\appendix
\section{Filtering Derivation and Approximation}
\label{app:filtering}

\subsection{Conditional feature-space model}
For one head and value channel $j$, let $m_{t-1,j}$ be column $j$ of the previous memory. The model in \Eqref{eq:latent_model} treats projected keys, transitions, and noise variances as given at each filtering step. Assume independent, zero-mean Gaussian process and observation noises, and a Gaussian previous state with mean $m_{t-1,j}$ and covariance $U_{t-1}$. Value channels share this covariance and have independent noises under the conditional model. Prediction gives
\begin{equation}
 m^-_{t,j}=A_tm_{t-1,j},\qquad
 P_t=A_tU_{t-1}A_t^\top+\operatorname{diag}(v_t).
 \label{eq:full_prediction}
\end{equation}
The joint conditional distribution of the latent state and its scalar observation is
\begin{equation}
 \begin{bmatrix}z_{t,j}\\x_{t,j}\end{bmatrix}
 \sim\mathcal{N}\left(
 \begin{bmatrix}m^-_{t,j}\\k_t^\top m^-_{t,j}\end{bmatrix},
 \begin{bmatrix}P_t&P_tk_t\\k_t^\top P_t&s_t+k_t^\top P_tk_t\end{bmatrix}
 \right).
\end{equation}
Conditioning on $x_{t,j}$ therefore gives
\begin{align}
 d_t&=s_t+k_t^\top P_tk_t=\gamma_t^{-1}, &g_t&=\gamma_t P_tk_t,\\
 m_{t,j}&=m^-_{t,j}+g_t(x_{t,j}-k_t^\top m^-_{t,j}),
 &U_t&=P_t-\gamma_t(P_tk_t)(P_tk_t)^\top.
 \label{eq:full_conditioning}
\end{align}
Stacking the channel means yields \Eqref{eq:memory_update}. The covariance update does not depend on the observed values, which makes it possible to compute gains separately from mean-memory updates.

The conditional construction motivates a neural memory update; it is not a Gaussian observation model for discrete language tokens. Projections and noise parameters are trained through the task loss. In MQAR, the model receives no labels identifying hidden changes or observation noise. Its feature-space variance parameters therefore need not equal the generator's change and replacement probabilities.

\subsection{Connection to scalar delta writes}
Suppose the predicted covariance at a step is $P_t=\rho_t I$ with $\rho_t>0$. For a unit-normalized key,
\begin{equation}
 g_t=\beta_t k_t,\qquad \beta_t=\frac{\rho_t}{s_t+\rho_t},\qquad
 M_t=(I-\beta_t k_tk_t^\top)A_tM_{t-1}+\beta_t k_tx_t^\top.
 \label{eq:isotropic_delta}
\end{equation}
With $A_t=I$, this has a delta-rule update; scalar decay $A_t=\alpha_t I$ gives the Gated DeltaNet form, and diagonal decay gives its coordinate-wise extension. This statement concerns the algebra of the mean update under an isotropic prediction. A rank-one observation correction generally produces non-isotropic posterior covariance, so isotropy is not automatically preserved by subsequent filtering steps. Nor does the connection identify separately learned erase and write vectors with a single Kalman gain.

In the isotropic case, the opposite effects of the two uncertainty terms follow directly from
\begin{equation}
 \frac{\partial\beta_t}{\partial\rho_t}=\frac{s_t}{(s_t+\rho_t)^2}>0,
 \qquad
 \frac{\partial\beta_t}{\partial s_t}=-\frac{\rho_t}{(s_t+\rho_t)^2}<0.
\end{equation}
Process noise increases predicted covariance in \Eqref{eq:full_prediction}. Observation noise instead increases the gain denominator. With non-isotropic covariance, the gain can also change direction; the scalar derivatives above do not assert that every signed gain coordinate changes monotonically with every variance parameter.

\subsection{Diagonal projection and short-block covariance}
The assumed-density projection in \Secref{sec:method_diagonal} minimizes the forward KL divergence from the Gaussian posterior to a diagonal Gaussian. To make its direction explicit, let the posterior at the current step be $p=\mathcal{N}(\mu,\Sigma)$ and the approximation be $q=\mathcal{N}(m,D)$, where $D$ is positive and diagonal. Then
\begin{equation}
 \operatorname{KL}(p\|q)=\frac12\left[
 \operatorname{tr}(D^{-1}\Sigma)
 +(m-\mu)^\top D^{-1}(m-\mu)-N
 +\log\frac{\det D}{\det\Sigma}\right].
 \label{eq:diagonal_projection_kl}
\end{equation}
Minimizing first over $m$ gives $m=\mu$. For a diagonal entry $D_{ii}$, the remaining derivative is $(D_{ii}-\Sigma_{ii})/(2D_{ii}^2)$, so the minimum is $D_{ii}=\Sigma_{ii}$. The projection therefore retains the mean and marginal variances and discards the covariances between coordinates. Here $p$ is the posterior obtained from the incoming approximate state.

Voltic-D applies this projection after every token. With diagonal $A_t$, its predicted covariance is diagonal, and taking the diagonal of \Eqref{eq:full_conditioning} gives \Eqref{eq:adf_gain}.

Voltic-Q applies the same projection only at boundaries of $Q$-token blocks. Within each block, it uses the full covariance update implied by the diagonal covariance received at the block's start. For this fixed-decay family, $Q=1$ gives the diagonal update with $A_t=I$, while a single block covering the whole sequence gives the full fixed-decay filter. The evaluated Voltic-D additionally learns $A_t$, as specified in \Secref{sec:method_diagonal}. The diagonal-plus-low-rank representation in \Appref{app:qadf_implementation} evaluates the within-block updates without forming a dense matrix.

For positive $s_t$ and positive semidefinite $P_t$, the posterior covariance remains positive semidefinite. To see this, set $w=P_t^{1/2}k_t$. Then
\begin{equation}
 U_t=P_t^{1/2}\left(I-\frac{ww^\top}{s_t+w^\top w}\right)P_t^{1/2}.
\end{equation}
The middle matrix has eigenvalues one orthogonal to $w$ and $s_t/(s_t+\|w\|^2)>0$ along $w$. Keeping only a covariance diagonal also preserves nonnegative marginal variances. These are properties of the mathematical updates; the precision and evaluation choices of the GPU implementation are specified in \Appref{app:kernel_details}.

\subsection{Computational scope}
A direct dense covariance update uses $O(N^2)$ work and storage per head at each step, in addition to the $N\times P$ mean memory. Voltic-D reduces uncertainty work and recurrent storage to $O(N)$. For fixed-decay Voltic-Q, at most $Q$ rank-one corrections give $O(QN)$ work per token and active uncertainty storage. Both variants preserve the same affine mean-memory form for the chunked kernel. These costs describe uncertainty computation; full training also stores or recomputes activations needed for gradients.

\FloatBarrier
\section{Implementation Details}
\label{app:implementation}

\subsection{Sequence-mixing block}
\label{app:voltic_block}

The MQAR and 45M language-model configurations use the following block. Let $h_t\in\mathbb{R}^{d_{\mathrm{model}}}$ be its normalized input and $H$ its number of heads. Each head has a key of dimension $N$, a value of dimension $P$, and a mean memory of size $N\times P$. Separate linear projections produce queries, keys, and values. To include nearby tokens in these features, each projected channel passes through a causal convolution of width four, followed by SiLU \citep{Elfwing_2018}. Writing the resulting vectors as $\widetilde q_{t,h}$, $\widetilde k_{t,h}$, and $x_{t,h}$, we use
\begin{equation}
 q_{t,h}=N^{-1/2}\widetilde q_{t,h},\qquad
 k_{t,h}=\frac{\widetilde k_{t,h}}{\|\widetilde k_{t,h}\|_2},\qquad
 o_{t,h}=M_{t,h}^{\top}q_{t,h}.
 \label{eq:block_readout}
\end{equation}
Thus, $q_t$ in the memory equations already includes the factor $N^{-1/2}$; queries receive no unit-length normalization. The implementation protects key normalization against division by zero.

The output combines the memory readout $o_{t,h}$ with the current value $x_{t,h}$. A trainable scalar $D_h$, initialized to one, controls this direct value contribution. After concatenating the heads, an input-dependent gate reweights the value channels before RMSNorm \citep{NEURIPS2019_1e8a1942} rescales the resulting vector:
\begin{equation}
 \begin{aligned}
 y_t&=\operatorname{concat}_{h=1}^{H}(o_{t,h}+D_hx_{t,h}),\\
 \operatorname{Mixer}(h_t)&=W_{\mathrm{out}}\,
 \operatorname{RMSNorm}\bigl(y_t\odot\operatorname{SiLU}(W_{\mathrm{gate}}h_t)\bigr).
 \end{aligned}
 \label{eq:block_output}
\end{equation}
RMSNorm operates over all $HP$ concatenated channels, with $\epsilon=10^{-5}$ and trainable weights initialized to one. The output projection returns to $d_{\mathrm{model}}$ for the residual addition. The benchmark appendices specify the number of layers, head dimensions, and separate feed-forward branches.

\subsection{Noise and decay parameterization}
\label{app:rates}

The model must produce positive noise variances from unconstrained projection outputs. Let $r^v_t$ and $r^s_t$ be the volatility and observation-noise projections of $h_t$, respectively. We first compute their log rates,
\begin{equation}
 \ell^v_t=-\exp(\theta_v)\odot\operatorname{softplus}(r^v_t+b_v),\qquad
 \ell^s_t=-\exp(\theta_s)\operatorname{softplus}(r^s_t+b_s),
 \label{eq:log_rates}
\end{equation}
and then set $v_t=\exp(\ell^v_t)$ and $s_t=\exp(\ell^s_t)$. The volatility bias and log scale have one entry per head and state coordinate; the observation-noise bias and log scale have one entry per head. All are trainable. For finite real-valued parameters, each log rate is negative, so $0<v_{t,i}<1$ and $0<s_t<1$. The filtering equations allow any positive variances; the evaluated networks use this bounded parameterization in their learned feature space. The selected configurations use no additional variance floor or convolution in these branches.

For learned-decay Voltic-D, the decay should remain between $\alpha_{\min}=0.2$ and one. Given the decay projection $r^\alpha_t$, define
\begin{equation}
 a^{\mathrm{raw}}_t=\exp(\theta_A)\operatorname{softplus}(r^\alpha_t+b_\alpha),\qquad
 F=-\log\alpha_{\min}>0.
\end{equation}
We obtain a smoothly bounded decay through
\begin{equation}
 \log\alpha_t=F\left[\exp\left(-a^{\mathrm{raw}}_t/F\right)-1\right].
\end{equation}
The operations are coordinate-wise, with a learned scale per head. Since $a^{\mathrm{raw}}_t$ is nonnegative, this maps the projection to $\alpha_{\min}\leq\alpha_t\leq1$. Voltic-Q uses $\alpha_t=1$ and has no learned decay projection.

\subsubsection{Initialization}
\label{app:initialization}
Each independent sequence starts with $M_0=0$ and a diagonal uncertainty $u_0$ stored as a fixed buffer. The two Voltic variants share the following initialization of $u_0$ and the noise parameters, which spreads initial retention across heads and state coordinates. The rates below are nominal values at zero projection output; actual token-dependent rates also depend on the initialized projection weights.

For head $h=1,\ldots,H$, the target retention logit is
\begin{equation}
 \eta_h=\operatorname{logit}(0.98)
 +\bigl[\operatorname{logit}(0.20)-\operatorname{logit}(0.98)\bigr]
 \left(\frac{h-1}{H-1}\right)^{1.1},
 \qquad \lambda_h=\operatorname{sigmoid}(\eta_h).
 \label{eq:init_head_retention}
\end{equation}
For one head, the fraction in this expression is defined as zero, giving $\lambda_1=0.98$. Set the nominal noise ranges to $s_{\min}=0.1$, $s_{\max}=0.98$, $v_{\min}=0.01$, and $v_{\max}=0.45$. We choose
\begin{equation}
 \begin{aligned}
 \log s_h^{(0)}&=\tfrac12\bigl[
 \max(\log s_{\min},\log v_{\min}+\eta_h)
 +\min(\log s_{\max},\log v_{\max}+\eta_h)\bigr],\\
 \xi_h^*&=1-\frac{\lambda_h v_{\min}}{(1-\lambda_h)s_h^{(0)}},
 \qquad \xi_h=\min(\max(\xi_h^*,0),1/2).
 \end{aligned}
 \label{eq:init_noise_budget}
\end{equation}
The fraction $\xi_h$ assigns part of the initial variance budget to $u_0$ and the rest to process noise. To spread that budget across coordinates, define
\begin{equation}
 \eta_h^{\mathrm{hi}}=\log\frac{(1-\xi_h)s_h^{(0)}}{v_{\min}},\qquad
 \eta_h^{\mathrm{lo}}=\log\frac{(1-\xi_h)s_h^{(0)}}{v_{\max}}.
\end{equation}
For $\xi_h^*>1/2$, coordinate $i=1,\ldots,N$ has retention logit
\begin{equation}
 \eta_{h,i}=\eta_h^{\mathrm{hi}}
 -(\eta_h^{\mathrm{hi}}-\eta_h^{\mathrm{lo}})
 \left(\frac{i-1}{N-1}\right)^{a_h},\qquad
 a_h=\frac{\log\bigl[(\eta_h^{\mathrm{hi}}-\eta_h)/(\eta_h^{\mathrm{hi}}-\eta_h^{\mathrm{lo}})\bigr]}{\log(1/2)}.
\end{equation}
Otherwise, the first half of coordinates use $\eta_h^{\mathrm{hi}}$ and the second half use
\begin{equation}
 \eta_{h,i}=\eta_h^{\mathrm{hi}}
 -(\eta_h^{\mathrm{hi}}-\eta_h^{\mathrm{lo}})
 \left(\frac{i-N/2}{N/2}\right)^2,\qquad i=N/2+1,\ldots,N.
\end{equation}
The evaluated state widths are even. These logits determine
\begin{equation}
 B_{h,i}=s_h^{(0)}\exp(-\eta_{h,i}),\qquad
 u_{0,h,i}=\xi_h B_{h,i},\qquad
 v_{h,i}^{(0)}=\operatorname{clip}\bigl((1-\xi_h)B_{h,i},v_{\min},v_{\max}\bigr).
 \label{eq:init_uncertainty}
\end{equation}
The budget satisfies $s_h^{(0)}/(s_h^{(0)}+B_{h,i})=\operatorname{sigmoid}(\eta_{h,i})$, specifying the nominal retention along an individual coordinate at unit decay. With $\phi_0=0.01$, the noise biases and log scales are initialized as
\begin{equation}
 \begin{aligned}
 (b_s)_h=(b_v)_{h,i}&=\operatorname{softplus}^{-1}(\phi_0),\\
 \theta_{s,h}&=\log\frac{-\log s_h^{(0)}}{\phi_0},\qquad
 \theta_{v,h,i}=\log\frac{-\log v_{h,i}^{(0)}}{\phi_0}.
 \end{aligned}
 \label{eq:init_noise_parameters}
\end{equation}
Substituting zero projection outputs in \Eqref{eq:log_rates} recovers the nominal rates above. Voltic-D initializes $\theta_A=0$ and $b_\alpha=\operatorname{softplus}^{-1}(10^{-4})$, giving $\alpha\approx0.9999$ at zero decay-projection output. Voltic-Q uses exact unit decay throughout.

\subsubsection{Fused rate computation and backward}
The noise projections have no recurrent dependencies, so one token-parallel kernel evaluates \Eqref{eq:log_rates} and its exponentials together. It writes the positive rates directly, without storing the log-rate arrays. For Voltic-D, the same kernel also computes $\alpha_t$ from $\log\alpha_t$, moving this exponential outside the uncertainty loop.

The backward pass reuses the saved rates to avoid retaining each intermediate activation. For one scalar noise coordinate, write $c=\exp(\theta)$, $\phi=\operatorname{softplus}(r+b)$, and $\nu=\exp(-c\phi)$. Here $\nu$ denotes either a volatility coordinate or the observation-noise variance. When $\nu>0$, the kernel reconstructs
\begin{equation}
 \phi=-\frac{\log\nu}{c},\qquad
 \frac{\partial\nu}{\partial r}=-c\nu(1-\exp(-\phi)),\qquad
 \frac{\partial\nu}{\partial\theta}=\nu\log\nu.
 \label{eq:fused_rate_backward}
\end{equation}
The bias has the same local derivative as $r$. The kernel multiplies these derivatives by the incoming loss gradient and sums parameter contributions over tokens. Rates that underflow to zero receive zero local gradients, avoiding logarithms of zero. Under deterministic training, parameter-gradient sums use separate partial reductions instead of unordered atomic additions.

\subsection{Evaluating Voltic-Q with diagonal and low-rank terms}
\label{app:qadf_implementation}

Each observation subtracts a rank-one term from the Voltic-Q covariance. We retain the vector and scalar defining each subtraction, allowing later updates to use these terms without constructing the full matrix. Consider a block starting after token $r$ and ending at $e=\min(r+Q,T)$, with uncertainty diagonal $u_r$ passed from the previous block. Define the accumulated diagonal $c_t$, update direction $w_t$, and inverse denominator $\gamma_t$ as
\begin{equation}
 c_t=u_r+\sum_{j=r+1}^{t}v_j,\qquad w_t=P_tk_t,\qquad
 \gamma_t=(s_t+k_t^{\top}w_t)^{-1}.
\end{equation}
With $\gamma_t$ as defined in \Eqref{eq:exact_filter}, the covariance before the observation update at token $t$ is
\begin{equation}
 P_t=\operatorname{diag}(c_t)-\sum_{j=r+1}^{t-1}\gamma_j w_jw_j^{\top}.
 \label{eq:qadf_low_rank}
\end{equation}
Multiplying this expression by $k_t$ gives an update using only the previous direction vectors:
\begin{equation}
 w_t=c_t\odot k_t-\sum_{j=r+1}^{t-1}\gamma_jw_j(w_j^{\top}k_t),
 \qquad g_t=\gamma_t w_t.
 \label{eq:qadf_direction}
\end{equation}
At the block's end, the uncertainty diagonal passed to the next block is
\begin{equation}
 u_e=c_e-\sum_{j=r+1}^{e}\gamma_j w_j^{\odot2}.
 \label{eq:qadf_diagonal_carry}
\end{equation}
These equations evaluate \Eqref{eq:qadf_update} without forming its dense matrices. The direction computation takes $O(QN)$ work per token and the active block contains $O(QN)$ direction entries. Training saves per-token directions, accumulated diagonals, and inverse denominators for the backward pass.

The main MQAR and language-model quality experiments use $N=128$ with $Q=4$. The uncertainty kernel processes tokens in order, distributing the vector coordinates across GPU threads. Threads synchronize before the next token uses a newly stored direction and inverse denominator. The throughput studies use $N=64$ and the register-based kernels described in \Appref{app:throughput_execution}.

\subsubsection{Backward through the uncertainty block}
\label{app:qadf_backward}
Let a bar denote a loss gradient, and let $\overline g_t$ and $\overline u_e$ be the incoming gradients from the memory updates and the next covariance block. The following reverse computation differentiates \Eqref{eq:qadf_direction} and \Eqref{eq:qadf_diagonal_carry} using vectors and scalars. Initialize all accumulated gradients to zero, then set
\begin{equation}
 \begin{aligned}
 \overline w_t&=\gamma_t\overline g_t-2\gamma_t\overline u_e\odot w_t,\\
 \overline\gamma_t&=\overline g_t^{\top}w_t-\overline u_e^{\top}w_t^{\odot2},
 \qquad \overline c_e=\overline u_e.
 \end{aligned}
 \label{eq:qadf_backward_seed}
\end{equation}
Visit tokens in reverse order, from $e$ to $r+1$. Differentiating the denominator gives $\delta_t=-\gamma_t^2\overline\gamma_t$. At each token, first apply
\begin{equation}
 \overline s_t=\delta_t,\qquad
 \overline k_t=\delta_t w_t,\qquad
 \overline w_t\mathrel{+}=\delta_t k_t,
 \label{eq:qadf_backward_denominator}
\end{equation}
where $\mathrel{+}=$ denotes adding a contribution to an accumulated gradient. Using the updated $\overline w_t$, the diagonal product contributes
\begin{equation}
 \overline c_t\mathrel{+}=\overline w_t\odot k_t,\qquad
 \overline k_t\mathrel{+}=\overline w_t\odot c_t.
\end{equation}
For every earlier token $j$ in the same block, define $\tau_{jt}=w_j^\top k_t$ and $\chi_{jt}=\overline w_t^\top w_j$. The rank-one terms contribute
\begin{equation}
 \begin{aligned}
 \overline\gamma_j&\mathrel{-}=\tau_{jt}\chi_{jt},\\
 \overline w_j&\mathrel{-}=\gamma_j(\tau_{jt}\overline w_t+\chi_{jt}k_t),\\
 \overline k_t&\mathrel{-}=\gamma_j\chi_{jt}w_j.
 \end{aligned}
 \label{eq:qadf_backward_directions}
\end{equation}
After reversing the block, the cumulative diagonal in $c_t$ gives
\begin{equation}
 \overline v_t=\sum_{i=t}^{e}\overline c_i,\qquad
 \overline u_r=\sum_{i=r+1}^{e}\overline c_i.
 \label{eq:qadf_backward_carry}
\end{equation}
Passing $\overline u_r$ to the preceding block accounts for all later uncertainty updates. Key gradients from the memory operator are added to $\overline k_t$ before differentiating key normalization and the input projection. The noise gradients pass through \Eqref{eq:fused_rate_backward}. This reverse computation needs $O(QN)$ active direction gradients, in addition to the forward arrays saved for training.

\subsection{Chunked memory computation and overlap}
\label{app:memory_kernel}

To use the generalized delta-rule operator in Flash Linear Attention \citep{yang2024fla}, expand \Eqref{eq:memory_update} as
\begin{equation}
 M_t=(A_t+b_ta_t^\top)M_{t-1}+\kappa_t x_t^\top,\qquad
 a_t=\alpha_t\odot k_t,\quad b_t=-g_t,\quad \kappa_t=g_t.
 \label{eq:memory_kernel_mapping}
\end{equation}
Because $A_t$ is diagonal, $a_t^\top=k_t^\top A_t$, and the right-hand side equals $(I-g_tk_t^\top)A_tM_{t-1}+g_tx_t^\top$. The kernel receives the diagonal decay as $\log\alpha_t$ and organizes the affine memory updates into chunks of $C=16$ tokens.

The gains depend on keys, noise variances, decay, and previous uncertainty. They do not depend on values or the mean memory, so the uncertainty kernel can run ahead of the memory kernel. We divide the sequence into groups of $S$ tokens, called panels, and assign the two kernels to separate CUDA streams \citep{nvidia2025cuda}. Once the gains for a panel are ready, the memory stream processes that panel while the uncertainty stream prepares the next one. CUDA events enforce the required dependencies. This overlap is available when a sequence contains multiple panels.

The covariance block length $Q$ determines when Voltic-Q discards off-diagonal covariance. The memory chunk length $C$ and panel length $S$ organize the evaluation of the resulting recurrence. Choosing $S$ as a multiple of both $C$ and $Q$ keeps panel boundaries aligned with memory chunks and covariance blocks. For Voltic-D, only alignment with $C$ is required. The benchmark appendices give the selected panel lengths.

\subsection{States and gradients across panels}
\label{app:backward}

At each panel boundary, the forward pass carries the uncertainty and mean memory to the next panel. The backward pass carries their loss gradients in the opposite direction, allowing a later prediction to affect an earlier memory update. A panel boundary therefore does not truncate either gradient path.

In \Eqref{eq:memory_kernel_mapping}, the same gain enters the memory operator's write term through $\kappa_t=g_t$ and its erase term through $b_t=-g_t$. With the gradient notation above, the chain rule gives
\begin{equation}
 \overline{g}_t=\overline{\kappa}_t-\overline{b}_t,
 \qquad \kappa_t=g_t,\quad b_t=-g_t.
 \label{eq:tied_gradient}
\end{equation}
The first term accounts for the gain's effect on writing the new value; the second accounts for its effect on erasing the old prediction. Both must reach the uncertainty backward pass because both depend on the same gain.

Learned decay also affects more than one computation: it changes predicted uncertainty, the factor $a_t=\alpha_t\odot k_t$, and the memory kernel's $\log\alpha_t$ input. The contribution through $a_t$ is
\begin{equation}
 \overline{\log\alpha_t}^{\,(a)}=\overline{a}_t\odot\alpha_t\odot k_t.
\end{equation}
We add this term to the direct decay-input gradient and the contribution from the uncertainty update before differentiating the decay parameterization in \Appref{app:rates}.

\subsection{Kernel specialization and execution settings}
\label{app:kernel_details}

We implement model layers and automatic differentiation in PyTorch \citep{NEURIPS2019_bdbca288}, with custom GPU kernels written in Triton \citep{Tillet_2019}. \Tabref{tab:software} lists the software versions. The following details describe the MQAR and language-model quality configurations; \Appref{app:throughput_execution} gives the choices for throughput measurements.

\subsubsection{Fused projections and short convolutions}
Query, key, value, output-gate, and noise projections share one matrix multiplication; Voltic-D includes the decay projection in the same operation. Their weights remain separately trainable parameters. The selected short-convolution backend is \texttt{causal-conv1d} \citep{dao2023causalconv}, which applies the convolution bias and SiLU activation in the convolution kernel. The three query, key, and value convolutions are invoked separately.

\subsubsection{Panel inputs and saved intermediates}
The Voltic-D uncertainty kernel writes the gain, $a_t=\alpha_t\odot k_t$, and the memory decay input into contiguous arrays for the current panel. The memory kernel consumes these arrays directly. This avoids concatenating all panels into a full-sequence gain array before starting memory computation. We retain the matrix products needed by the analytic backward pass. In the sign-specialized kernels, a product involving $-g_t$ is reconstructed by negating the corresponding product involving $g_t$ after loading it into registers. This removes separate arrays and GPU launches for these negations in both forward and backward computation.

\subsubsection{Compact memory computation for Voltic-Q}
For Voltic-Q, fixed decay implies $\log\alpha_t=0$ and $a_t=k_t$. When the processed sequence fits in one panel, the specialized path omits cumulative decay calculations, multiplication by unit decay factors, and their backward derivatives. A single preparation kernel forms the within-chunk products, replacing separate input-packing and decay-prefix operations. It also omits the stored zero-decay array and reuses the normalized key directly. This path applies to processed document segments as well as complete sequences. Longer inputs use the panel pipeline while carrying uncertainty and memory between panels.

\subsubsection{Precision and execution choices}
Parameters and optimizer states are FP32. Under BF16 autocasting, the projections and memory-kernel inputs use BF16, while uncertainty updates and their reductions use FP32. For Voltic-D, the kernel evaluates the posterior diagonal as $u_{t,i}=p_{t,i}(d_t-p_{t,i}k_{t,i}^{2})/d_t$, an algebraic rearrangement of \Eqref{eq:adf_gain}. The surrounding PyTorch layers execute eagerly. MQAR fixes $S=256$. The 45M language models select $S$ by timing supported values before training, restoring model parameters and random-number state afterward. \Appref{app:lm_optimization} gives the timing candidates.

\FloatBarrier
\section{MQAR Covariance-Block Ablation}
\label{app:covariance_blocks}

The main experiments use Voltic-Q with $Q=4$. Here we compare it with $Q=16$ on MQAR. The longer block retains covariance between state coordinates over more tokens before discarding the off-diagonal entries. Both configurations use fixed decay and the same model dimensions, parameter count, and training protocol in \Appref{app:mqar_models}. Each selects its learning rate independently by the procedure in \Appref{app:mqar_evaluation}. For $Q=16$, the selected rates are $0.003$ on Standard and Volatile MQAR and $0.01$ on Stochastic MQAR and Vol \& Sto.

\Tabref{tab:covariance_mqar} compares the block lengths at all four memory sizes. Averaging the two training sizes within each run, their mean accuracies differ by less than 0.3 percentage points on every task. Both are nearly perfect on Standard MQAR; $Q=16$ has a slightly higher mean on Volatile MQAR, while $Q=4$ has slightly higher means on Stochastic MQAR and Vol \& Sto.

\begin{table}[!htbp]
\centering
\caption{MQAR covariance-block comparison: accuracy (\%), mean $\pm$ SEM over five runs. Bold marks the higher displayed mean for each task and memory size; ties are both bold.}
\label{tab:covariance_mqar}
\small
\setlength{\tabcolsep}{3.5pt}
\begin{tabular}{lrrrr}
\toprule
Model & Standard & Volatile & Stochastic & Vol \& Sto \\
\midrule
\multicolumn{5}{l}{\textbf{(a) $K=16$, $T=128$}} \\
\rowcolor{VolticTint}
Voltic-Q ($Q=4$) & $\mathbf{100.00}\,\pm\,0.00$ & $99.82\,\pm\,0.06$ & $\mathbf{88.66}\,\pm\,0.13$ & $\mathbf{73.52}\,\pm\,0.23$ \\
\rowcolor{VolticTint}
Voltic-Q ($Q=16$) & $\mathbf{100.00}\,\pm\,0.00$ & $\mathbf{99.88}\,\pm\,0.05$ & $88.45\,\pm\,0.24$ & $73.14\,\pm\,0.29$ \\
\midrule
\multicolumn{5}{l}{\textbf{(b) $K=64$, $T=512$}} \\
\rowcolor{VolticTint}
Voltic-Q ($Q=4$) & $\mathbf{100.00}\,\pm\,0.00$ & $99.13\,\pm\,0.42$ & $\mathbf{86.56}\,\pm\,0.53$ & $72.25\,\pm\,0.34$ \\
\rowcolor{VolticTint}
Voltic-Q ($Q=16$) & $\mathbf{100.00}\,\pm\,0.00$ & $\mathbf{99.53}\,\pm\,0.11$ & $86.39\,\pm\,0.59$ & $\mathbf{72.42}\,\pm\,0.26$ \\
\midrule
\multicolumn{5}{l}{\textbf{(c) $K=128$, $T=1,024$}} \\
\rowcolor{VolticTint}
Voltic-Q ($Q=4$) & $97.65\,\pm\,0.62$ & $69.36\,\pm\,5.21$ & $68.96\,\pm\,1.44$ & $61.23\,\pm\,1.16$ \\
\rowcolor{VolticTint}
Voltic-Q ($Q=16$) & $\mathbf{98.01}\,\pm\,0.96$ & $\mathbf{70.46}\,\pm\,2.91$ & $\mathbf{73.27}\,\pm\,2.89$ & $\mathbf{62.21}\,\pm\,0.93$ \\
\midrule
\multicolumn{5}{l}{\textbf{(d) $K=256$, $T=2,048$}} \\
\rowcolor{VolticTint}
Voltic-Q ($Q=4$) & $\mathbf{43.51}\,\pm\,6.97$ & $15.33\,\pm\,3.98$ & $27.74\,\pm\,5.57$ & $23.70\,\pm\,2.81$ \\
\rowcolor{VolticTint}
Voltic-Q ($Q=16$) & $41.28\,\pm\,6.24$ & $\mathbf{20.67}\,\pm\,4.32$ & $\mathbf{32.34}\,\pm\,5.83$ & $\mathbf{25.52}\,\pm\,2.31$ \\
\bottomrule
\end{tabular}
\end{table}

At the extrapolation sizes, $Q=16$ has higher mean accuracy on Volatile MQAR, Stochastic MQAR, and Vol \& Sto at both $K=128$ and $K=256$. On Standard MQAR, $Q=16$ is slightly higher at $K=128$, while $Q=4$ is higher at $K=256$. Thus, retaining covariance over a longer block can help with extrapolation, even though its effect at the training sizes is small in these experiments.

\FloatBarrier
\section{Software, Hardware, and Training Time}
\label{app:resources}

\subsection{Software environment}
\label{app:software}

\Tabref{tab:software} lists the main packages in the MQAR and 45M language-model environment. PyTorch and Triton provide the framework and custom kernels described in \Appref{app:kernel_details}; FLA provides the generalized delta-rule operations. The Mamba package \citep{dao2023mambasoftware} and TileLang \citep{ICLR2026_76fb9228} support the Mamba-3 baselines. For language modeling, Transformers \citep{wolf-etal-2020-transformers} and Accelerate \citep{accelerate} provide model and training utilities, while Datasets \citep{lhoest-etal-2021-datasets} and Tokenizers \citep{moi2023tokenizers} support data preparation and tokenization. NumPy \citep{Harris_2020} supports data generation and array processing, and \href{https://github.com/arogozhnikov/einops}{Einops} supplies tensor reshaping operations. Language Model Evaluation Harness is used as described in \Secref{sec:lm_setup}.

The xLSTM baselines use the original mLSTM and sLSTM blocks from the \href{https://github.com/NX-AI/xlstm/tree/ab22eadbd245f293dd8dec38ed29963d73758a12}{official xLSTM repository}, version 2.0.6 at revision \texttt{ab22eadbd245}. The source is unchanged; the MQAR interface and removal of the hybrid's external feed-forward branch are described in \Appref{app:mqar_models}.

Retrieval evaluations use the PyTorch and CUDA versions in \Tabref{tab:software}. Training-throughput and cached-decoding measurements use PyTorch 2.13.0+cu130 and Triton 3.7.1; their Python, CUDA, FLA, \texttt{causal-conv1d}, \texttt{mamba-ssm}, and TileLang versions match the corresponding table entries.

KLA language modeling and retrieval use the \texttt{cuda\_v3\_fast} backend based on the official implementation at revision \texttt{ad5e64de903e}, with deterministic gradient reductions. Training-throughput measurements use that release's unmodified CUDA backend; cached decoding uses its state-producing Triton prefill and single-token step implementation (\Appref{app:throughput}). LongBench's official metrics use the \texttt{rouge} package, version 1.0.1; benchmark code and dataset revisions are specified in \Appref{app:retrieval}.

\begin{table}[!htbp]
\centering
\caption{Main software versions for MQAR and 45M language modeling.}
\label{tab:software}
\small
\begin{tabular}{ll}
\toprule
Package or runtime & Version \\
\midrule
Python & 3.11.14 \\
PyTorch & 2.12.0+cu130 \\
CUDA & 13.0 \\
Triton & 3.7.0 \\
Flash Linear Attention / FLA Core & 0.5.1 / 0.5.1 \\
\texttt{causal-conv1d} & 1.6.2.post1 \\
\texttt{mamba-ssm} & 2.3.2.post1 \\
TileLang & 0.1.8 \\
Transformers & 5.9.0 \\
Accelerate & 1.13.0 \\
Datasets & 4.8.5 \\
Tokenizers & 0.22.2 \\
NumPy & 2.4.6 \\
Einops & 0.8.2 \\
Language Model Evaluation Harness & 0.4.12 \\
\bottomrule
\end{tabular}
\end{table}

\subsection{Hardware}
\label{app:hardware}

Each training run uses one GPU. In the MQAR campaign, Voltic, KLA, and Mamba-3 MIMO use NVIDIA H200 NVL GPUs; the other models use RTX A6000 GPUs. Evaluation-only variance interventions use H200 NVL or RTX A6000, with each intervention and its full control evaluated on the same device. At 45M scale, Voltic, KLA, and both Mamba-3 variants use H200-class GPUs, while KDA and GDN-2 use RTX A6000 GPUs; the Transformer runs use both GPU types. All runs within each study retain the model dimensions, data, and optimization settings specified in its experimental setup.

The original RULER and BASED retrieval evaluations use NVIDIA Ampere and Ada GPUs: RTX 3090, RTX A4000/A6000, L4/L40, RTX 4060 Ti, RTX 4070 Ti SUPER, RTX 4080 SUPER, RTX 4090, and RTX 4500/6000 Ada. LongBench and BABILong use RTX 3090 and RTX A6000 GPUs. KLA's evaluations across all four groups use H200 and H200 NVL GPUs. The same prompts and scoring rules are used across devices.

All reported training-throughput and cached-decoding measurements run serially on the same physical H200 NVL at a 600\,W power limit. The fixed-token training view is a subset of this device's batch-size grid; it replaces the earlier standalone sequence-length measurements on a different H200 NVL. Baseline updates and recovered large-batch measurements retain the same device and runtime.

\subsection{Approximate training time}
\label{app:training_resources}

\Tabref{tab:training_time} gives approximate training time over the recorded optimizer updates. For language modeling, we reconstruct this time from the token counts and processing rates in the training logs, counting each completed update once across checkpoint restarts. MQAR records its training-loop duration directly. These times include data loading and checkpoint writes within the training intervals. They exclude data preparation, model initialization, final downstream evaluation, and time waiting to restart a job.

\begin{table}[!htbp]
\centering
\caption{Approximate training time: completed MQAR search, fresh-seed, and learning-rate follow-up runs, and 45M final runs.}
\label{tab:training_time}
\small
\begin{tabular}{lrrr}
\toprule
Study & Runs & Hours per run & Summed hours \\
\midrule
MQAR, 10,000 updates & 316 & 0.02--0.24 & 31.6 \\
45M, 4,800 updates & 24 & 2.1--34.5 & 285 \\
\bottomrule
\end{tabular}
\end{table}

The MQAR row covers completed learning-rate searches and fresh runs for the four tasks, including the covariance-block study in \Appref{app:covariance_blocks} and the 20 additional Transformer runs at peak rate $0.001$ in \Appref{app:mqar_transformer}. It excludes twelve numerically failed search candidates, evaluation-only interventions, and other Transformer diagnostics, including the curriculum experiment. The 45M row covers three final seeds per reported model. These times cover the recorded intervals of completed runs, rather than total allocated GPU time.

\section{MQAR Experimental Details}
\label{app:mqar}

\subsection{Sequence generation}
\label{app:mqar_generation}

The four tasks in \Secref{sec:mqar_setup} use sequences of length $T=8K$ with $K$ distinct keys. Keys are sampled without replacement from token IDs 1--4,095. Initial values are sampled without replacement from IDs 4,096--8,191. Unassigned positions contain uniform random tokens from the full vocabulary, including IDs that may also appear as keys or values. Each key has one designated query. Only predictions at these query-key positions contribute to the loss and accuracy; the output at a query is compared directly with its target value, without shifting the labels to another position.

\phantomsection\label{app:mqar_stochastic_generation}\label{app:mqar_joint_generation}\textbf{Observation rounds and queries.} All four tasks share three observation rounds and the same token layout. Each round visits all $K$ keys in an independently shuffled order, occupying $2K$ tokens. Queries occupy every other position in the last $2K$ tokens, in an independently shuffled key order. The target is the final hidden value.

Let $h$ be the probability that a key's hidden value changes before rounds two and three, and let $e$ be the probability that a displayed value is incorrect. A hidden change samples uniformly from the 4,095 values different from the preceding hidden value. Independently, an incorrect observation samples uniformly from the 4,095 alternatives to the current hidden value. Observation noise applies in all three rounds, including the first. Values may coincide across keys or rounds after a change or replacement. The four settings are
\begin{equation}
 (h,e)\in\{(0,0),(0.30,0),(0,0.25),(0.30,0.25)\},
\end{equation}
corresponding to Standard MQAR, Volatile MQAR, Stochastic MQAR, and Vol \& Sto MQAR. Models receive no change or noise indicators. Random draws for keys, initial values, observation order, query order, and filler positions are shared across the four conditions; the thresholds $h$ and $e$ determine which candidate changes and replacements take effect. Thus, these comparisons hold the number and positions of observations fixed.

\textbf{Data sharing and splits.} Training, validation, and test generation use separate random-number seeds so that the splits can be reproduced independently. Training data depend on the task, run seed, batch index, and sequence configuration. They do not depend on the model or learning-rate candidate. All rate candidates share initialization and training data within a model--task pair. For each run seed, all models also receive identical initial embedding weights, sampled from a zero-mean normal distribution with standard deviation $0.02$. These embeddings are then trained independently. Validation and test examples are fixed across run seeds and generated independently of training data. Evaluation generates examples in batches of 16, then processes each batch through the model in groups of four. Changing this forward-batch size therefore does not change the examples.

\subsection{Models and optimization}
\label{app:mqar_models}

\Tabref{tab:mqar_models} gives parameter counts and selected learning rates. All models use two layers, hidden width $d_{\mathrm{model}}=128$, and 1,048,576 parameters in a shared embedding/output matrix. We remove the separate feed-forward branch and its normalization while retaining each mixer's internal gates and projections. Total counts span 1.18--1.39 million parameters.

\begin{table}[!htbp]
\centering
\caption{MQAR parameter counts and selected peak learning rates. Other parameters exclude the shared embedding/output matrix.}
\label{tab:mqar_models}
\small
\setlength{\tabcolsep}{3.5pt}
\begin{tabular}{lrrrrrr}
\toprule
Model & Total & Other & Std. & V & S & Vol \& Sto \\
\midrule
\rowcolor{VolticTint}
Voltic-D & 1,283,464 & 234,888 & 0.003 & 0.003 & 0.003 & 0.003 \\
\rowcolor{VolticTint}
Voltic-Q & 1,250,438 & 201,862 & 0.003 & 0.003 & 0.003 & 0.010 \\
KLA & 1,263,234 & 214,658 & 0.003 & 0.003 & 0.003 & 0.003 \\
KDA & 1,315,202 & 266,626 & 0.003 & 0.030 & 0.003 & 0.003 \\
GDN-2 & 1,380,482 & 331,906 & 0.003 & 0.003 & 0.003 & 0.003 \\
Mamba-3 SISO & 1,286,800 & 238,224 & 0.003 & 0.010 & 0.003 & 0.003 \\
Mamba-3 MIMO & 1,394,320 & 345,744 & 0.003 & 0.003 & 0.010 & 0.003 \\
xLSTM (mLSTM) & 1,267,600 & 219,024 & 0.003 & 0.003 & 0.003 & 0.003 \\
xLSTM (hybrid) & 1,192,328 & 143,752 & 0.003 & 0.003 & 0.003 & 0.003 \\
\midrule
Transformer & 1,180,032 & 131,456 & 0.001 & 0.001 & 0.003 & 0.003 \\
\bottomrule
\end{tabular}
\end{table}

\textbf{Voltic.} Both variants use one head with key/state width $N=128$ and value width $P=128$, giving a $128\times128$ mean memory per layer. They use the block in \Appref{app:voltic_block}, including width-four query, key, and value convolutions and gating before full-width RMSNorm. The memory-kernel chunk length is $C=16$ and the panel length is $S=256$, with fused projections, precomputed noise rates, and saved intermediates in the specialized memory kernel. Voltic-D's learned decay is initialized near one and bounded below by $0.2$ (\Appref{app:initialization}). Voltic-Q uses fixed decay $\alpha_t=1$ and covariance-block length $Q=4$.

\textbf{KDA and GDN-2.} Both use head dimension 128 and causal short convolutions of width four. GDN-2 uses value expansion one.

\textbf{KLA.} We use state dimension 64, expansion one, rank-64 time-step projection, Ornstein--Uhlenbeck dynamics, and learned process noise initialized at $0.01$. The value stream comes from the width-four convolution. The observation-variance projection uses the time-step rank, and its minimum observation variance is $10^{-4}$. Queries and keys are normalized, the output is gated, and the composed Triton backend uses scan checkpointing.

\textbf{Mamba-3.} Both variants use expansion two, head dimension 64, state dimension 64, one group, and rotary fraction $0.5$. The chunk lengths are 64 for SISO and 16 for MIMO, with MIMO rank four.

\textbf{xLSTM.} The mLSTM configuration uses two matrix-memory blocks. Each expands the hidden width to 256 and uses four heads with $64\times64$ matrix memories, giving 16,384 matrix-state elements per layer, equal to Voltic's mean-memory size. The hybrid uses an mLSTM block followed by a scalar-memory sLSTM block of width 128 with four heads. Its scalar layer has fewer recurrent-state elements, and its total parameter count is smaller. Both use width-four causal convolutions, tied embeddings, zero dropout, and the upstream output normalization. The hybrid's separate feed-forward branch is removed to follow the common MQAR configuration; it is an adaptation of the original architecture. We use the official parallel mLSTM implementation and the official CUDA sLSTM implementation with FP32 recurrent arithmetic. The source revision is recorded in \Appref{app:software}.

\textbf{Transformer.} We use four causal attention heads and rotary position embeddings \citep{Su_2024} over the full head dimension, with base 10,000 and maximum position 2,048. Attention dropout is zero. Like the other baselines, this is a small configuration of the model defined in \Secref{sec:model_variants}, using the common MQAR network described above. Tables and figures place this full-attention reference after the recurrent models.

\textbf{Optimizer and numerical settings.} AdamW uses moment coefficients $(0.9,0.95)$, numerical-stability constant $\epsilon=10^{-8}$, and global gradient-norm clipping at one. The learning rate warms up for 50 updates, then follows cosine decay to $0.1$ times its peak at update 10,000. Every batch produces one optimizer update, without gradient accumulation. All parameter groups use learning-rate multiplier one, including KLA dynamics and Voltic rate parameters. Weight decay is $0.1$ for the matrix parameter groups, including the shared embedding and Voltic rate projections; biases, normalization parameters, and static dynamics marked for exclusion by the model implementation use zero decay.

Every run uses 10,000 updates with batch size 16, alternating $(T,K)=(128,16)$ and $(512,64)$. Loss is computed only at designated query positions. Parameters are stored in FP32. The forward pass uses bfloat16 (BF16) autocasting while retaining each implementation's recurrent arithmetic. Query logits are converted to FP32 for the loss. Both matrix-multiplication and cuDNN TensorFloat-32 (TF32) modes are disabled. There is no outer \texttt{torch.compile} wrapper. \Appref{app:resources} lists the software versions and hardware.

\subsection{Selection and metrics}
\label{app:mqar_evaluation}

The initial peak learning-rate grid is $\{0.003,0.01,0.03\}$. For each model and task, we train each candidate with seed 600 and select the finite final checkpoint with the highest validation accuracy. Validation uses 2,048 sequences at each of $K=16$ and $K=64$, with equal weight for the two sizes. Ties are resolved by lower query negative log-likelihood (NLL), then by the smaller rate.

After investigating the Transformer's low Standard MQAR accuracy, we add $0.001$ to its grid and apply the same validation rule (\Appref{app:mqar_transformer}). The recurrent models retain the original grid. At each selected rate, we report seeds 600--604: the tuning run and four fresh runs. Test data contain 4,096 shared sequences at each size. Test accuracy is not used for learning-rate selection.

For each query, the model predicts the vocabulary token with the highest score. Accuracy is the percentage of predictions equal to the target value. Each matched task has one condition per $K$, without averaging over update fractions. The main table averages $K=16$ and $K=64$ equally within each run before summarizing across runs; \Appref{app:mqar_results} reports each size separately. All MQAR summaries use mean $\pm$ standard error of the mean (SEM), computed as $s/\sqrt{n}$ from the sample standard deviation $s$ across $n$ training runs. We include all five evaluated runs, including the selected tuning seed. SEM describes uncertainty in the mean; individual-seed scores are shown in \Appref{app:mqar_seed_results}.

To test generalization to longer sequences containing more associations, we evaluate $(T,K)=(1024,128)$ and $(2048,256)$. We evaluate the same five checkpoints at every test size. We also record query NLL, the fraction of sequences with every query correct, and accuracy grouped by whether the last observation was noisy or the hidden value changed. Hidden indicators are used only for this analysis after prediction.

Twelve xLSTM search candidates stop with nonfinite gradients. Each reported model--task pair has five selected and fresh runs with finite evaluations. Numerical failures are recorded separately and are never assigned zero test accuracy.

\subsection{Results at every memory size}
\label{app:mqar_results}

\Tabref{tab:mqar_matched} reports the four tasks at each memory size, using mean $\pm$ SEM over all five runs. Bold and underline mark the highest and second-highest means at the displayed precision. \Appref{app:covariance_blocks} gives the separate covariance-block comparison.

\begin{table}[!htbp]
\centering
\caption{MQAR accuracy (\%) at each memory size: mean $\pm$ SEM over five runs. $K=128/256$ are extrapolation settings. Bold/underline mark the top two displayed means within each task and size, including ties.}
\label{tab:mqar_matched}
\small
\setlength{\tabcolsep}{3.5pt}
\begin{tabular}{lrrrr}
\toprule
Model & Standard & Volatile & Stochastic & Vol \& Sto \\
\midrule
\multicolumn{5}{l}{\textbf{(a) $K=16$, $T=128$}} \\
\rowcolor{VolticTint}
Voltic-D & $\mathbf{100.00}\,\pm\,0.00$ & $\mathbf{99.86}\,\pm\,0.04$ & $\mathbf{88.92}\,\pm\,0.05$ & $72.69\,\pm\,0.40$ \\
\rowcolor{VolticTint}
Voltic-Q & $\mathbf{100.00}\,\pm\,0.00$ & $\underline{99.82}\,\pm\,0.06$ & $88.66\,\pm\,0.13$ & $\mathbf{73.52}\,\pm\,0.23$ \\
KLA & $99.69\,\pm\,0.16$ & $95.64\,\pm\,2.86$ & $84.86\,\pm\,0.54$ & $60.55\,\pm\,7.97$ \\
KDA & $\mathbf{100.00}\,\pm\,0.00$ & $59.03\,\pm\,22.66$ & $86.95\,\pm\,2.00$ & $\underline{73.09}\,\pm\,0.10$ \\
GDN-2 & $99.98\,\pm\,0.02$ & $80.10\,\pm\,17.71$ & $73.04\,\pm\,15.88$ & $69.41\,\pm\,3.69$ \\
Mamba-3 SISO & $21.81\,\pm\,3.92$ & $7.62\,\pm\,2.96$ & $8.79\,\pm\,1.74$ & $2.71\,\pm\,1.11$ \\
Mamba-3 MIMO & $24.11\,\pm\,2.81$ & $4.72\,\pm\,1.15$ & $36.62\,\pm\,14.00$ & $0.47\,\pm\,0.14$ \\
xLSTM (mLSTM) & $\mathbf{100.00}\,\pm\,0.00$ & $89.43\,\pm\,4.22$ & $88.64\,\pm\,0.15$ & $70.04\,\pm\,0.91$ \\
xLSTM (hybrid) & $\mathbf{100.00}\,\pm\,0.00$ & $85.63\,\pm\,0.55$ & $\underline{88.70}\,\pm\,0.03$ & $67.86\,\pm\,1.04$ \\
\midrule
Transformer & $\underline{99.99}\,\pm\,0.00$ & $5.19\,\pm\,0.02$ & $5.93\,\pm\,0.03$ & $5.51\,\pm\,0.00$ \\
\midrule
\multicolumn{5}{l}{\textbf{(b) $K=64$, $T=512$}} \\
\rowcolor{VolticTint}
Voltic-D & $99.96\,\pm\,0.04$ & $\underline{98.72}\,\pm\,0.60$ & $\mathbf{87.75}\,\pm\,0.36$ & $67.03\,\pm\,2.87$ \\
\rowcolor{VolticTint}
Voltic-Q & $\mathbf{100.00}\,\pm\,0.00$ & $\mathbf{99.13}\,\pm\,0.42$ & $\underline{86.56}\,\pm\,0.53$ & $\mathbf{72.25}\,\pm\,0.34$ \\
KLA & $70.19\,\pm\,4.14$ & $54.89\,\pm\,4.47$ & $54.03\,\pm\,2.56$ & $34.70\,\pm\,8.57$ \\
KDA & $\underline{99.99}\,\pm\,0.00$ & $38.50\,\pm\,16.63$ & $81.21\,\pm\,5.12$ & $\underline{68.67}\,\pm\,0.63$ \\
GDN-2 & $99.73\,\pm\,0.25$ & $60.40\,\pm\,15.35$ & $69.75\,\pm\,17.17$ & $60.84\,\pm\,9.97$ \\
Mamba-3 SISO & $1.52\,\pm\,0.26$ & $2.66\,\pm\,1.05$ & $0.89\,\pm\,0.15$ & $0.39\,\pm\,0.14$ \\
Mamba-3 MIMO & $1.82\,\pm\,0.32$ & $0.59\,\pm\,0.12$ & $16.82\,\pm\,6.62$ & $0.10\,\pm\,0.01$ \\
xLSTM (mLSTM) & $99.89\,\pm\,0.05$ & $83.35\,\pm\,4.65$ & $80.40\,\pm\,2.90$ & $63.52\,\pm\,3.40$ \\
xLSTM (hybrid) & $99.87\,\pm\,0.06$ & $88.58\,\pm\,1.54$ & $82.57\,\pm\,0.14$ & $62.08\,\pm\,4.24$ \\
\midrule
Transformer & $99.97\,\pm\,0.00$ & $0.73\,\pm\,0.00$ & $0.85\,\pm\,0.01$ & $0.81\,\pm\,0.01$ \\
\midrule
\multicolumn{5}{l}{\textbf{(c) $K=128$, $T=1,024$}} \\
\rowcolor{VolticTint}
Voltic-D & $94.62\,\pm\,3.92$ & $\underline{67.04}\,\pm\,6.54$ & $\mathbf{72.02}\,\pm\,2.91$ & $45.72\,\pm\,8.18$ \\
\rowcolor{VolticTint}
Voltic-Q & $\underline{97.65}\,\pm\,0.62$ & $\mathbf{69.36}\,\pm\,5.21$ & $\underline{68.96}\,\pm\,1.44$ & $\mathbf{61.23}\,\pm\,1.16$ \\
KLA & $24.42\,\pm\,2.94$ & $14.78\,\pm\,1.80$ & $23.60\,\pm\,3.05$ & $10.98\,\pm\,2.91$ \\
KDA & $96.03\,\pm\,0.37$ & $16.25\,\pm\,8.76$ & $65.55\,\pm\,5.70$ & $\underline{49.63}\,\pm\,1.85$ \\
GDN-2 & $91.34\,\pm\,5.15$ & $17.03\,\pm\,6.44$ & $59.98\,\pm\,14.90$ & $47.17\,\pm\,10.07$ \\
Mamba-3 SISO & $0.44\,\pm\,0.03$ & $0.72\,\pm\,0.28$ & $0.31\,\pm\,0.04$ & $0.17\,\pm\,0.05$ \\
Mamba-3 MIMO & $0.50\,\pm\,0.06$ & $0.25\,\pm\,0.03$ & $5.60\,\pm\,2.21$ & $0.07\,\pm\,0.00$ \\
xLSTM (mLSTM) & $92.55\,\pm\,1.93$ & $42.61\,\pm\,4.66$ & $61.39\,\pm\,5.72$ & $46.83\,\pm\,5.65$ \\
xLSTM (hybrid) & $92.36\,\pm\,1.98$ & $45.47\,\pm\,1.91$ & $64.91\,\pm\,0.34$ & $43.37\,\pm\,6.12$ \\
\midrule
Transformer & $\mathbf{99.08}\,\pm\,0.27$ & $0.12\,\pm\,0.02$ & $0.32\,\pm\,0.00$ & $0.27\,\pm\,0.00$ \\
\midrule
\multicolumn{5}{l}{\textbf{(d) $K=256$, $T=2,048$}} \\
\rowcolor{VolticTint}
Voltic-D & $44.97\,\pm\,6.77$ & $\underline{14.47}\,\pm\,2.16$ & $27.83\,\pm\,4.73$ & $16.65\,\pm\,4.90$ \\
\rowcolor{VolticTint}
Voltic-Q & $43.51\,\pm\,6.97$ & $\mathbf{15.33}\,\pm\,3.98$ & $27.74\,\pm\,5.57$ & $\mathbf{23.70}\,\pm\,2.81$ \\
KLA & $1.37\,\pm\,0.57$ & $1.15\,\pm\,0.35$ & $3.36\,\pm\,1.08$ & $0.79\,\pm\,0.33$ \\
KDA & $46.10\,\pm\,0.94$ & $3.18\,\pm\,1.97$ & $30.06\,\pm\,3.35$ & $16.15\,\pm\,1.14$ \\
GDN-2 & $40.57\,\pm\,7.62$ & $2.62\,\pm\,1.32$ & $31.83\,\pm\,7.98$ & $15.43\,\pm\,4.55$ \\
Mamba-3 SISO & $0.16\,\pm\,0.01$ & $0.19\,\pm\,0.06$ & $0.13\,\pm\,0.01$ & $0.08\,\pm\,0.02$ \\
Mamba-3 MIMO & $0.17\,\pm\,0.00$ & $0.10\,\pm\,0.01$ & $0.69\,\pm\,0.27$ & $0.04\,\pm\,0.00$ \\
xLSTM (mLSTM) & $50.81\,\pm\,3.73$ & $13.35\,\pm\,1.93$ & $\underline{33.11}\,\pm\,5.53$ & $\underline{18.31}\,\pm\,3.25$ \\
xLSTM (hybrid) & $\underline{50.89}\,\pm\,3.70$ & $14.08\,\pm\,0.97$ & $\mathbf{35.54}\,\pm\,0.43$ & $15.12\,\pm\,2.73$ \\
\midrule
Transformer & $\mathbf{61.74}\,\pm\,2.85$ & $0.02\,\pm\,0.00$ & $0.11\,\pm\,0.01$ & $0.09\,\pm\,0.01$ \\
\bottomrule
\end{tabular}
\end{table}

\FloatBarrier

\begin{figure}[!htbp]
\centering
\includegraphics[width=\linewidth]{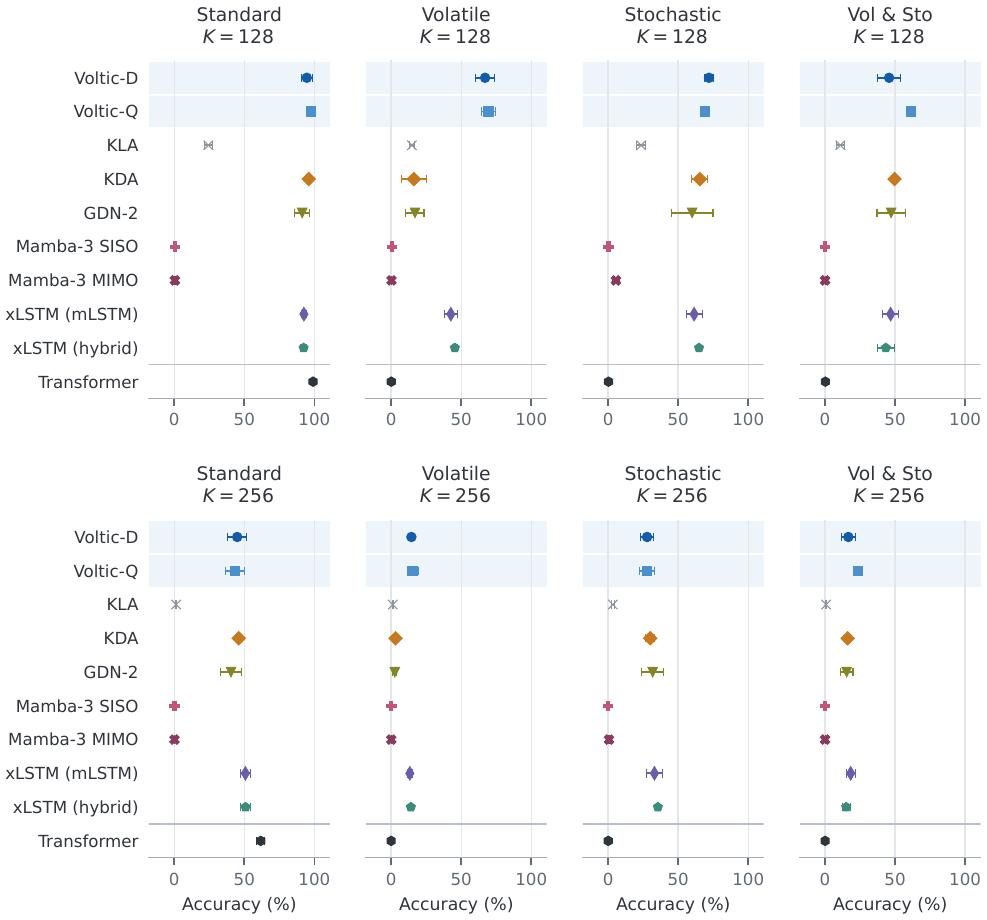}
\caption{MQAR extrapolation at $K=128$ ($T=1,024$) and $K=256$ ($T=2,048$). Points and horizontal bars show mean accuracy and SEM over five runs.}
\label{fig:mqar_extrapolation}
\end{figure}

\textbf{Extrapolation.} \Figref{fig:mqar_extrapolation} shows five-run results beyond the training sizes. Voltic-Q retains a higher Vol \& Sto mean than the other main models at both sizes. Voltic-D and Voltic-Q both remain strong on Volatile MQAR at $K=128$, while accuracy drops substantially at $K=256$. The Transformer has the highest Standard MQAR mean at both extrapolation sizes. The xLSTM hybrid leads Stochastic MQAR at $K=256$.

\FloatBarrier

\subsection{Variation across seeds}
\label{app:mqar_seed_results}

\begin{figure}[!htbp]
\centering
\includegraphics[width=\linewidth]{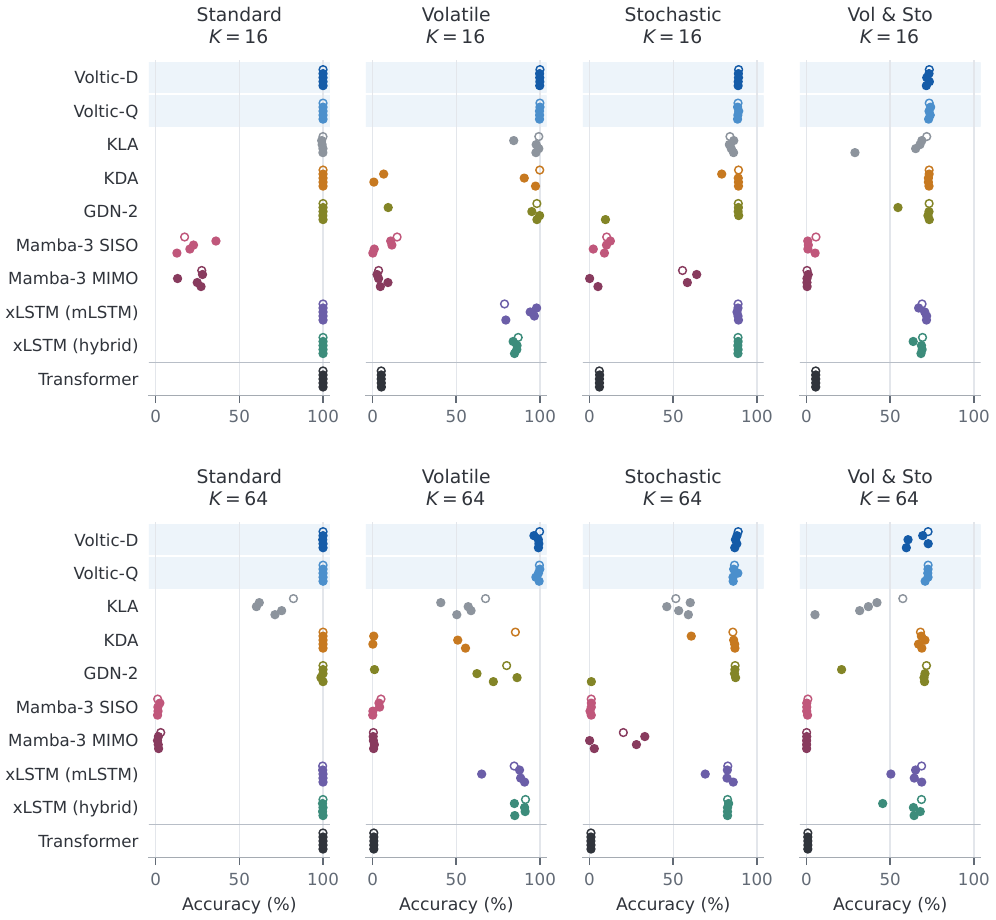}
\caption{MQAR test accuracy for individual runs at $K=16$ and $K=64$. Open points mark the tuning seed; filled points mark the four fresh seeds.}
\label{fig:mqar_seed_accuracy}
\end{figure}

\Figref{fig:mqar_seed_accuracy} shows every run at each training size. On Stochastic MQAR, Mamba-3 MIMO shows substantial variation across initializations at both training sizes. These evaluations remain numerically finite. The individual points show the variation across runs that SEM alone does not describe.

\FloatBarrier

\subsection{Transformer learning rate and curriculum controls}
\label{app:mqar_transformer}

\textbf{Learning-rate follow-up.} The Transformer's low Standard MQAR accuracy in the initial search was unexpected, since this task provides three faithful observations of each fixed association. This prompted a check of the optimization settings. On all four tasks, the initial search selected $0.003$, its lowest candidate. We therefore evaluate an additional peak rate of $0.001$ on all four tasks and seeds 600--604, keeping the architecture, initial weights, ordered training batches, and 10,000-update budget unchanged. The checkpoints use the same validation and test sequences as the main comparison.

The smaller rate restores nearly perfect Standard MQAR accuracy, while accuracy remains low on Volatile MQAR, Stochastic MQAR, and Vol \& Sto. Applying the validation rule in \Appref{app:mqar_evaluation} to the expanded grid selects $0.001$ for Standard and Volatile MQAR and retains $0.003$ for Stochastic MQAR and Vol \& Sto. These are the Transformer rates used in the reported results.

\textbf{One-observation warm-up task.} We use Zoology's \texttt{multiquery\_ar} generator with one query per key, random non-query tokens, and \texttt{power\_a=0.01}. The first $2K$ tokens alternate between keys and their values. The remaining region has candidate query positions at every other token. We sample $K$ of these positions without replacement, using weights proportional to $j^{-0.99}$, where $j$ numbers the candidate positions starting from one. We assign one key to each sampled position, and its target is that key's initial value. This task is used only for the curriculum diagnostic below.

\textbf{Matched training-budget comparison.} To test whether learning the one-observation task first helps with Volatile MQAR, Stochastic MQAR, and Vol \& Sto, we compare two training paths for each target task and seed 600--604. One path trains on the one-observation task for 10,000 updates and then on the target task for 2,000 updates. The other trains on the target task for all 12,000 updates. Both paths use the same original initialization and unchanged Transformer architecture. Each receives 61.44 million input tokens and 7.68 million supervised query positions.

Both paths use peak learning rate $0.003$ in the first phase. At update 10,000, both reset AdamW and the scheduler, then use peak rate $0.001$ for the first 2,000 steps of the same 10,000-step warmup/cosine schedule. The second phase uses identical ordered target batches that do not repeat sequences from either first phase or the test set. All training uses RTX A6000 GPUs; the warm-up runs are retrained on this device to match the target runs. Other optimizer and numerical settings follow \Appref{app:mqar_models}.

We evaluate the final checkpoints on the same 4,096 held-out sequences per target task and memory size. \Tabref{tab:mqar_transformer_curriculum} averages the two sizes within each run, then reports mean $\pm$ SEM over all five seeds. This is a separate diagnostic with a larger training budget than the main benchmark. It uses no further learning-rate search or checkpoint selection.

\begin{table}[!htbp]
\centering
\caption{Transformer curriculum control after 12,000 updates. Warm-up uses the one-observation task described in \Appref{app:mqar_transformer}. Accuracy (\%) averages $K=16/64$: mean $\pm$ SEM over five paired seeds.}
\label{tab:mqar_transformer_curriculum}
\small
\setlength{\tabcolsep}{3.5pt}
\begin{tabular}{lrr}
\toprule
Target task & Target $\rightarrow$ target & Warm-up $\rightarrow$ target \\
\midrule
Volatile & $21.05\,\pm\,18.22$ & $99.43\,\pm\,0.12$ \\
Stochastic & $3.36\,\pm\,0.01$ & $88.49\,\pm\,0.11$ \\
Vol \& Sto & $3.11\,\pm\,0.01$ & $72.29\,\pm\,0.13$ \\
\bottomrule
\end{tabular}
\end{table}

\textbf{Interpretation.} Training first on the one-observation task improves accuracy in every paired seed/task comparison. Extra target-only training produces uneven outcomes on Volatile MQAR: one seed reaches high accuracy, while the others remain low. On Stochastic MQAR and Vol \& Sto, all target-only runs remain low. Thus, the unchanged architecture can achieve much higher accuracy, and extra updates or the optimizer restart alone do not explain the curriculum's advantage under this recipe.

These results support difficulty learning a useful retrieval computation from scratch under the tested training settings. They do not establish a representational limitation of full attention. The warm-up changes both the first-phase task and the placement of observations and queries, so the experiment does not identify which learned computation produces the benefit. Different optimization settings or longer target-only training may also improve performance.
\FloatBarrier

\subsection{Volatility and stochasticity ablations}
\label{app:mqar_ablations}

\textbf{Evaluation-only interventions.} We fix the effective observation variance $s_t=0.5$, the volatility vector $v_t=0.5\mathbf{1}$, or both, at every token and layer during evaluation. The fixed value lies within the implemented variance range (\Appref{app:rates}). Each branch is fixed by zeroing its projection, bias, and log scale; under the parameterization in \Eqref{eq:log_rates}, this gives $\exp[-\operatorname{softplus}(0)]=0.5$. The memory update, keys, values, initial uncertainty, and decay configuration are retained.

We evaluate each full-model checkpoint without changing its remaining weights and compare each intervention with an unmodified replay on the same device and examples. All evaluated runs contribute to the reported mean and SEM.

The reported interventions use 40 full-model checkpoints, four variance settings (including the unmodified control), and four memory sizes. \Tabref{tab:mqar_variance_ablations} summarizes accuracy averaged equally over $K=16/64$.

\textbf{Reliance in trained models.} Replacing learned volatility at evaluation sharply reduces accuracy on every task, including Standard MQAR. Replacing stochasticity has a more variable effect: it strongly affects Voltic-D on Stochastic MQAR, while Voltic-Q largely retains Vol \& Sto accuracy. These interventions measure sensitivity to changing trained internal values. The strong effect on Standard MQAR also shows why this sensitivity alone is not evidence that a branch specifically detects changes or observation noise.

These evaluation-only interventions do not establish that input-dependent variance branches are necessary, or that their outputs are calibrated estimates of the generator's probabilities.

\begin{table}[!htbp]
\centering
\caption{Evaluation-only variance interventions: accuracy (\%) averaged equally over $K=16/64$, mean $\pm$ SEM over five runs. Full controls are unmodified matched replays.}
\label{tab:mqar_variance_ablations}
\small
\setlength{\tabcolsep}{3.5pt}
\begin{tabular*}{\linewidth}{@{\extracolsep{\fill}}lrrrr@{}}
\toprule
Model & Full & $s_t=0.5$ & $v_t=0.5$ & $s_t=v_t=0.5$ \\
\midrule
\multicolumn{5}{l}{\textit{Standard}} \\
Voltic-D & $99.98\,\pm\,0.02$ & $93.49\,\pm\,6.27$ & $22.33\,\pm\,12.10$ & $21.10\,\pm\,11.51$ \\
Voltic-Q & $100.00\,\pm\,0.00$ & $80.08\,\pm\,19.91$ & $13.72\,\pm\,6.09$ & $13.27\,\pm\,5.38$ \\
\midrule
\multicolumn{5}{l}{\textit{Volatile}} \\
Voltic-D & $99.29\,\pm\,0.31$ & $89.09\,\pm\,7.23$ & $23.45\,\pm\,15.86$ & $22.99\,\pm\,15.88$ \\
Voltic-Q & $99.48\,\pm\,0.23$ & $96.60\,\pm\,1.78$ & $1.02\,\pm\,0.38$ & $1.19\,\pm\,0.52$ \\
\midrule
\multicolumn{5}{l}{\textit{Stochastic}} \\
Voltic-D & $88.33\,\pm\,0.20$ & $53.88\,\pm\,13.26$ & $2.31\,\pm\,1.71$ & $2.17\,\pm\,1.40$ \\
Voltic-Q & $87.61\,\pm\,0.33$ & $86.42\,\pm\,0.68$ & $12.47\,\pm\,5.63$ & $12.66\,\pm\,5.76$ \\
\midrule
\multicolumn{5}{l}{\textit{Vol \& Sto}} \\
Voltic-D & $69.86\,\pm\,1.63$ & $66.88\,\pm\,2.94$ & $7.90\,\pm\,3.35$ & $7.07\,\pm\,3.17$ \\
Voltic-Q & $72.89\,\pm\,0.26$ & $72.98\,\pm\,0.24$ & $24.26\,\pm\,8.06$ & $22.40\,\pm\,7.96$ \\
\bottomrule
\end{tabular*}
\end{table}

\FloatBarrier

\FloatBarrier
\section{Language-Modeling Experimental Details}
\label{app:lm}

\subsection{Data preparation and document boundaries}
\label{app:lm_data}

The 45M data use FineWeb-Edu's \texttt{sample-10BT} release at revision \texttt{87f09149ef47}. Source documents are processed in shard and row order. A fixed hash of the document ID assigns each document to one of four disjoint partitions: $1\%$ for tokenizer training, $0.25\%$ for hyperparameter selection, $0.75\%$ for final reporting, and $98\%$ for training. The tokenizer uses byte-level byte-pair encoding with 32,000 entries, including padding, beginning-of-sequence, end-of-sequence, and unknown-token symbols. It is trained on a deterministic sample of at least 400 MiB from its own partition and then shared by all models.

To reduce overlap with evaluation data, we normalize source and evaluation text with Unicode NFKC normalization, lowercasing, ASCII-punctuation removal, and whitespace splitting. We reject a source document if it contains any contiguous 13-word sequence from an evaluation prompt. For prompts of 4--12 words, we instead match the complete normalized prompt. This filter applies to all four partitions and includes the language-task prompts and the six BASED retrieval datasets. LongBench and BABILong were added after the training corpus was fixed and are not included in this overlap filter.

The selection set has 8,388,608 tokens in 4,096 sequences, and the reporting set has 25,165,824 tokens in 12,288 sequences.

Each accepted document ends with an end-of-sequence token. The resulting stream is split into sequences of 2,048 tokens, with document boundaries retained. A recurrent model resets its memory at every sequence and document start, and a Transformer uses causal attention separately within each document segment. Targets at these starts are excluded from the loss, preventing prediction from the preceding document or sequence. Training reads the same ordered sequences for all models and seeds.

\subsection{Architecture and parameter matching}
\label{app:lm_models}

We fix the token-representation dimension $d_{\mathrm{model}}$, layer count $L$, vocabulary, and input/output embedding sharing. We then adjust the intermediate dimension $d_{\mathrm{ff}}$ of each block's gated feed-forward branch to match the parameter target. \Tabref{tab:lm_configuration} gives the resulting sizes and the rates selected by the procedure in \Appref{app:lm_optimization}. The target is 45,508,352 parameters, corresponding to the chosen Mamba-3 SISO configuration; every model is within $0.5\%$ of this target.

\begin{table}[!htbp]
\centering
\caption{45M model dimensions and selected learning rates, with $L=8$ and $d_{\mathrm{model}}=512$. Projection and dynamics columns give learning-rate multipliers.}
\label{tab:lm_configuration}
\small
\setlength{\tabcolsep}{4pt}
\begin{tabular}{lrrrrr}
\toprule
Model & Parameters & $d_{\mathrm{ff}}$ & Peak rate & Projection & Dynamics \\
\midrule
Voltic-D & 45,322,880 & 1,152 & 0.0032 & 1.0 & 0.3 \\
Voltic-Q & 45,580,896 & 1,344 & 0.0032 & 1.0 & 0.1 \\
KLA & 45,576,712 & 1,712 & 0.0032 & --- & 0.1 \\
KDA & 45,434,400 & 1,504 & 0.0032 & --- & --- \\
GDN-2 & 45,483,552 & 1,168 & 0.0032 & --- & --- \\
Mamba-3 SISO & 45,508,352 & 1,280 & 0.0032 & --- & --- \\
Mamba-3 MIMO & 45,655,808 & 1,152 & 0.0032 & --- & --- \\
\midrule
Transformer & 45,425,152 & 1,680 & 0.0016 & --- & --- \\
\bottomrule
\end{tabular}
\end{table}

\textbf{Voltic, KDA, and GDN-2.} Each head uses a 128-dimensional key and a 128-dimensional value, with four heads. This gives the same main memory size as the Mamba-3 configurations below. All use causal short convolutions of width four. Voltic uses the block and noise parameterization in \Appref{app:voltic_block} and \Appref{app:rates}. Voltic-D's learned decay is initialized near one and bounded below by $0.2$ (\Appref{app:initialization}). GDN-2 uses value expansion one. The internal memory chunk lengths are 16 for Voltic and 64 for KDA and GDN-2.

\textbf{KLA.} KLA uses state dimension 64 and expansion one, giving 512 channels. Its observation-noise and time-step projections use rank 64, while values come from the convolution output. The model uses a causal convolution of width four with SiLU, normalized queries and keys, learned process noise, Ornstein--Uhlenbeck discretization, output gating, and a learned scalar skip connection. Readout uses the updated state. We use the optimized CUDA implementation specified in \Appref{app:software}, with checkpointing inside the state-space scan.

\textbf{Mamba-3.} Both variants use state dimension 64, head dimension 64, expansion two, one group, and rotary fraction $0.5$. The expanded dimensions give 16 heads. SISO uses chunk length 64; MIMO uses chunk length 16 and rank four. These configurations include the separate gated feed-forward branch in every block.

\textbf{Transformer.} The full-attention baseline uses four heads, with head dimension 128. Rotary position embeddings cover the full head dimension, with base 10,000 and maximum position 2,048. Attention dropout is zero, and attention respects the document boundaries used by the recurrent models.

\textbf{State budget.} Voltic, KDA, GDN-2, and Mamba-3 each store 65,536 main-memory scalars per layer. KLA stores two $512\times64$ arrays, its information vector and diagonal precision, giving 65,536 recurrent-state scalars in total. Convolution histories and Voltic's uncertainty state are additional storage. Transformer attention retains the keys and values of previous tokens, so its stored history grows with context length.

\subsection{Optimization and hyperparameter selection}
\label{app:lm_optimization}

Every update processes 256 sequences of 2,048 tokens, or 524,288 input tokens. The 4,800-update budget is 2,516,582,400 input tokens. All models accumulate gradients over 32 batches of eight sequences. Hardware and software are specified in \Appref{app:resources}. Parameters and optimizer states use FP32; the forward pass uses BF16 autocasting. Dropout and general gradient checkpointing are disabled.

All models use AdamW with moment coefficients $(0.9,0.95)$ and global gradient-norm clipping at one. The learning rate warms up over 48 updates ($1\%$ of training), then follows cosine decay to $10\%$ of its peak at update 4,800. Weight decay $0.1$ applies to matrix, convolution, and embedding weights, including Voltic's rate-projection weights. Biases, normalization parameters, and the architectures' designated static dynamics parameters use zero decay.

\textbf{Learning-rate selection.} All eight reported configurations, including the Transformer, begin with the same peak-rate candidates: $10^{-4}$, $2\times10^{-4}$, $4\times10^{-4}$, $8\times10^{-4}$, $1.6\times10^{-3}$, and $3.2\times10^{-3}$. We score candidates by mean selection-set NLL over seeds 100 and 101. An initial screen uses 477 updates; the two best candidates are evaluated after 954 updates. If their mean NLLs differ by at most $0.01$, both are evaluated after 1,717 updates. Each stage restarts with the same seed and uses the corresponding prefix of the 4,800-update learning-rate schedule. When the best screened rate is a grid endpoint, one additional rate is tested outward from that endpoint; this adds $6.4\times10^{-3}$ for seven models, while the Transformer requires no extension.

For Voltic, multiplier candidates are $\{0.3,1\}$ for noise/decay projection weights and $\{0.1,0.3,1\}$ for learned log-rate scales and biases. We select these settings after the base rate, using the same screening and confirmation procedure. Exact ties prefer the lower base rate, then the lower projection multiplier, then the lower dynamics multiplier. The selected settings are fixed before training three fresh final seeds, 0--2; reported scores use their final checkpoints.

KLA retains a fixed learning-rate multiplier of $0.1$ and zero weight decay for its static dynamics parameters. Its learned projection matrices use the base learning rate and weight decay $0.1$. KLA's base rate is selected by the same search above; its confirmation stage resolves the two candidates without a near-tie extension.

Voltic uses fused projections, precomputed noise rates, and saved memory-kernel intermediates. Panel length is selected for runtime from $\{256,512,1024,2048\}$ before training, with model parameters and random-number state restored afterward.

\subsection{Evaluation tasks and scoring}
\label{app:lm_evaluation}

The downstream tasks cover final-word prediction with LAMBADA; continuation and reference resolution with HellaSwag \citep{zellers-etal-2019-hellaswag} and WinoGrande \citep{Sakaguchi_2020}; passage comprehension with BoolQ; physical and social commonsense with PIQA \citep{Bisk_2020} and Social IQa \citep{sap-etal-2019-social}; and science questions with ARC-Easy, ARC-Challenge, and OpenBookQA. \Tabref{tab:lm_tasks} lists the splits and their full sizes.

We use the frozen task definitions from Language Model Evaluation Harness, with document-level WikiText-103 and the OpenAI-preprocessed LAMBADA task. Evaluation uses batches of 64, no beginning-of-sequence token, and a maximum model input of 2,048 tokens. Candidate log probabilities are computed with FP32 log-softmax. Inputs are right-padded, and padding positions are excluded from scores. For a candidate that exceeds the context limit, the evaluator retains the rightmost 2,049 tokens before forming the shifted input and targets. WikiText uses rolling windows to score the full documents.

\begin{table}[!htbp]
\centering
\caption{Language-model evaluation splits and scoring metrics.}
\label{tab:lm_tasks}
\small
\begin{tabular}{llrl}
\toprule
Task & Split & Examples & Metric \\
\midrule
WikiText-103 & Test & 62 documents & Word perplexity \\
LAMBADA & Test & 5,153 & Accuracy; perplexity \\
HellaSwag & Validation & 10,042 & Normalized accuracy \\
PIQA & Validation & 1,838 & Accuracy \\
ARC-Easy & Test & 2,376 & Accuracy \\
ARC-Challenge & Test & 1,172 & Normalized accuracy \\
OpenBookQA & Test & 500 & Normalized accuracy \\
WinoGrande & Validation & 1,267 & Accuracy \\
Social IQa & Validation & 1,954 & Accuracy \\
BoolQ & Validation & 3,270 & Accuracy \\
\bottomrule
\end{tabular}
\end{table}

For each accuracy task, the harness applies its fixed prompt and answer-scoring rule. In HellaSwag, ARC-Challenge, and OpenBookQA, it divides each candidate's summed log probability by its character count before choosing the highest score. The remaining multiple-choice tasks use unnormalized conditional log probabilities. LAMBADA accuracy requires the final word's tokens to be predicted correctly. Its perplexity uses the log probability of that target word. WikiText word perplexity divides the summed negative log probability by the number of words before exponentiation; byte perplexity and bits per byte are also recorded.

The eight-task reasoning and comprehension average includes HellaSwag, PIQA, ARC-Easy, ARC-Challenge, OpenBookQA, WinoGrande, Social IQa, and BoolQ. The nine-task average adds LAMBADA. Both give equal weight to their tasks and were recorded by the frozen evaluator for every final checkpoint. We compute each average within each run, then summarize the run scores with their mean and sample standard deviation. \Tabref{tab:lm45_likelihood} reports WikiText perplexity and FineWeb-Edu NLL alongside the downstream averages.

\subsection{Complete 45M results}
\label{app:lm_results}

\Tabpair{tab:lm45_likelihood}{tab:lm45_tasks} report all eight reported 45M configurations, including the Transformer, using their final checkpoints.
\begin{table}[!htbp]
\centering
\caption{45M results after 4,800 updates: mean $\pm$ sample SD over three seeds. NLL uses FineWeb-Edu; WikiText-103 reports word perplexity. The nine-task average adds LAMBADA. Bold/underline mark the top two means.}
\label{tab:lm45_likelihood}
\small
\setlength{\tabcolsep}{4pt}
\begin{tabular}{lrrrr}
\toprule
Model & NLL $\downarrow$ & WikiText PPL $\downarrow$ & 8-task (\%) $\uparrow$ & 9-task (\%) $\uparrow$ \\
\midrule
\rowcolor{VolticTint}
Voltic-D & $3.284\,\pm\,0.004$ & $59.09\,\pm\,0.40$ & $\underline{42.26}\,\pm\,0.37$ & $39.73\,\pm\,0.34$ \\
\rowcolor{VolticTint}
Voltic-Q & $3.306\,\pm\,0.002$ & $60.80\,\pm\,0.21$ & $\mathbf{42.79}\,\pm\,0.14$ & $\underline{39.98}\,\pm\,0.20$ \\
KLA & $3.329\,\pm\,0.003$ & $66.52\,\pm\,0.13$ & $41.95\,\pm\,0.72$ & $39.14\,\pm\,0.57$ \\
KDA & $\mathbf{3.229}\,\pm\,0.002$ & $\mathbf{54.72}\,\pm\,0.07$ & $42.08\,\pm\,0.46$ & $\mathbf{39.99}\,\pm\,0.33$ \\
GDN-2 & $\underline{3.262}\,\pm\,0.003$ & $\underline{56.79}\,\pm\,0.43$ & $41.80\,\pm\,0.58$ & $39.46\,\pm\,0.45$ \\
Mamba-3 SISO & $3.284\,\pm\,0.008$ & $60.14\,\pm\,0.66$ & $41.95\,\pm\,0.91$ & $39.47\,\pm\,0.82$ \\
Mamba-3 MIMO & $3.269\,\pm\,0.001$ & $58.58\,\pm\,0.52$ & $41.84\,\pm\,0.48$ & $39.48\,\pm\,0.51$ \\
\midrule
Transformer & $3.347\,\pm\,0.003$ & $62.08\,\pm\,0.68$ & $42.13\,\pm\,0.42$ & $39.59\,\pm\,0.39$ \\
\bottomrule
\end{tabular}
\end{table}

\begin{table}[!htbp]
\centering
\caption{45M zero-shot accuracy (\%) after 4,800 updates: mean $\pm$ sample SD over three training seeds. Bold/underline mark the top two means.}
\label{tab:lm45_tasks}
\small
\setlength{\tabcolsep}{3.5pt}
\begin{tabular*}{\linewidth}{@{\extracolsep{\fill}}lrrrrr@{}}
\toprule
\multicolumn{6}{l}{\textbf{(a) LAMBADA, HellaSwag, PIQA, and ARC}} \\
Model & LAMBADA & HellaSwag & PIQA & ARC-E & ARC-C \\
\midrule
\rowcolor{VolticTint}
Voltic-D & $19.52\,\pm\,0.46$ & $28.98\,\pm\,0.22$ & $60.97\,\pm\,0.60$ & $48.64\,\pm\,0.23$ & $23.75\,\pm\,0.88$ \\
\rowcolor{VolticTint}
Voltic-Q & $17.52\,\pm\,0.76$ & $28.84\,\pm\,0.35$ & $60.61\,\pm\,0.25$ & $47.91\,\pm\,1.42$ & $\mathbf{24.37}\,\pm\,0.44$ \\
KLA & $16.61\,\pm\,0.69$ & $29.12\,\pm\,0.17$ & $\mathbf{62.15}\,\pm\,0.32$ & $48.25\,\pm\,0.60$ & $23.52\,\pm\,0.36$ \\
KDA & $\mathbf{23.27}\,\pm\,0.90$ & $\mathbf{29.80}\,\pm\,0.27$ & $60.66\,\pm\,0.38$ & $\mathbf{49.21}\,\pm\,0.02$ & $23.81\,\pm\,0.74$ \\
GDN-2 & $\underline{20.70}\,\pm\,0.57$ & $\underline{29.52}\,\pm\,0.05$ & $60.74\,\pm\,0.35$ & $48.20\,\pm\,0.95$ & $23.24\,\pm\,0.34$ \\
Mamba-3 SISO & $19.65\,\pm\,0.36$ & $28.99\,\pm\,0.53$ & $60.70\,\pm\,0.72$ & $48.02\,\pm\,0.78$ & $23.35\,\pm\,0.73$ \\
Mamba-3 MIMO & $20.61\,\pm\,1.01$ & $29.40\,\pm\,0.24$ & $\underline{61.30}\,\pm\,0.23$ & $\underline{48.71}\,\pm\,0.50$ & $23.75\,\pm\,0.57$ \\
\midrule
Transformer & $19.30\,\pm\,0.73$ & $28.07\,\pm\,0.06$ & $59.88\,\pm\,0.44$ & $47.42\,\pm\,0.72$ & $\underline{23.86}\,\pm\,0.58$ \\
\bottomrule
\end{tabular*}
\par\medskip
\begin{tabular*}{\linewidth}{@{\extracolsep{\fill}}lrrrr@{}}
\toprule
\multicolumn{5}{l}{\textbf{(b) OpenBookQA, WinoGrande, Social IQa, and BoolQ}} \\
Model & OpenBookQA & WinoGrande & Social IQa & BoolQ \\
\midrule
\rowcolor{VolticTint}
Voltic-D & $29.20\,\pm\,1.74$ & $51.09\,\pm\,0.59$ & $36.63\,\pm\,0.75$ & $58.83\,\pm\,2.91$ \\
\rowcolor{VolticTint}
Voltic-Q & $\mathbf{31.07}\,\pm\,1.10$ & $50.99\,\pm\,0.48$ & $36.78\,\pm\,0.44$ & $\mathbf{61.74}\,\pm\,0.43$ \\
KLA & $28.47\,\pm\,0.42$ & $\underline{51.93}\,\pm\,1.11$ & $36.22\,\pm\,0.18$ & $55.98\,\pm\,6.25$ \\
KDA & $29.67\,\pm\,1.10$ & $51.85\,\pm\,0.44$ & $36.52\,\pm\,0.78$ & $55.11\,\pm\,2.55$ \\
GDN-2 & $28.87\,\pm\,1.22$ & $50.57\,\pm\,1.28$ & $\underline{36.90}\,\pm\,0.09$ & $56.41\,\pm\,5.04$ \\
Mamba-3 SISO & $29.67\,\pm\,1.70$ & $51.25\,\pm\,0.43$ & $36.44\,\pm\,0.37$ & $57.18\,\pm\,4.63$ \\
Mamba-3 MIMO & $\underline{29.93}\,\pm\,0.46$ & $51.25\,\pm\,1.03$ & $\mathbf{37.34}\,\pm\,0.34$ & $53.05\,\pm\,1.47$ \\
\midrule
Transformer & $29.40\,\pm\,0.40$ & $\mathbf{52.22}\,\pm\,0.56$ & $36.63\,\pm\,0.34$ & $\underline{59.52}\,\pm\,3.88$ \\
\bottomrule
\end{tabular*}
\end{table}

\FloatBarrier
\section{In-Context Retrieval Experimental Details}
\label{app:retrieval}

\subsection{Checkpoints and generation}
\label{app:retrieval_generation}

We evaluate eight 45M configurations, including the Transformer, using the final checkpoints reported in \Secref{sec:lm_setup}, with the same model dimensions, tokenizer, and weights. All models and seeds use identical evaluation examples. There is no additional training, instruction tuning, or model-specific prompt selection.

Generation uses BF16 autocasting and selects the highest-scoring token from FP32 logits. Each example starts with a fresh state. The evaluator scores tokens using causal full-prefix computation; right padding used for kernel alignment cannot provide information to earlier positions. Where a recurrent decoder proposes several tokens together, the full-prefix model verifies them before acceptance, preserving greedy output. These are quality evaluations; their execution times are not used in the throughput comparisons.

\Appref{app:resources} lists the retrieval software and hardware.

For the Transformer, we extend the allowed position range to the evaluation budget. Rotary position embeddings retain their training frequency base of 10,000, without rescaling or long-context adaptation. The recurrent models likewise receive no adaptation for the longer inputs.

KLA uses the same CUDA implementation as its language-model evaluation, with full-prefix greedy decoding. Its input-length limit is raised to 8,192 for RULER without changing model weights; this leaves its 2K logits unchanged. The benchmark scores do not depend on a cached-state decoder.

\subsection{RULER tasks and context budgets}
\label{app:retrieval_ruler}

We use revision \texttt{c3f5e3b4f87f} of the official RULER generator. S1, S2, and S3 correspond to \texttt{niah\_single\_1}, \texttt{niah\_single\_2}, and \texttt{niah\_single\_3}; MK1 is \texttt{niah\_multikey\_1}. S1 uses repeated filler text; the other three use the generator's essay background. Keys are word strings. S1, S2, and MK1 use numeric values, while S3 uses 36-character random identifier strings. MK1 contains four key--value facts with distinct keys and one query; the other tasks contain one fact and one query.

Each task/context combination contains 500 frozen prompts, shared by every model and training seed. Context budgets are 1,024, 2,048, 4,096, and 8,192 tokens, including a reserved generation allowance of 128 tokens. These are upper bounds: generated prompts need not fill the entire budget. \Tabref{tab:retrieval_prompt_lengths} gives the observed mean prompt lengths. Prompts retain the generator's question and answer prefix. Generation stops at the end-of-sequence token or the 128-token limit, without newline stopping.

The RULER scorer strips surrounding whitespace, replaces control characters with newlines, and checks whether the reference string occurs in the answer, ignoring case. Each selected task has one reference answer, so the score is the percentage of examples recovered correctly. We evaluate $4\times4\times500=8,000$ examples per checkpoint. Our overall RULER score averages all sixteen task/context combinations equally, including the S3 and MK1 8K conditions. It is a score for this retrieval subset, not the full thirteen-task RULER benchmark.

\begin{table}[!htbp]
\centering
\caption{Mean RULER prompt length at each total context budget, reserving 128 tokens for generation.}
\label{tab:retrieval_prompt_lengths}
\small
\setlength{\tabcolsep}{4pt}
\begin{tabular}{lrrrr}
\toprule
Task & 1K & 2K & 4K & 8K \\
\midrule
S1 & 880.9 & 1,905.8 & 3,955.8 & 8,039.4 \\
S2 & 703.7 & 1,535.7 & 3,750.1 & 7,699.2 \\
S3 & 519.5 & 1,767.9 & 3,639.1 & 7,693.8 \\
MK1 & 656.0 & 1,817.9 & 3,766.9 & 7,967.1 \\
\bottomrule
\end{tabular}
\end{table}

\subsection{BASED data and scoring}
\label{app:retrieval_natural}

We use the six BASED-format validation splits in \Tabref{tab:retrieval_datasets}, with the shared language-model tokenizer. Prompt formatting follows the released \href{https://github.com/HazyResearch/prefix-linear-attention/tree/7490f22bda3e38dc057bcbaffe6bdb09b4d475e6}{\texttt{prefix-linear-attention} evaluation code} at revision \texttt{7490f22bda3e}. SWDE \citep{lockard-etal-2019-openceres} and FDA \citep{arora2023evaporate} append a requested field to the source text. SQuAD \citep{rajpurkar-etal-2016-squad} appends an unfinished statement whose missing ending is the answer, following the released prompt format. TriviaQA \citep{joshi-etal-2017-triviaqa}, Natural Questions (NQ) \citep{kwiatkowski-etal-2019-natural}, and DROP \citep{dua-etal-2019-drop} retain the released context and question formatting.

We tokenize each formatted prompt and retain its last 2,000 tokens. This leaves 48 tokens for generation within the 2,048-token budget. Generation stops at the first newline, the end-of-sequence token, or the output limit. This rule shortens some FDA and NQ inputs; the other four datasets fit without truncation. All 12,129 examples are retained, including cases where truncation removes answer text. We use no answer-centered cropping or filtering based on answer location.

The score is the fraction of generated answers containing a reference string, using case-insensitive matching. It does not require equality with the entire generated string. For the question-answering datasets, this differs from their usual exact-match or token-F1 evaluation. The six-task mean averages task percentages equally within each training seed. Evaluation produces 12,129 predictions per checkpoint.

\begin{table}[!htbp]
\centering
\caption{BASED validation splits and frozen revisions under \texttt{hazyresearch/}.}
\label{tab:retrieval_datasets}
\small
\begin{tabular}{lllr}
\toprule
Task & Dataset & Revision & Examples \\
\midrule
SWDE & \texttt{based-swde-v2} & \texttt{27f0223028a5} & 1,111 \\
SQuAD & \texttt{based-squad} & \texttt{d152c14c2d7d} & 2,984 \\
FDA & \texttt{based-fda} & \texttt{42569d301e12} & 1,102 \\
TriviaQA & \texttt{based\_triviaqa} & \texttt{5c39988c6a2b} & 1,688 \\
NQ & \texttt{based\_nq\_2048} & \texttt{d3338c4e6d8e} & 3,157 \\
DROP & \texttt{based\_drop} & \texttt{ad796806f1b9} & 2,087 \\
\bottomrule
\end{tabular}
\end{table}

\subsection{Selected LongBench tasks}
\label{app:retrieval_longbench}

We use the official LongBench prompts and scorers at code revision \texttt{2e00731f8d0b} and dataset revision \texttt{5e628be450b7}. Qasper \citep{dasigi-etal-2021-dataset} asks questions about research papers, while MultiFieldQA-en covers passages from several domains. SAMSum \citep{gliwa-etal-2019-samsum} summarizes a dialogue, and TREC \citep{li-roth-2002-learning} predicts a question category from demonstrations in the prompt. We retain the released prompt examples and instructions.

For each task, we rank example IDs by the SHA256 hash of the selection seed 1729, task name, and original ID, then select the first 100. The same examples are used at both 1,024- and 2,048-token input budgets. When a prompt is too long, we remove tokens from its middle, keeping the beginning and a suffix long enough to contain the complete final question or dialogue. The suffix contains at least half the input budget and at least the tokenized final input plus 32 tokens. This retains the query but can remove supporting text; examples are not filtered by answer visibility. The budget limits the input alone. Generation allows up to 128 tokens for Qasper and SAMSum and 64 for MultiFieldQA-en and TREC.

Generation stops at the end-of-sequence token or output limit. SAMSum also stops at a newline after generating at least one token. Following the official scorer, SAMSum and TREC discard leading newlines and score the first remaining line. Question answering uses token F1 and summarization uses ROUGE-L, taking the best score over the supplied references. TREC gives full credit for the correct category alone; if several category names occur, it divides credit among them. We scale scores to 0--100 and average the eight task/budget scores within each seed. This gives 800 predictions per checkpoint.

\subsection{Selected BABILong tasks}
\label{app:retrieval_babilong}

We use the official BABILong prompts and answer scorer at code revision \texttt{7a6efee29f5c} and dataset revision \texttt{ee0d588794c7}. QA1 asks for a fact stated in the context; QA2 and QA3 require combining two and three supporting facts, respectively. We use all 100 examples in each released short (\texttt{0k}), \texttt{1k}, and \texttt{2k} setting for these tasks. The released instructions and demonstrations remain in every prompt.

Prompts are not truncated. The release names describe nominal lengths: with our tokenizer and the full prompt, the 1K examples contain 896--1,057 tokens and the 2K examples contain 1,775--2,139 tokens. Short controls contain the task statements without the added long distractor text, but their lengths also vary; QA3 controls can reach 1,252 tokens. The releases are not fully paired across lengths, so their differences can also reflect which questions are included. We report these conditions as a short-context reasoning evaluation; RULER supplies the controlled 4K/8K context extrapolation.

Generation allows 32 new tokens and stops at the end-of-sequence token or this limit. The official scorer lowercases the answer, retains text before the first period or a generated context/example delimiter, and checks whether the reference answer occurs in that text. We average the nine task/length accuracies within each seed. This gives 900 predictions per checkpoint.

\subsection{RULER across context lengths}
\label{app:retrieval_ruler_results}

\begin{figure}[!htbp]
\centering
\includegraphics[width=\linewidth]{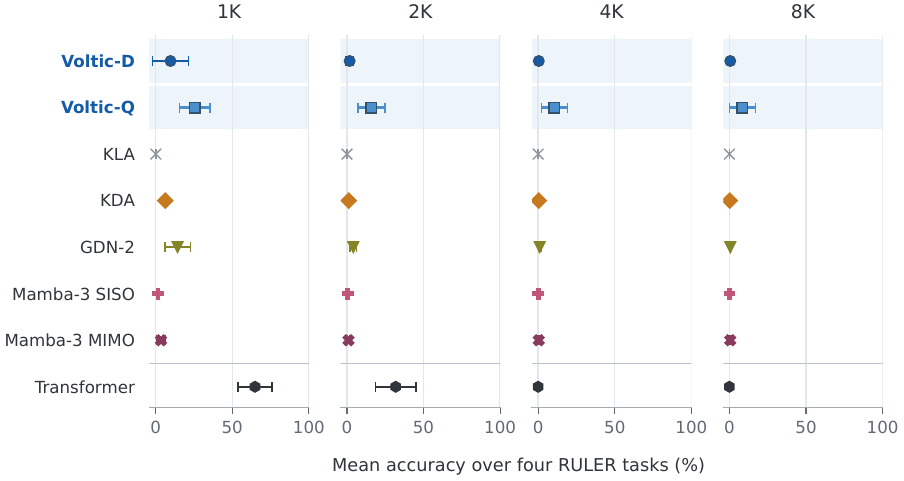}
\caption{45M RULER accuracy averaged over four tasks at each context budget: mean $\pm$ sample SD over three seeds. Training uses 2K sequences.}
\label{fig:retrieval45_contexts}
\end{figure}

\Figref{fig:retrieval45_contexts} compares RULER accuracy at the four context budgets, including the Transformer as a full-attention reference. We average the four task accuracies within each training seed before computing the mean and sample standard deviation across seeds.

\FloatBarrier
\section{Throughput Experimental Details}
\label{app:throughput}

\subsection{Model dimensions and timed operation}
\label{app:throughput_models}

The throughput models use $L=7$, $d_{\mathrm{model}}=512$ at 45M and $L=9$, $d_{\mathrm{model}}=1024$ at 180M. Both use a vocabulary of 32,000 tokens, tied input and output embeddings, and a gated feed-forward branch. We adjust the feed-forward dimension to match parameter targets of 44,620,384 and 180,134,592. Every model is within $0.3\%$ of its target. \Tabref{tab:throughput_models} gives the exact counts.

Voltic, KDA, and GDN-2 use key and value dimensions of 64, with 16 heads at 45M and 32 at 180M. Mamba-3 uses state and head dimensions of 64, expansion two, and rank four for MIMO. KLA uses state dimension 64 and expansion one. Transformer head dimension is 64. The main recurrent budget is 65,536 scalars per layer at 45M and 131,072 at 180M: this counts the mean memory for matrix recurrences and the two sufficient-statistic vectors for KLA. Auxiliary uncertainty, convolution state, and intermediate activations are additional storage.

The training and decoding measurements reuse these configurations. Models start from seed 1701; weights are fixed throughout timing. All architectures receive the same synthetic token IDs at a given input shape. In training measurements, there is no document packing or recurrent state carried between operations. Each operation runs \texttt{zero\_grad(set\_to\_none=True)}, causal forward computation, FP32 full-vocabulary cross-entropy, and backward. Throughput counts $BT$ input tokens per operation; next-token loss uses $B(T-1)$ targets. Cached decoding is measured separately below.

\begin{table}[!htbp]
\centering
\caption{Dedicated throughput configurations: seven layers at 45M and nine at 180M, with matched parameter budgets.}
\label{tab:throughput_models}
\small
\begin{tabular}{lrrrr}
\toprule
 & \multicolumn{2}{c}{45M} & \multicolumn{2}{c}{180M} \\
\cmidrule(lr){2-3}\cmidrule(lr){4-5}
Model & Parameters & $d_{\mathrm{ff}}$ & Parameters & $d_{\mathrm{ff}}$ \\
\midrule
\rowcolor{VolticTint}
Voltic-D & 44,684,224 & 224 & 180,152,448 & 528 \\
\rowcolor{VolticTint}
Voltic-Q & 44,619,600 & 560 & 180,281,184 & 1,216 \\
KLA & 44,680,199 & 1,968 & 180,142,089 & 4,112 \\
KDA & 44,648,496 & 1,120 & 180,060,000 & 2,448 \\
GDN-2 & 44,705,840 & 448 & 180,354,912 & 1,104 \\
Mamba-3 SISO & 44,620,384 & 1,536 & 180,134,592 & 3,200 \\
Mamba-3 MIMO & 44,749,408 & 1,408 & 180,466,368 & 3,072 \\
\midrule
Transformer & 44,547,584 & 1,936 & 180,243,456 & 3,968 \\
\bottomrule
\end{tabular}
\end{table}

\subsection{Execution settings}
\label{app:throughput_execution}

\Appref{app:resources} gives the shared software versions and hardware. BF16 autocasting is enabled, TF32 is disabled, and FP32 matrix multiplication uses the highest precision setting. Native kernels retain their supported internal precision; KLA's recurrence is FP32.

Execution settings are selected before the final timing repetitions. Every baseline is offered eager execution and compilation of supported model components. Voltic also tests panel lengths, short-convolution backends, separate or combined query/key/value input preparation, and the available Voltic-Q launch settings. Selection uses synchronized trial times after warmup; its measurements are excluded from the reported results. The batch-size study retains the selected settings at each model size and sequence length, without retuning them for each batch.

Voltic uses fused noise rates and separate query, key, and value convolution calls in the selected settings. Voltic-D uses $S=512$ throughout. Voltic-Q panel lengths and convolution choices are listed in \Tabref{tab:throughput_plans}. At state dimension 64, Voltic-Q holds a block's update vectors in GPU registers while computing its token-ordered forward updates.

KDA and Mamba-3 SISO use eager execution at both scales; GDN-2 uses eager execution at 45M and compiled model components at 180M. Eager here describes the outer model: KDA, GDN-2, and Mamba-3 still call their native accelerated recurrence kernels. Mamba-3 MIMO and the Transformer use compiled model components. Voltic compiles the projections and output operations around custom kernels, except for the 45M Voltic-D 1K setting, which compiles the full model.

KLA training uses the unmodified official \texttt{cuda\_v3\_fast} backend at revision \texttt{ad5e64de903e}, replacing the earlier \texttt{triton\_composed} measurements. The original weights are loaded strictly into the updated model; dimensions, parameter counts, noise parameterization, and inputs are retained. Its execution settings are selected again over eager execution, compiled components, and compiled components with scan checkpointing, using five warmup and five pilot operations. Compiled components without additional scan checkpointing are selected at all sizes and lengths; the CUDA kernel's internal recomputation remains enabled. Each selection is fixed across batches. Thus the comparison uses optimized baseline kernels, not Python reference recurrences, without claiming that every implementation is equally optimized.

\begin{table}[!htbp]
\centering
\caption{Selected Voltic-Q $S$/convolution backend: C = causal-conv1d, F = FLA Triton, P = PyTorch. Voltic-D uses 512/C throughout.}
\label{tab:throughput_plans}
\small
\begin{tabular}{llrrrr}
\toprule
Scale & Variant & 1K & 2K & 4K & 8K \\
\midrule
45M & Voltic-Q & 1024/C & 256/C & 512/C & 512/F \\
180M & Voltic-Q & 1024/F & 2048/F & 256/C & 512/C \\
\bottomrule
\end{tabular}
\end{table}

\subsection{Training timing and checks}
\label{app:throughput_training}

The measured training grid covers 256 combinations: two scales, eight models, four sequence lengths, and batches in $\{1,2,4,8\}$. The fixed-token view selects its 64 entries with $BT=8,192$; these are not additional independent measurements. All reported training and decoding timings use the same physical H200 NVL. The updated KLA measurements replace its previous backend, and newly completed large-batch cases fill the previous omissions; other batch-grid measurements are retained.

Each combination is measured in three fresh processes. Model and shape order are randomized within each repetition using seeds 9013--9015. Ten warmup operations precede fifty synchronized wall-clock measurements. For each process, we divide the input-token count by its mean operation time, then report the mean and sample standard deviation of the three process throughputs. The 150 operations are not treated as 150 independent runs. Compilation, optimizer updates, data loading, packing, and gradient accumulation are excluded.

Before timing, each selected execution setting is compared with its accepted eager implementation at the full input shape, checking outputs, loss, and parameter gradients. KLA's updated backend is compared with its previous Triton implementation and additionally checks recurrence outputs, variance, and input gradients against the upstream sequential reference. Voltic checks complete-sequence memory outputs and output-loss gradients against an independent FP64 recurrence at sampled first and final layers. These checks apply to the measured training operation, which does not reuse returned states.

\textbf{Reference-memory limits.} Where the whole-batch reference runs out of memory, we compute each independent row separately, concatenate all logits, and average the equal-length row losses and parameter gradients. This reference procedure is first checked against whole-batch execution at length 1K and batch two, then used at the complete target shape with unchanged numerical bounds. The candidate is qualified and timed with the actual full batch, never split into smaller timed batches. Together with the updated KLA backend, this completes all 256 measured combinations, including the previously omitted 180M Voltic cases at 8K and batch eight. Their first- and final-layer sequence-output/gradient oracle checks pass, but the separately recorded final-state API comparisons exceed their oracle bounds for both Voltic variants. The training timing checks therefore do not certify that separate API or arbitrary cached execution.

\subsection{Training throughput}
\label{app:throughput_results}

\begin{figure}[!htbp]
\centering
\includegraphics[width=\linewidth]{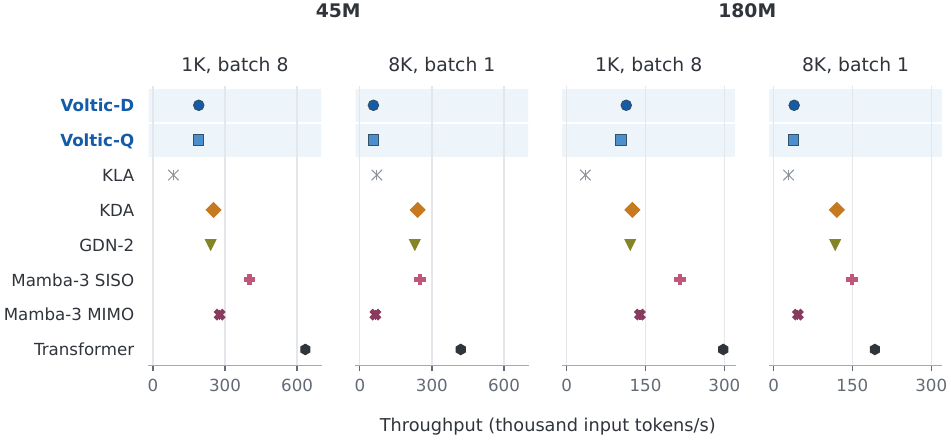}
\caption{Training throughput at 8,192 input tokens per operation: mean $\pm$ sample SD over three timing processes.}
\label{fig:throughput_sequence}
\end{figure}

\begin{figure}[!htbp]
\centering
\includegraphics[width=\linewidth]{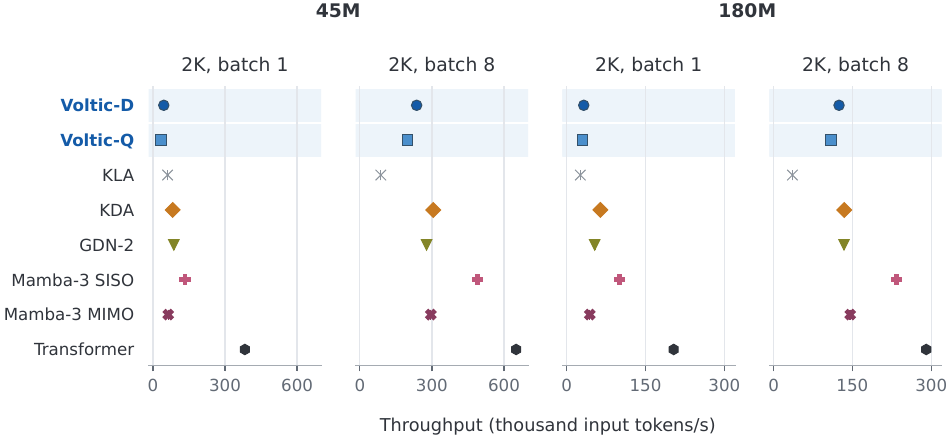}
\caption{Training throughput at 2K sequence length, batches one and eight: mean $\pm$ sample SD over three timing processes.}
\label{fig:throughput_batch}
\end{figure}

The fixed-token view in \Figref{fig:throughput_sequence} uses the entries with $BT=8,192$, so length and batch size change together. \Figref{fig:throughput_batch} holds sequence length at 2K and compares batches one and eight at both model scales.


\FloatBarrier
\subsection{Cached autoregressive decoding}
\label{app:throughput_decoding}

We measure all eight configurations at both scales, with batch one, prompt lengths 1K, 2K, 4K, and 8K, and 128 cached decode steps per generation. The prompt cache is reconstructed before each timed block. Time per output token (TPOT) is the synchronized wall-clock time for the complete decode block divided by 128. It includes host dispatch, every model layer, the full vocabulary projection, and GPU greedy selection. It excludes prompt processing, tokenization, networking, and backward computation. Each of three fresh timing processes uses two warmup generations and five measured generations. We average each process's five block times before computing the three-process mean and sample standard deviation. These output-token latencies are not training input-token throughput.

\textbf{Cached implementations.} Voltic uses its Triton recurrent inference path; the Voltic-Q cache retains within-block information and chunk phase. KDA and GDN-2 use their chunked prefill and fused recurrent decode kernels. Mamba-3 uses its official cached step kernels with a singleton-sequence shape adapter. The Transformer uses its frozen projections, rotary positions, normalization, and feed-forward layers with a key/value-cache adapter and PyTorch scaled dot-product attention. No full-prefix recomputation is substituted for cached decoding. KLA uses its updated official release, but its CUDA training backend cannot return a prefill state: cached generation instead uses the official state-producing Triton prefill and \texttt{kla\_step}. We do not label these KLA decoding results as CUDA-v3-fast timings.

\textbf{Checks and limits.} Before timing, cached teacher-forced logits are compared with complete-sequence computation at all 64 continuation positions following each full-length prompt, including Voltic-Q block boundaries. Admission requires finite logits and global and per-token relative $\ell_2$ error at most $0.05$. Pointwise errors and argmax agreement are recorded as diagnostics, not additional pass criteria. All 64 tested configurations pass this finite continuation check; it does not establish exact greedy-token agreement for arbitrary prompts or replace a final-state oracle certificate.

\textbf{Results.} \Figref{fig:throughput_main} shows the decoding latencies for 2K prompts. Across the measured 1K--8K range, most latencies change little. KLA's 8K points were measured separately after increasing only the wrapper's length limit to cover the prompt and continuation. Its roughly $18\%$ lower 8K latency than the earlier 1K--4K points has no established cause; we do not interpret this as improved scaling with context length. Weights, kernels, precision, timing rules, and physical GPU were unchanged.

\end{document}